\documentclass{bmvc2k}

\usepackage{graphicx} 
\usepackage{mathptmx} 
\usepackage{times} 
\usepackage{amsmath} 
\usepackage{amssymb}  
\usepackage{multirow}
\usepackage{makecell}
\usepackage{pifont}
\usepackage{color}
\usepackage{colortbl}
\usepackage{xcolor}
\usepackage[export]{adjustbox}
\usepackage{booktabs}
\usepackage{arydshln}
\usepackage{soul}
\usepackage{xspace}
\usepackage{upgreek}
\usepackage{textcomp}
\usepackage{gensymb}
\usepackage{enumitem}
\setlist[itemize]{noitemsep, nolistsep}
\usepackage[
acronyms,
nohypertypes={acronym},
shortcuts,
nonumberlist,
nogroupskip,
nopostdot
]{glossaries}
\usepackage[sort&compress,capitalize,nameinlink]{cleveref}
\graphicspath{{./figures}}

\newcommand{\name}{\mbox{EgoFlowNet}} 
\newcommand*{\cf}{\emph{c.f.}\@\xspace}

\title{\name{}: Non-Rigid Scene Flow from Point Clouds with Ego-Motion Support}

\addauthor{Ramy Battrawy}{ramy.battrawy@dfki.de}{1}
\addauthor{Ren{\'e} Schuster}{rene.schuster@dfki.de}{1}
\addauthor{Didier Stricker}{didier.stricker@dfki.de}{1,2}

\addinstitution{
	Augmented Vision\\
	German Research Center for\\ 
	Artificial Intelligence (DFKI)\\
	Kaiserslautern, Germany
}

\addinstitution{
	Computer Science Department\\
	The University of Kaiserslautern-Landau (RPTU)\\
	Kaiserslautern, Germany
}

\runninghead{Battrawy, Schuster, Stricker}{\name{}}

\def\eg{\emph{e.g}\bmvaOneDot}

\def\etal{\emph{et al}\bmvaOneDot}
\def\ie{\emph{i.e}\bmvaOneDot}
\def\etc{\emph{etc}\bmvaOneDot}

\newcommand{\mytilde}{{\raise.17ex\hbox{$\scriptstyle\sim$}}}

\newacronym{fps}{FPS}{Farthest-Point-Sampling}
\newacronym{rs}{RS}{Random-Sampling}
\newacronym{ft3d}{$\mathrm{FT3D_s}$}{FlyingThings3D subset}
\newacronym{ft3do}{$\mathrm{FT3D_o}$}{FlyingThings3D}%

\newacronym{ft3d_bold}{$\mathrm{\mathbf{FT3D_s}}$}{FlyingThings3D Subset_}

\newacronym{lfa}{LFA}{Local-Feature-Aggregation}
\newacronym{knn}{KNN}{K-Nearest-Neighbor}
\newacronym{us}{US}{Up-Sampling}
\newacronym{ds}{DS}{Down-Sampling}
\newacronym{fe}{FE}{Flow-Embedding}
\newacronym{wl}{WL}{Warping-Layer}
\newacronym{max}{Max-Pooling}{Max-Pooling}
\newcommand\Tstrut{\rule{0pt}{2.1ex}}         
\newcommand\Bstrut{\rule[-0.9ex]{0pt}{0pt}}   
\newcommand{\cmark}{\ding{51}}%
\newcommand{\xmark}{\ding{55}}%

\begin{document}

\maketitle

\begin{abstract}
	Recent weakly-supervised methods for scene flow estimation from LiDAR point clouds are limited to explicit reasoning on object-level.
    These methods perform multiple iterative optimizations for each rigid object, which makes them vulnerable to clustering robustness.
	In this paper, we propose our \name{} -- a \textbf{point-level} scene flow estimation network trained in a weakly-supervised manner and without object-based abstraction.
	Our approach predicts a binary segmentation mask that implicitly drives two parallel branches for ego-motion and scene flow.
	Unlike previous methods, we provide both branches with all input points and carefully integrate the binary mask into the feature extraction and losses.
    We also use a shared cost volume with local refinement that is updated at multiple scales without explicit clustering or rigidity assumptions.
	On realistic KITTI scenes, we show that our \name{} performs better than state-of-the-art methods in the presence of ground surface points.
\end{abstract}

\section{Introduction}
\label{sec:intro} \glsresetall
Scene flow estimation is an important computer vision problem for navigation, planning, and autonomous driving systems.     
It provides a representation of the dynamic environment by estimating the 3D motion field relative to the observer. 
Until a few years ago, stereo images were used for joint disparity estimation and optical flow estimation to represent scene flow~\cite{saxena2019pwoc, jiang2019sense, ma2019deep, chen2020consistency, schuster2020sceneflowfields++}. 
However, the two-view geometry used in self-driving cars has inherent limitations, such as inaccuracies in depth estimation in distant regions.

With the advent of LiDAR, many learning-based methods have been developed to estimate scene flow directly from point clouds in a fully-supervised manner~\cite{gu2019hplflownet, wu2020pointpwc, cheng2022bi, wang2022matters, kittenplon2021flowstep3d}.
They differ from each other in their basic feature extraction framework and the way they design their cost volume.
Due to the lack of annotated data on realistic sequences, some methods train their end-to-end models with self-supervised losses~\cite{wu2020pointpwc, kittenplon2021flowstep3d, mittal2020just, li2021self}. 

\begin{figure*}[t]
	\begin{center}
		\includegraphics[width=1.0\linewidth]{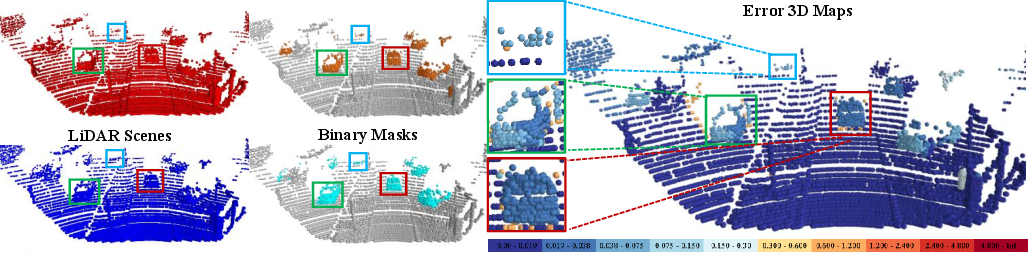}
		\caption{Our \name{} operates non-rigidly at the \textbf{point-level} and shows high accuracy for regions of varying local density (\eg, the red, blue, and green rectangles).}
		\label{Figure2_Teaser}
	\end{center}
     \vspace{-5mm}
\end{figure*}

Apart from the point-wise estimation of scene flow, some methods perform better when using self-supervised losses under conditions of rigidity~\cite{li2022rigidflow, deng2023rsf}.
Other methods support scene flow estimation with ego-motion~\cite{behl2019pointflownet, tishchenko2020self, wang2022unsupervised}.
All of the above methods work well on ideal conditions (\eg, no ground points, no occlusions, or with nearly direct correspondences between consecutive scenes).

A recent breakthrough has been achieved by WSLR~\cite{gojcic2021weakly}, where a multi-task prediction network is designed and trained with real scenes in a weakly-supervised manner in the presence of ground points.
This approach segments the scene into static parts (\ie background ($BG$)) and moving agents (\ie foreground ($FG$)). 
It then optimizes the initial estimate of ego-motion and scene flow via non-parametric object-based optimizations using explicit rigidity constraints. 
Towards learning-based optimization, ERC~\cite{dong2022exploiting} uses the predicted segmentation mask from \cite{gojcic2021weakly} and proposes a novel optimization method with an error-driven Gated Recurrent Unit and residual scene flow heads.
These methods \cite{gojcic2021weakly, dong2022exploiting} show impressive results for more difficult scenes (\eg with ground points, outliers, occlusions, \etc).
However, both methods rely on the DBSCAN clustering algorithm \cite{ester1996density}, which may limit their ability to work on low-density regions or under-sampled objects (\eg, distant cars or small objects).
In addition, they must perform iterative optimizations for each clustered region, which negatively impacts efficiency when the scene contains a large number of clusters.

Compared to these methods, our approach is far removed from any clustering strategy and instead predicts unconstrained scene flow at the point-level.
To this end, we design our multi-task network to predict a binary $FG$/$BG$ segmentation mask, which is then carefully used to estimate ego-motion and scene flow (\cf \Cref{Figure2_Teaser}).
Unlike object-based methods, we feed the ego-motion and scene flow branches with all input points, integrate our predicted mask into both branches and combine everything with point-based coarse-to-fine refinement to obtain accurate scene flow.
For robust estimation in both branches, we also develop a hybrid feature extraction to provide both branches with well-suited features. 

Our contributions are summarized as follows:
\begin{itemize}[noitemsep] 
	\item We propose \name{} -- a multi-task neural network architecture to estimate scene flow directly from \textit{raw} point clouds that jointly estimates binary segmentation masks, ego-motion, and scene flow. 
	\item We propose a hybrid feature extraction along with a hybrid warping layer and integrate the binary masks to obtain robust scene flow.
	\item We work with a point-level refinement of the scene flow, which is free of explicit clustering mechanisms or rigidity assumptions for dynamic objects.
	\item On difficult real LiDAR scenes (\ie, with ground points, occlusions, and outliers), we show that our proposed approach outperforms recent clustering-based methods.
\end{itemize}

\section{Related Work}
\label{sec:related}
3D scene flow was first introduced in the image domain using RGB-D~\cite{jaimez2015primal, jaimez2017fast, qiao2018sf, shao2018motion} for indoor scenarios and stereo images~\cite{vogel20153d, ilg2018occlusions, jiang2019sense, ma2019deep, chen2020consistency, schuster2020sceneflowfields++, teed2021raft} for outdoor scenarios.
However, learning scene flow directly from point clouds without relying on RGB images opens up a wide field of research~\cite{wang2018deep, liu2019flownet3d, gu2019hplflownet, wu2020pointpwc, puy20flot, wei2020pv, kittenplon2021flowstep3d, li2021hcrf, wang2021hierarchical, gojcic2021weakly, gu2022rcp, cheng2022bi, dong2022exploiting, deng2023rsf}.

\textbf{GRU-based Scene Flow from Point Cloud:}
The Gated Recurrent Unit (GRU) \cite{cho2014learning} is used to iteratively refine the global cost volume to provide an accurate estimate of the scene flow \cite{teed2021raft}.
FlowStep3D \cite{kittenplon2021flowstep3d} updates the cost volume locally using GRU with multiple reconstructions and iterative point cloud alignment.   
To encode a large correspondence set within the cost volume, PV-RAFT~\cite{wei2020pv} combines a voxel representation with a point-wise cost volume.
A point-wise optimization combined with a recurrent network regularization is proposed by RCP~\cite{gu2022rcp}.
Our \name{} avoids strict iterative updates and works from coarse-to-fine, driven by a binary segmentation mask and jointly estimates scene flow and the ego-motion.

\textbf{Hierarchical Scene Flow from Point Cloud:}
FlowNet3D~\cite{liu2019flownet3d}  is the first work to introduce a cost volume layer from a point cloud with hierarchical refinement. However, it is limited to a single cost volume layer.
To overcome this limitation, HPLFlowNet~\cite{gu2019hplflownet} introduces multi-scale correlation layers by projecting points into a permutohedral grid~\cite{su2018splatnet}. 
Moving away from the grid representation, PointPWC-Net~\cite{wu2020pointpwc} improves the direct estimation of scene flow from \textit{raw} point clouds by constructing cost volumes at a range of scales from coarse-to-fine. 
Following the hierarchical point-based designs, intensive improvements are proposed in the development of cost volume using dual attentions as in~\cite{wang2021hierarchical, wang2022residual, wang2022matters, battrawy2022rms}.
Our network is basically hierarchical, but integrates further multi-task estimates of ego-motion and segmentation. 
It operates in challenging outdoor scenes with typical occlusions and in dense scenes with ground points.

\textbf{Scene Flow from Point Cloud with Constraints:} 
Axiomatic concepts of rigidity assumptions are explored in~\cite{deng2023rsf, li2022rigidflow} along with cluster-based or object-level optimization.
However, the above methods are not well explored with typical outdoor scenes in the presence of ground surface points. 
More recently, WSLR~\cite{gojcic2021weakly} has proposed pioneering weakly-supervised learning along with non-parametric optimization, and ERC~\cite{dong2022exploiting} extends this to learning-based optimization.
Both work well on challenging outdoor scenes where ground points are present. 
However, both require multiple optimization steps and work under object-level constraints using DBSCAN clustering~\cite{ester1996density}.
Chodosh~\etal~\cite{chodosh2023re} is a very recent conventional and cluster-based method that works by test time optimization using ICP~\cite{besl1992method, segal2009generalized} and RANSAC to achieve appropriate piece-wise rigidity.
In contrast to the above methods, we do not use clustering algorithms and work with point-level optimization, which allows us to estimate non-rigid motion and is more accurate and robust than state-of-the-art methods.

\section{Network Design}
\label{sec:network}
\begin{figure*}[t]
	\begin{center}
		\includegraphics[width=1.0\linewidth]{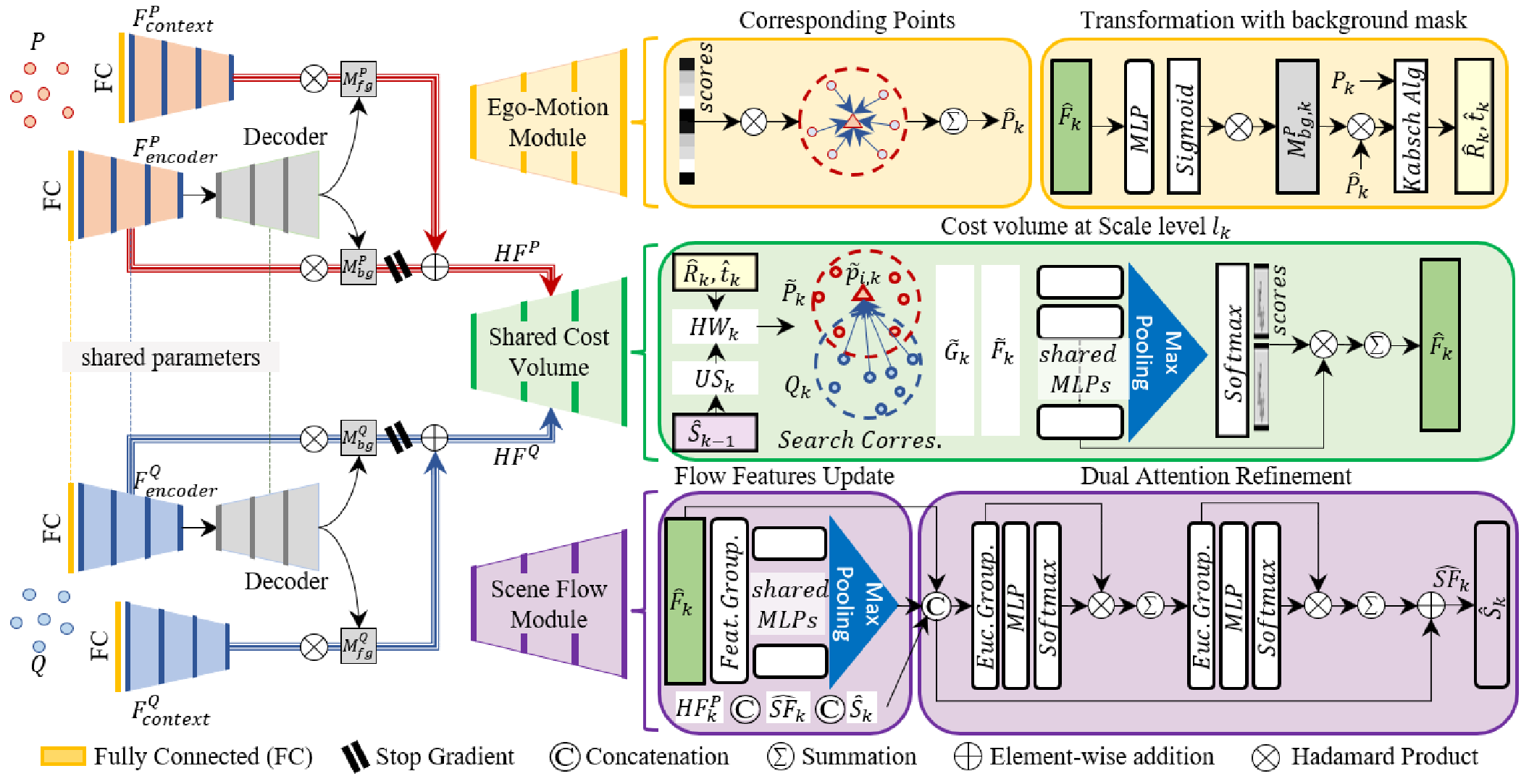}
		\caption{Our \name{} architecture predicts a binary segmentation masks ($M^P_{fg}$ and $M^Q_{fg}$) for foreground points ($FG$) and ($M^P_{bg}$ and $M^Q_{bg}$) for background points ($BG$). We use the binary mask to jointly estimate ego-motion and scene flow at the point-level. For this, we extract hybrid features ($HF^P$ and $HF^Q$) and hierarchically refine our point-wise scene flow.}
		\label{Figure3_Pipeline}
	\end{center}
     \vspace{-5mm}
\end{figure*}

Our \name{} estimates scene flow as translational vectors from two consecutive frames of point clouds, with no assumptions about object rigidity.
Given Cartesian 3D point cloud frames $P$ and $Q$ at timestamps $t$ and $t+1$, our goal is to estimate point-wise 3D flow vectors $\hat{S}$ for each point within $P$.
Our network is designed to combine segmentation, ego-motion, and scene flow tasks at four scales $\{L\}^3_{k=0}$, where $l_0$ is the full resolution of $P$ and $Q$.
An illustration of our hierarchical modules of our feature extraction, cost volume, ego-motion and scene flow is given in \Cref{Figure3_Pipeline}, where the right part of the figure illustrates a single layer or scale $l_k$ for each of these modules.
The following sections describe the components of each module in detail. 
\subsection{Feature Extraction} \label{feature_extraction}
Our feature extraction module consists of two networks: The first one is an encoder-decoder module, while the second one consists only of a context encoder.
The backbone of our feature extraction is inspired by RandLA-Net~\cite{hu2020randla}.

\textbf{Encoder Module:} Each scale in the encoder module essentially consists of two layers, where \gls{lfa} is applied to aggregate the features at the $l_k$ scale, followed by a \gls{ds} layer to aggregate the features from the $l_{k}$ level to $l_{k+1}$, resulting in a simultaneous decrease in resolution.
The backbone of the \gls{lfa} is inspired by RandLA-Net \cite{hu2020randla}, which uses attentive pooling based on self-attention as in \cite{yang2020robust, zhang2019pcan}.
At all scales, we search for 16 neighbors in Euclidean space using \glspl{knn}, then their weighted features are summed based on attentive pooling. 
The \Gls{ds} layer samples the points based on \gls{fps} to the defined resolution $l_{k+1}$, and aggregates $16$ nearest neighbors in the higher resolution $l_{k}$ for each selected sample simply by using \acrshort{max}. 

\textbf{Decoder Module:} The decoder module of the hourglass network consists of ${\{L\}}^3_{k=0}$ layers for extracting the features up to the full (input) resolution $l_0$ of $P$ and $Q$, respectively. 
To up-sample from the $l_{k+1}$ level to $l_k$, we simply assign the one nearest neighbor for each point of the higher resolution to the lower one, followed by a simple Multi-Layer Perceptron (MLP).  
To increase the quality of the features in the encoder-decoder network, lateral connections are added to each layer.

\textbf{Segmentation Features:} The encoder-decoder of the first network extracts the features $F^P_{s,0}$ and $F^Q_{s,0}$ at the input resolution, which are used to predict the binary segmentation masks ($M^P_{fg}$ and $M^Q_{fg}$) and/or ($M^P_{bg}$ and $M^Q_{bg}$) for $P$ and $Q$, respectively.

\textbf{Hybrid Features:}
The encoder module of the first network and the context network compute the features ($F^P_{encoder,k}$, $F^Q_{encoder,k}$) and ($F^P_{context,l}$, $F^Q_{context,k}$), at each scale level $l_k$.
The encoder modules down-sample the input to the resolutions $l_1 = 2048, l_2 = 512, l_3 = 128$ with feature dimensions $C_0 = 32, C_1 = 128, C_2 = 256, C_3 = 512$.
The output features of the two encoders at each scale level $l_k$ are merged using the predicted and down-sampled segmentation masks $M_{fg, k}$ as follows:
\begin{equation}
	\begin{aligned}
		 HF_{k} = M_{fg,k} \cdot{F_{context,k}} + (1 - M_{fg,k})\cdot{\perp{(F_{encoder,k})}},
		\label{eq_features}
	\end{aligned}	
\end{equation}
where $M_{fg,k}$ refers to the binary mask of foreground points, ($1 - M_{fg,k}$) refers to the background mask (\ie, $M_{bg,k}$) and $\perp$ is an operator that sets the gradient of the operand to zero, $\triangledown{(x)} = 0$~(\ie, stop gradient).
Since the number of background points ($BG$) within a scene, including the ground points, is usually much higher than the number of foreground points ($FG$), using the stop gradient eliminates the negative effect of the ego-motion branch on the segmentation head and the scene flow branch. 
By merging the context encoders, the features of the $FG$ points can be enhanced to provide an accurate estimate of scene flow for these points.
We apply \cref{eq_features} at each scale level $l_k$ using $M^P_{fg,k}$ and $M^Q_{fg,k}$, resulting in $HF^P_{k}$ and $HF^Q_{k}$ for $P$ and $Q$, respectively, which are then used for the shared cost volume.

\subsection{Segmentation Head} \label{sub:segmentation_head}
We apply three layers of Multi-Layer Perceptions (MLPs) with 64, 32 and 1 output channels to the computed segmentation features $F^P_{s,0}$ and $F^Q_{s,0}$.
The output of the last layer provides the segmentation probabilities at full (input) resolution layer $l_0$, allowing us to define the binary segmentation masks $M^P_{fg}$ and $M^Q_{fg}$ for $P$ and $Q$, respectively.

\subsection{Shared Cost Volume} \label{cost_volume}
We learn the geometric and feature correlations based on the hybrid features $HF^P_{k}$ and $HF^Q_{k}$ (\cf~\cref{eq_features}).

\textbf{Searching for Correspondences:} As a first step, we need to find the correlation set in $Q_k$ for each point in $P_k$.
Since finding correlations based on Euclidean space may not be sufficient to capture distant correspondences, we use the feature space to find the correspondences at the coarsest scale $l_{3}$.
This provides a high quality initial estimate of the scene flow $\hat{S}_{3}$ and a high quality initial estimate of the ego-motion parameters represented by the rotation $\hat{R}_{3}$ and translation $\hat{t}_3$ components. 
For the scene flow, when searching for correspondences in feature space, the distant matches on the upper scales are approximated by our hybrid warping layer so that the warped point cloud $\widetilde{P}_{k}$ is close to its match in $Q_{k}$.
With this initialization, it becomes worthwhile to search for the closest $16$ matches in Euclidean space for the upper scales $\{L\}^2_{k=0}$.
 
After grouping the correspondence set with its geometric features ${\{q_{j,k}\}}^{16}_{j=1}$ and hybrid features ${\{{hf}^{q}_{j,k}\}}^{16}_{j=1}$, we compute the differences to the point $P_k$ and its hybrid feature ${hf}^P_{i,k}$, respectively. 
This yields the geometric and feature differences $\widetilde{G}_{k}$ and $\widetilde{F}_{k}$, which are then concatenated.
We apply~\acrshort{max} along the feature dimension to compute attentive weights similar to HRegNet~\cite{lu2021hregnet}.
The geometric and feature differences are then smoothly weighted by the attentional weights and summed to obtain ${\hat{F}}_{k}$.

\textbf{Hybrid Warping Layer:}
Our hybrid warping layer (${HW}_{k}$) is jointly driven by the ego-motion, the scene flow, and the predicted segmentation $M^P_{fg,k}$ of frame $P_k$.
After obtaining the initial ego-motion and the initial scene flow at the coarse scale $l_3$, we apply hybrid warping to refine the estimate at the upper scales $\{L\}^2_{k=0}$. 
Fo this purpose, we use the the corresponding binary masks ($M^P_{fg,k}$ and $M^P_{bg,k}$), the up-sampled scene flow ${\hat{S}}_k$ from the coarser scale through a simple Up-Sampling layer (${US}_{k}$), and the ego-motion transformation to warp the points in $P_k$ towards the target $Q_k$ and obtain ${\widetilde{P}}_k$.
For all upper scales, we apply the following equation:
\begin{equation}
	\begin{aligned}
		{\widetilde{P}}_k = M_{fg,k} \cdot{(P_k + {\hat{S}}_k})  + (M_{bg,k} \cdot{((\hat{R}_kP^T_k)^{T} + \hat{t}_k)}
		\label{eq_warping}
	\end{aligned}	
\end{equation}

\subsection{Ego-Motion branch} \label{ego_motion}
We compute point correspondences and apply the Kabsch algorithm~\cite{kabsch1976solution} to estimate the ego-motion parameters $\hat{R}_k$ and $\hat{t}_k$.

\textbf{Corresponding Points:}
Inspired by HRegNet~\cite{lu2021hregnet}, after obtaining the correspondence set ${\{q_{j,k}\}}^{16}_{j=1}$ for each point in $Q_k$ as described in the cost volume, we multiply the computed attentive weights with them and sum over the nearest neighbors to obtain the corresponding points $\hat{P}_{k}$.

\textbf{Optimal Transformation:}
Multi-Layer Perceptrons (MLPs) are applied to the cost volume output ${\hat{F}}_{k}$, followed by a Sigmoid function to obtain confidence values inspired by \cite{lu2021hregnet, gojcic2021weakly, dong2022exploiting}.
However, unlike previous approaches~\cite{dong2022exploiting, gojcic2021weakly}, we do not filter out the ($FG$) points to feed the ego-motion branch with only ($BG$) points.
Instead, we feed this branch with all points and multiply the confidence values by $M^P_{bg,k}$, to refine the corresponding points $\hat{P}_{k}$ so that the transformation matrix can be computed according to \cite{kabsch1976solution} to obtain $\hat{R}_k$ and $\hat{t}_k$.     

\subsection{Scene Flow branch} \label{scene_flow}
Across all scales, our scene flow branch consists of three refinement stages and four scene flow predictors with simple nearest-neighbor \acrfull{us}.
The total number of layers with attention-based refinement is inspired by RMS-FlowNet~\cite{battrawy2022rms}, which is designed to estimate scene flow only, but we add three feature updating units.
 
\textbf{Feature Updates:}
With the obtained cost volume features ${\hat{F}}_{k}$, we search for the 16 nearest neighbors in the feature space, group them and then apply MLPs followed by \acrshort{max}. 
This helps to capture similar features and implicitly extends features to semantic objects as inspired by DGCNN~\cite{wang2019dynamic}. 

\textbf{Dual Attention Refinement:}
We concatenate the updated features ${\hat{F}}_{k}$ with $HF^P_{k}$, $\hat{SF}_{k}$ and $\hat{S}_k$, where the latter two components are the scene flow features and the scene flow, respectively. Both are initialized to zero in the coarse layer $l_3$ and are only used in the upper scales ${\{L\}}^2_{k=1}$. 
We use the defined nearest neighbors ($16$) in Euclidean space to group the concatenated features and we apply dual attentions to refine the corresponding features as performed in \cite{battrawy2022rms}.

\textbf{Scene Flow Predictor:}
Our \name{} predicts scene flow at multiple scales, inspired by~\cite{wu2020pointpwc, battrawy2022rms}. 
The scene flow estimation head takes the resulting scene flow features at each scale $\hat{SF}_{k}$ and applies three layers of MLPs with 64, 32 and 3 output channels.
Then, the estimated scene flow $\hat{S}_k$ and the features from the attention-based refinement are up-sampled to the next higher scale using a simple \acrshort{knn} search.

\subsection{Scene Flow of $\mathbf{BG}$ Points} \label{sub:final_sceneflow}
At the input point resolution $l_0$, we compute the scene flow of the background $BG$ from the predicted rotation and translation (\ie, $\hat{R}_0$ and $\hat{t}_0$) of the ego-motion branch. We use the binary segmentation mask $M^P_{bg}$ to merge the scene flow of the $BG$ points with the output of the scene flow $\hat{S}_0$ obtained from the scene flow branch. 

\subsection{Loss Function} \label{losses}
To guide the training of segmentation, ego-motion, and scene flow, we combine three losses:
\begin{equation}
	\begin{aligned}
		\mathcal{L}_{total} = \mathcal{L}_{seg} + \mathcal{L}_{ego} + \mathcal{L}_{sf},
		\label{eq_totalloss}
	\end{aligned}	
\end{equation}

\textbf{Segmentation Loss:}
We use the Weighted Binary Cross-Entropy loss to overcome the severe imbalance of $FG$ and $BG$ classes as follows:
\begin{equation}
	\begin{aligned}
		\mathcal{L}_{seg} = -\frac{1}{N}\sum_{i=1}^{N} \gamma\cdot{y_i}\cdot{log(\sigma{(p(y_i))})} + (1-y_i)\cdot{log(1-\sigma{(p(y_i))})},
		\label{eq_segmentloss}
	\end{aligned}	
\end{equation}
where $i$ is the index in $P$ or $Q$, $y_i$ is the ground truth label, $p(y_i)$ is the probability of the predictions and $\gamma$ is the $FG$ class weight, which is set to 20.

\textbf{Ego-Motion Loss:}
Inspired by \cite{lu2021hregnet}, the ego-motion loss is computed hierarchically.
Given a four-scale estimate of the transformation parameters $\hat{R}_k$ and $\hat{t}_k$ and the ground truth $R$ and $t$, we compute the final ego-motion loss as follows: 
\begin{equation}
	\begin{aligned}
		\mathcal{L}_{ego} = \frac{1}{4} \sum_{k=0}^{3} \beta {\| {\hat{R}}^{T}_{k} R - I\|}_2 + {\| {\hat{t}}_{k} - t \|}_2
		\label{eq_egoloss}
	\end{aligned}	
\end{equation}
where ${\|.\|}_2$ denotes the $L_2$-norm and $\beta$ is set to $1.8$.

\textbf{Scene Flow Losses}:
To train the scene flow branch, we apply a bidirectional Chamfer loss $\mathcal{L}_{cd,k}$ and Smoothness loss $\mathcal{L}_{sm,k}$ per scale, both driven by $M^P_{fg,k}$ and $M^Q_{fg,k}$ as follows:
\begin{equation}
	\begin{aligned}
		\mathcal{L}_{cd,k} = \sum_{\tilde{p_k} \in {\tilde{P}}_k} m^P_{fg,k}\cdot{\underset{q_k \in Q_k}{min} {\| {\tilde{p}}_k - q_k\|}_2} + \sum_{q_k \in Q_k} m^Q_{fg,k}\cdot{\underset{{\tilde{p}}_k \in {\tilde{P}}_k}{min} {\| {\tilde{p}}_k - q_k\|}_2}
		\label{eq_chamferloss}
	\end{aligned}	
\end{equation}
\begin{equation}
	\begin{aligned}
		\mathcal{L}_{sm,k} = \sum_{p^i_{k} \in P_k} m^P_{fg, k}\cdot{\frac{1}{N_k(p^i_k)} \sum_{p^j_{k} \in N_k(p^i_k)} {\|{\hat{S}_k}(p^j_k) - \hat{S}_k(p^i_k)\|}_1}
		\label{eq_smoothloss}
	\end{aligned}	
\end{equation}
where ${\|.\|}_1$ denotes the $L_1$-norm, and the number of neighborhood points are $ N_0=16, N_1=12, N_2=8, N_3=4$.
Both losses are then combined as follows:
\begin{equation}
	\begin{aligned}
		\mathcal{L}_{sf} = \sum_{k=0}^{3} {\alpha}_k(\mathcal{L}_{cd,k} + \mathcal{L}_{sm,k}),
		\label{eq_sfloss}
	\end{aligned}	
\end{equation}
and the weights per scale are ${\alpha}_0 = 0.02, {\alpha}_1 = 0.04, {\alpha}_2 = 0.08, {\alpha}_3 = 0.16$.

\section{Experiments} 
\label{sec:exp}
First, we give a brief description of the data sets and metrics used for evaluation.
We also demonstrate the accuracy of the method in comparison to state-of-the-art methods. 
Finally, there is a verification of our design choices.

\subsection{Evaluation Metrics} \label{Data_Metrics}
Let $\hat{S}$ denotes the predicted scene flow, and $S$ denotes the ground truth scene flow. 
The evaluation metrics for the 3D motion are averaged over all points and computed as follows:
\begin{itemize}
	\item {\textbf{EPE3D [m]}}: The 3D end-point error computed in meters as ${\|\hat{S}-S\|}_2$. 
	\item {\textbf{Acc3DS [\%]}}: The strict 3D accuracy which is the ratio of points whose EPE3D $< 0.05~m$ \textbf{or} relative error $< 5\%$. 
	\item {\textbf{Acc3DR [\%]}}: The relaxed 3D accuracy which is the ratio of points whose EPE3D $< 0.1~m$ \textbf{or} relative error $< 10\%$. 
	\item {\textbf{Out3D [\%]}}: The ratio of outliers whose EPE3D $> 0.3~m$ \textbf{or} relative error $> 10\%$.  
\end{itemize}
To evaluate the predicted ego-motion parameters (\ie, $\hat{R}$ and $\hat{t}$), compared to the ground truth ($R$ and $t$), respectively, we report the following metrics averaged over all the consecutive scenes: 
\begin{itemize}
	\item {\textbf{RAE [$\mathbf{\degree}$]}}: The relative angular error computed in degrees as: $\arccos(\mathrm{Tr}(\hat{R}^TR-1)/2)$.
	\item {\textbf{RTE [m]}}: The relative translation error computed in meters as ${\|\hat{t}-t\|}_2$.
\end{itemize}	

\subsection{Data Sets and Preprocessing} \label{sec:datasets} 
As with all related methods, the point clouds generated from the following data sets are randomly sub-sampled to be evaluated at a defined resolution (\eg, 8192 points) and are shuffled in a random order to resolve possible correlations between consecutive point clouds. 
We evaluate all of the methods in the different versions of KITTI that are described below.
All data include ground surface points. 

\textbf{semKITTI~\cite{behley2019semantickitti}}: 
It contains semantic labels of point clouds and ego-motion ground truth, including many sequences of real-world autonomous driving scenes. 
WSLR~\cite{gojcic2021weakly} has created a preprocessed version of this data set, including large sequences for training and a test split.  

\textbf{stereoKITTI~\cite{menze2015object}}: 
This is a real scene flow data set with scene flow labels.
As with most LiDAR-based methods, it is preprocessed using HPLFlowNet~\cite{gu2019hplflownet}.
This processing creates direct correlations across the consecutive scenes, and exhibits non-uniform point cloud density.

\textbf{lidarKITTI~\cite{geiger2012we}}: Unlike $\mathrm{stereoKITTI}$, the consecutive point clouds in this data set are not in direct correspondence and some points have typical occlusions.
The scene flow vectors of the ground truth are obtained by mapping the points to the corresponding pixels in the $\mathrm{stereoKITTI}$ data set.
The point clouds have a non-uniform density that mimics the sampling pattern of a typical LiDAR scan.
We use exactly the preprocessed and published data from WSLR~\cite{gojcic2021weakly}.

\subsection{Comparison to State-of-the-Art} \label{Quantitative}
To demonstrate the accuracy of our model, we compare our segmentation, ego-motion estimation, and scene flow estimation with state-of-the-art methods.

\textbf{Segmentation and Ego-Motion:}
\begin{table}[t]
	\caption{The segmentation accuracy of our \name{} generalizes better to $\mathrm{lidarKITTI}$ and shows better results for ego-motion estimation (\ie, $\mathrm{RAE}$ and $\mathrm{RTE}$) on $\mathrm{semKITTI}$.}
	\label{Table1_Comparison1} 
	\vspace{-5mm}
	\begin{center}
		\resizebox{1.0\linewidth}{!}
		{
			\begin{tabular}{c|cccccc|cccc}				
				\multirow{3}{*}{\textbf{Method}} 
				& \multicolumn{6}{c|}{$\mathbf{semKITTI}$~\textbf{\cite{behley2019semantickitti}}}
				& \multicolumn{4}{c}{$\mathbf{lidarKITTI}$~\textbf{\cite{geiger2012we}}} \\ 
				& \textbf{prec. FG~$\mathbf{\uparrow}$}  & \textbf{rec. FG~$\mathbf{\uparrow}$} & \textbf{prec. BG~$\mathbf{\uparrow}$} & \textbf{rec. BG~$\mathbf{\uparrow}$} 
				& \textbf{RAE~$\mathbf{\downarrow}$}  & \textbf{RTE~$\mathbf{\downarrow}$} 
				& \textbf{prec. FG~$\mathbf{\uparrow}$}  & \textbf{rec. FG~$\mathbf{\uparrow}$} & \textbf{prec. BG~$\mathbf{\uparrow}$} & \textbf{rec. BG~$\mathbf{\uparrow}$} 
				\\
				& [\%]               & [\%]              & [\%]               & [\%]        
				& [$\degree$]        & [m]
				& [\%]               & [\%]              & [\%]               & [\%] 
				\Tstrut\Bstrut\\
				\hline
				WSLR~\cite{gojcic2021weakly}
				& \textbf{0.950} & 0.892        & 0.991       & \textbf{0.996}  
				& 0.116          & 0.029
				& 0.734          & 0.855        & 0.991        & \textbf{0.980}               
				\Tstrut\Bstrut\\
				\textbf{Ours }   
			    & 0.898        & \textbf{0.922 }       & \textbf{0.997}       & \textbf{0.996}  
				& \textbf{0.097 }        & \textbf{0.024}
				& \textbf{0.797}         & \textbf{0.887}        & \textbf{0.992}        & 0.975              
			\end{tabular}
		}
	\end{center}
    \vspace{-4mm}
\end{table}

\begin{table}[t]
	\caption{We outperform point-wise models that are fully supervised and methods that optimize for rigid motion at the object-level~\cite{gojcic2021weakly, dong2022exploiting, chodosh2023re}.}
	\vspace{-3mm}
	\label{Table2_Comparison2}
	\begin{center}
		\resizebox{1.0\linewidth}{!}
		{
			\begin{tabular}{c|lcc|cccc|cccc}				
				\multirow{3}{*}{\textbf{Data Set}} & 
				\multirow{3}{*}{\textbf{Method}} & 
				\multirow{3}{*}{\textbf{Sup.}} & 
				\multirow{3}{*}{\textbf{Rigid.}} &
				\multicolumn{4}{c|}{$\mathbf{stereoKITTI}$~\textbf{\cite{menze2015object}}} & 
				\multicolumn{4}{c}{$\mathbf{lidarKITTI}$~\textbf{\cite{geiger2012we}}} \\ 
				& & & 
				& \textbf{EPE3D}~$\mathbf{\downarrow}$  & \textbf{Out3D}~$\mathbf{\downarrow}$ & \textbf{Acc3DS}~$\mathbf{\uparrow}$ & \textbf{Acc3DR}~$\mathbf{\uparrow}$ 
				& \textbf{EPE3D}~$\mathbf{\downarrow}$  & \textbf{Out3D}~$\mathbf{\downarrow}$ & \textbf{Acc3DS}~$\mathbf{\uparrow}$ & \textbf{Acc3DR}~$\mathbf{\uparrow}$ \\
				 & & & 
				& [m]              & [\%]              & [\%]            & [\%]              
				& [m]              & [\%]              & [\%]            & [\%] 
				\Tstrut\Bstrut\\
				\hline
				\multirow{5}{*}{$\mathbf{FT3D_{s}}$~\textbf{\cite{mayer2016large}}}		
				& PointPWC-Net~\cite{wu2020pointpwc}     & \textit{full} & \xmark
				& 0.204            & 0.645         & 0.292       & 0.556       
				& 0.710       & 0.932         & 0.114     & 0.219
				\Tstrut\Bstrut\\
				& FlowStep3D~\cite{kittenplon2021flowstep3d}     & \textit{full} & \xmark
				& 0.109            & 0.391         & 0.577       & 0.765       
				& 0.797       & 0.929         & 0.087     &  0.184
				\Tstrut\Bstrut\\
				& RMS-FlowNet~\cite{battrawy2022rms}      & \textit{full} & \xmark            
				& 0.199           & 0.547          & 0.391       & 0.618       
				& 0.652       & 0.920       & 0.120    & 0.233
				\Tstrut\Bstrut\\
				& WM3D~\cite{wang2022matters}             & \textit{full}  & \xmark 
				& 0.119       & 0.487         & 0.488      & 0.721     
				& 0.646       & 0.928         & 0.165      & 0.270          
				\Tstrut\Bstrut\\ 
				& Bi-PointFlowNet~\cite{cheng2022bi}      & \textit{full}  & \xmark
				& 0.135         & 0.439         & 0.578      & 0.760      
				& 0.686         & 0.905         & 0.179     & 0.268                         
				\Tstrut\Bstrut\\	
				\hline
				& Chodosh \etal~\cite{chodosh2023re}    & \textit{None} & \cmark
				& -             &  -            &  -              &  -           
				& 0.061         & -             & 0.917           & 0.962            
				\Tstrut\Bstrut\\
				\hline
				\multirow{3}{*}{$\mathbf{semKITTI}$~\textbf{\cite{behley2019semantickitti}}}	
				& WSLR~\cite{gojcic2021weakly}          & \textit{Weak} & \cmark
				& 0.068         & 0.263          & 0.836          & 0.897       
				& 0.080         & 0.369          & 0.742          & 0.850               
				\Tstrut\Bstrut\\
				& ERC~\cite{dong2022exploiting}           & \textit{Weak} & \cmark 
				& 0.053         & 0.269          & 0.858         & 0.917       
				& 0.065         & 0.290          & 0.857         & 0.940              
				\Tstrut\Bstrut\\
				& \textbf{Ours} & \textbf{\textit{Weak}}   & \textbf{\xmark}
				& \textbf{0.039}         & \textbf{0.212}        & \textbf{0.922}        & \textbf{0.966}            
				& \textbf{0.049}         & \textbf{0.267}        & \textbf{0.918}        & \textbf{0.964}          
				\Tstrut\Bstrut\\	
			\end{tabular}
		}
	\end{center}
    \vspace{-6mm}
\end{table}

We compare the predicted mask and ego-motion estimates of our \name{} with the pioneering work of WSLR~\cite{gojcic2021weakly}, which is the first to jointly predict binary segmentation, ego-motion, and scene flow for point clouds in a single network.
The comparison is shown in \cref{Table1_Comparison1}.
The results shown by WSLR~\cite{gojcic2021weakly} are the best optimized results presented in their paper, obtained by pre-training on FlyingThings3D \cite{mayer2016large} and subsequent further training on $\mathrm{semKITTI}$~\cite{behley2019semantickitti}.
With the exception of $FG$ precision, our model trained from scratch on the same training split of $\mathrm{semKITTI}$ outperforms WSLR on all other segmentation metrics.
Since we predict point-wise scene flow for the $FG$, errors in the segmentation have less impact compared to other methods that predict rigid object motion.
That said, our segmentation generalizes better to $\mathrm{lidarKITTI}$ and outperforms WSLR in the $FG$ category.    
In addition, our ego-motion estimation (\ie, rotation $\mathrm{RAE}$, and translation $\mathrm{RTE}$ errors ) on $\mathrm{semKITTI}$ surpasses that of WSLR~\cite{gojcic2021weakly}. 

\textbf{Scene Flow:}
The ultimate goal of our model is to predict the scene flow for each input point in the scene with respect to point cloud $P$.
To this end, we evaluate our final scene flow estimate against point-wise methods \cite{wu2020pointpwc, kittenplon2021flowstep3d, battrawy2022rms, wang2022matters, cheng2022bi}, which are state-of-the-art methods that perform best on $\mathrm{stereoKITTI}$ when ground points are omitted.
However, the accuracy of these methods is severely limited in the presence of ground points on $\mathrm{stereoKITTI}$ and even worse on $\mathrm{lidarKITTI}$, a data set that resembles real LiDAR scenes with occlusions and no direct correspondences between successive LiDAR scans.
Our point-based scene flow is comparable to the latest conventional method~\cite{chodosh2023re} on $\mathrm{lidarKITTI}$, which integrates the ego-motion and rigidity assumptions into the scene flow estimation, but our method performs significantly better with respect to EPE3D.
We also outperform the object-based weakly supervised methods WSLR~\cite{gojcic2021weakly} and ERC~\cite{dong2022exploiting} on both KITTI versions in all metrics (\cf \cref{Table2_Comparison2}).
We also visualize our qualitative results on $\mathrm{lidarKITTI}$ in \Cref{Figure10_QualitativeComparison}. 
Further qualitative results can be found in the supplementary material.

\textbf{Efficiency:}
Our model contains about 16 million parameters. For $8192$ input points, it requires 11 GFLOPs, which takes an average of $140ms$ for a pair of point clouds on a single NVIDIA Titan V.
\begin{figure}[t]
	\centering
	\begin{tabular}{p{0.05cm}cccc}
		& \textit{Example 1} & \textit{Example 2} & \textit{Example 3} 
		\Tstrut\Bstrut\\ 
		\rotatebox[origin=c]{90}{\textit{Scene}} &
		\includegraphics[width=0.29\linewidth,valign=c]{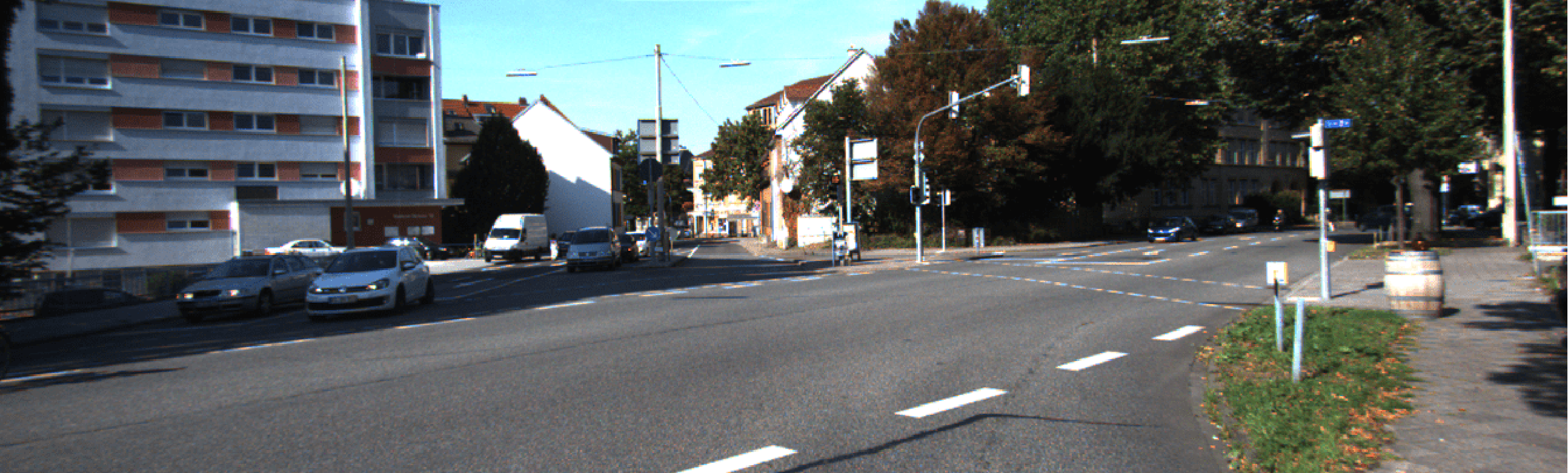}& 
		\includegraphics[width=0.29\linewidth,valign=c]{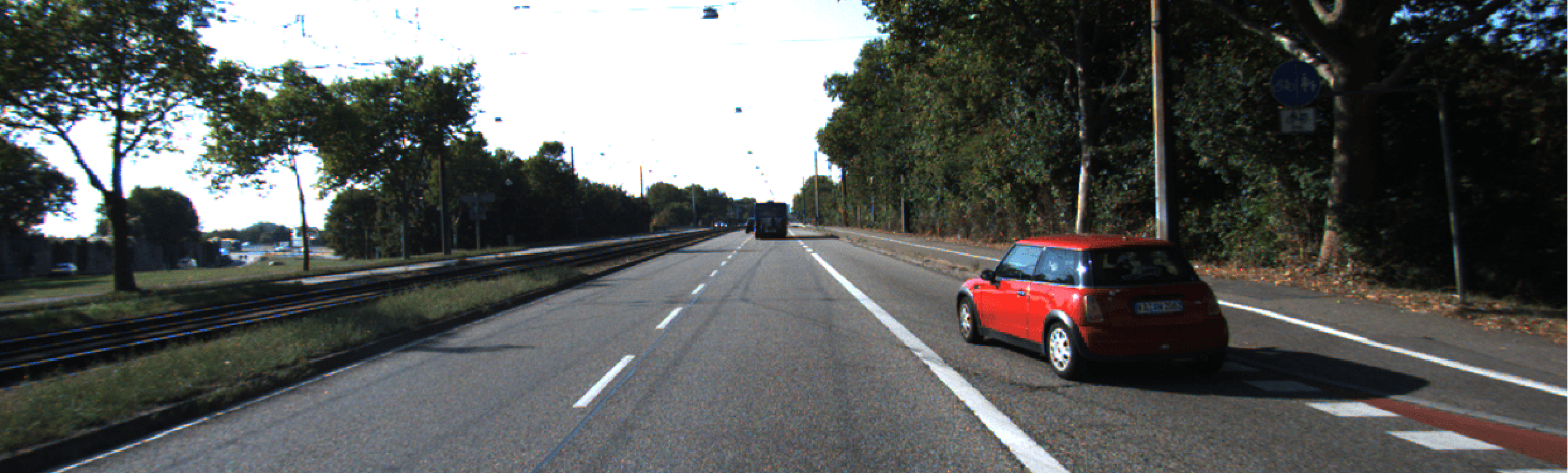}&
		\includegraphics[width=0.29\linewidth,valign=c]{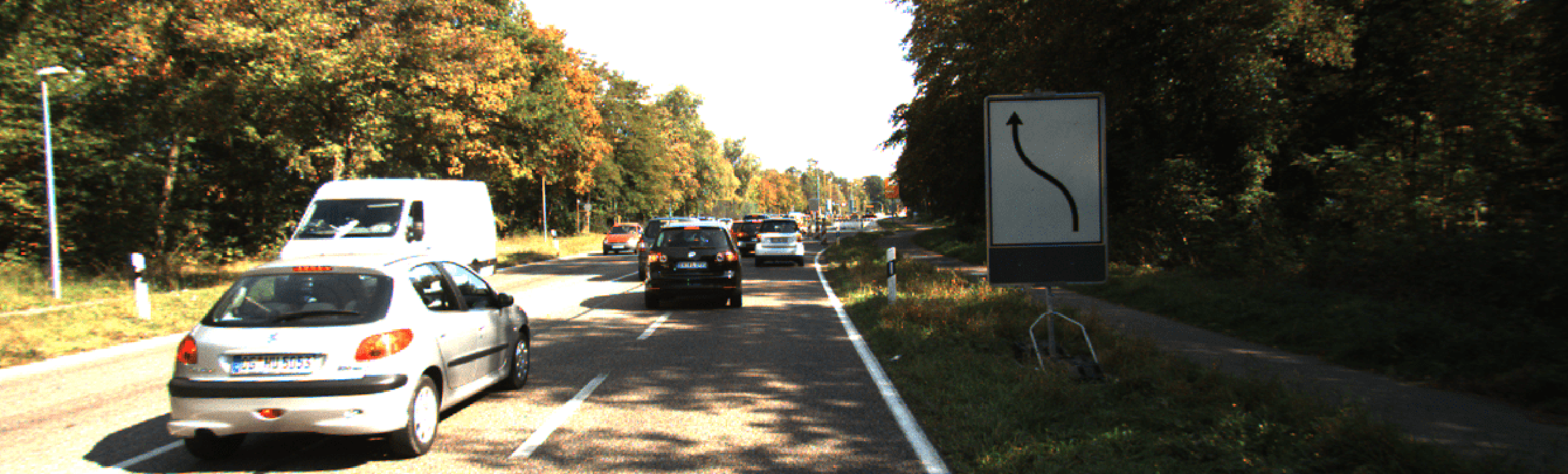}
		\Tstrut\Bstrut\\		
		\rotatebox[origin=c]{90}{\textit{$M^P_{fg}$}} &
		\includegraphics[width=0.29\linewidth,valign=c]{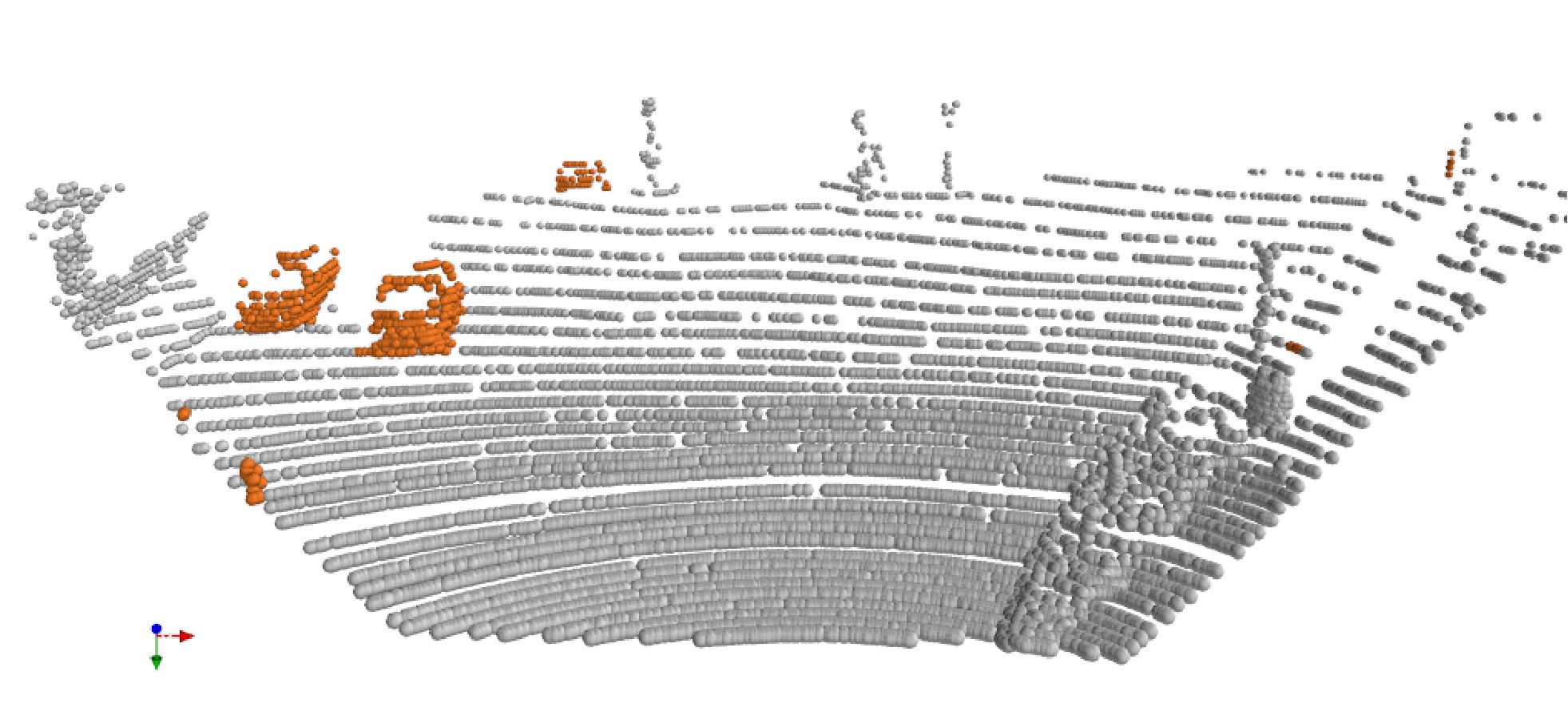}&
		\includegraphics[width=0.29\linewidth,valign=c]{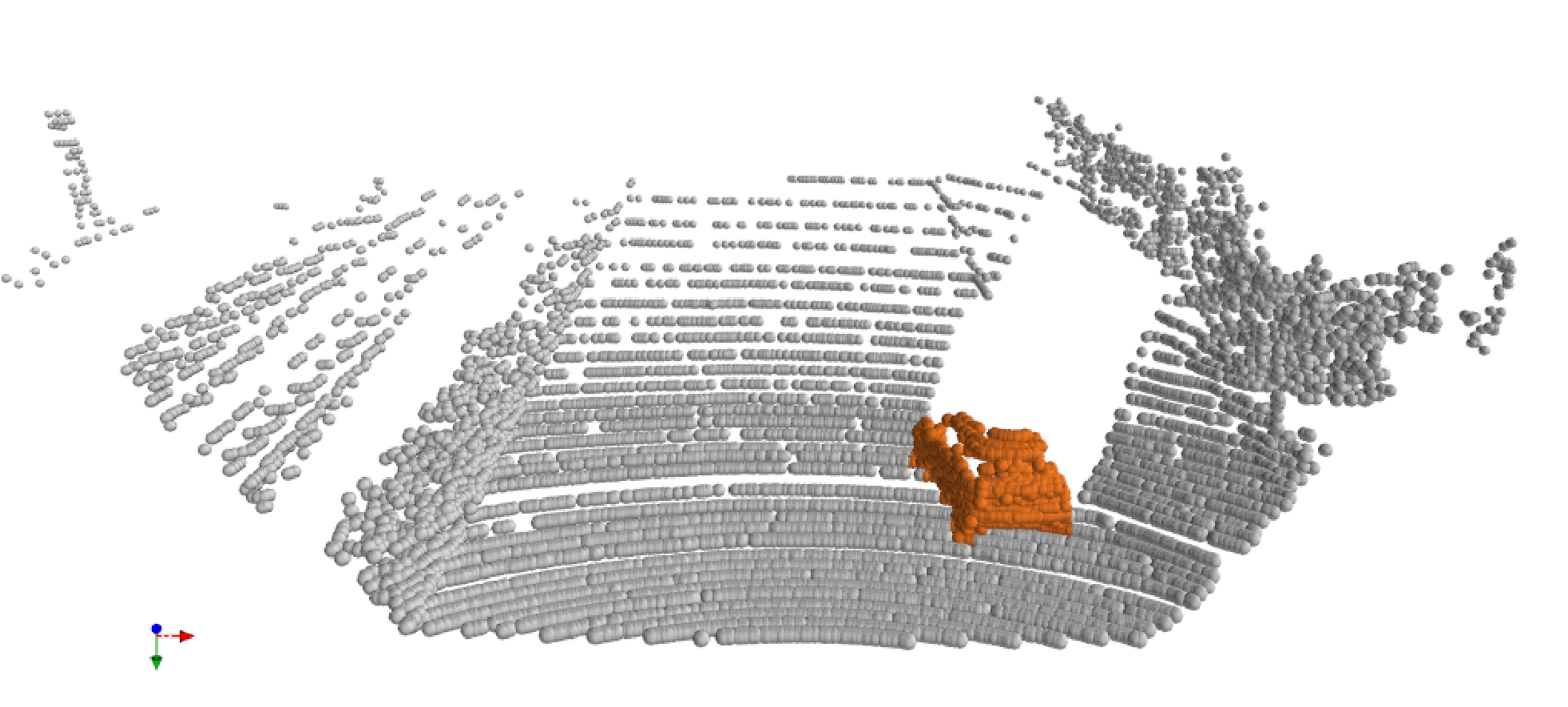}&
		\includegraphics[width=0.29\linewidth,valign=c]{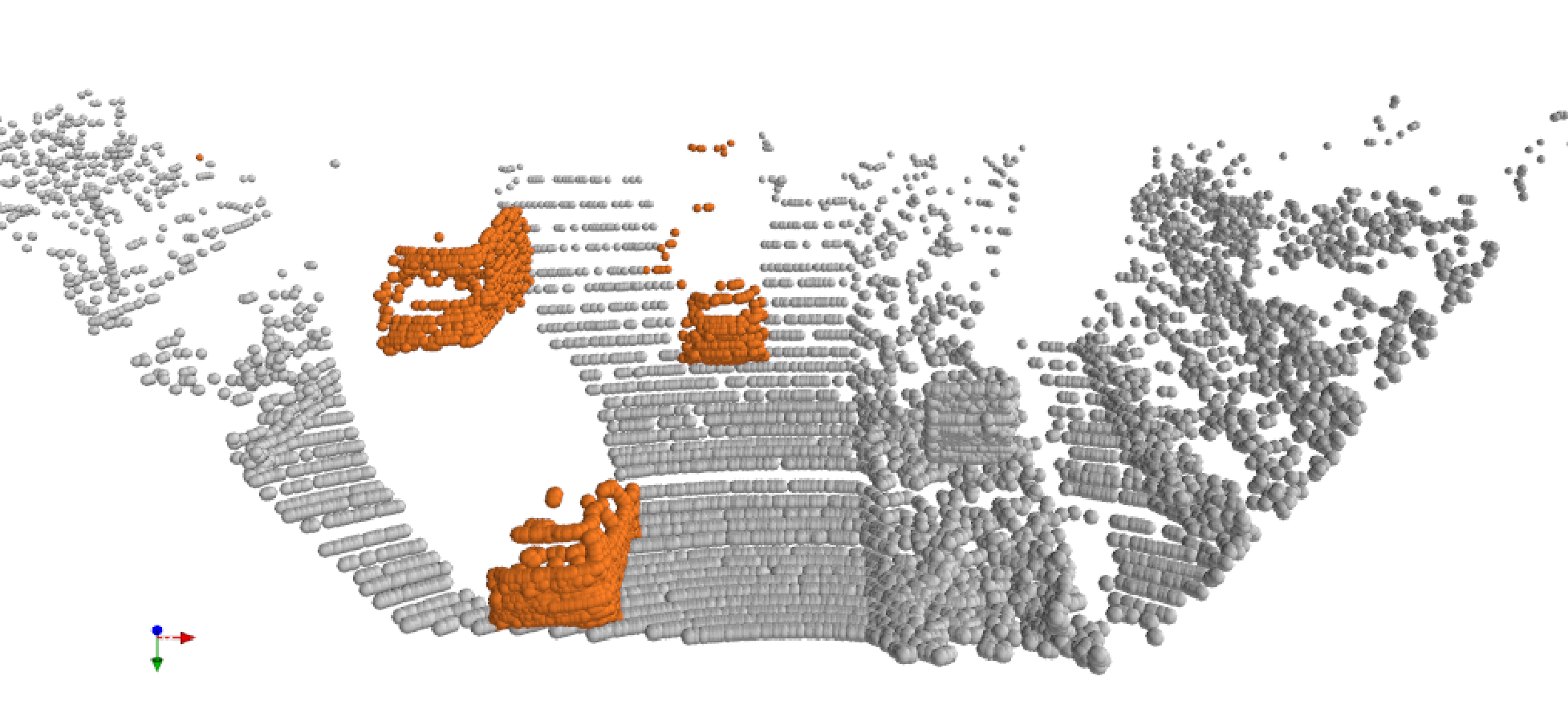}
		\Tstrut\Bstrut\\
		
		\rotatebox[origin=c]{90}{\textit{Error Map}} &
		\includegraphics[width=0.29\linewidth,valign=c]{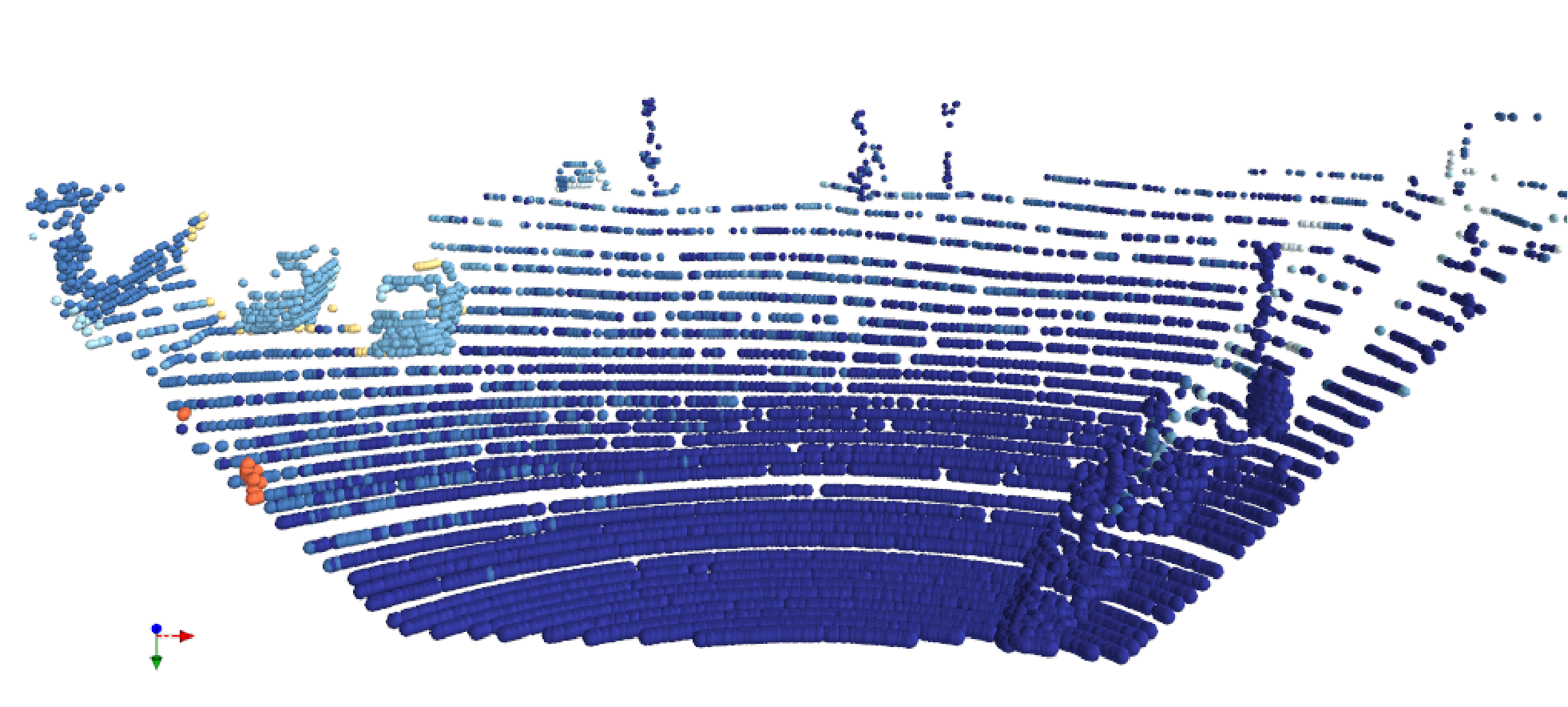}&
		\includegraphics[width=0.29\linewidth,valign=c]{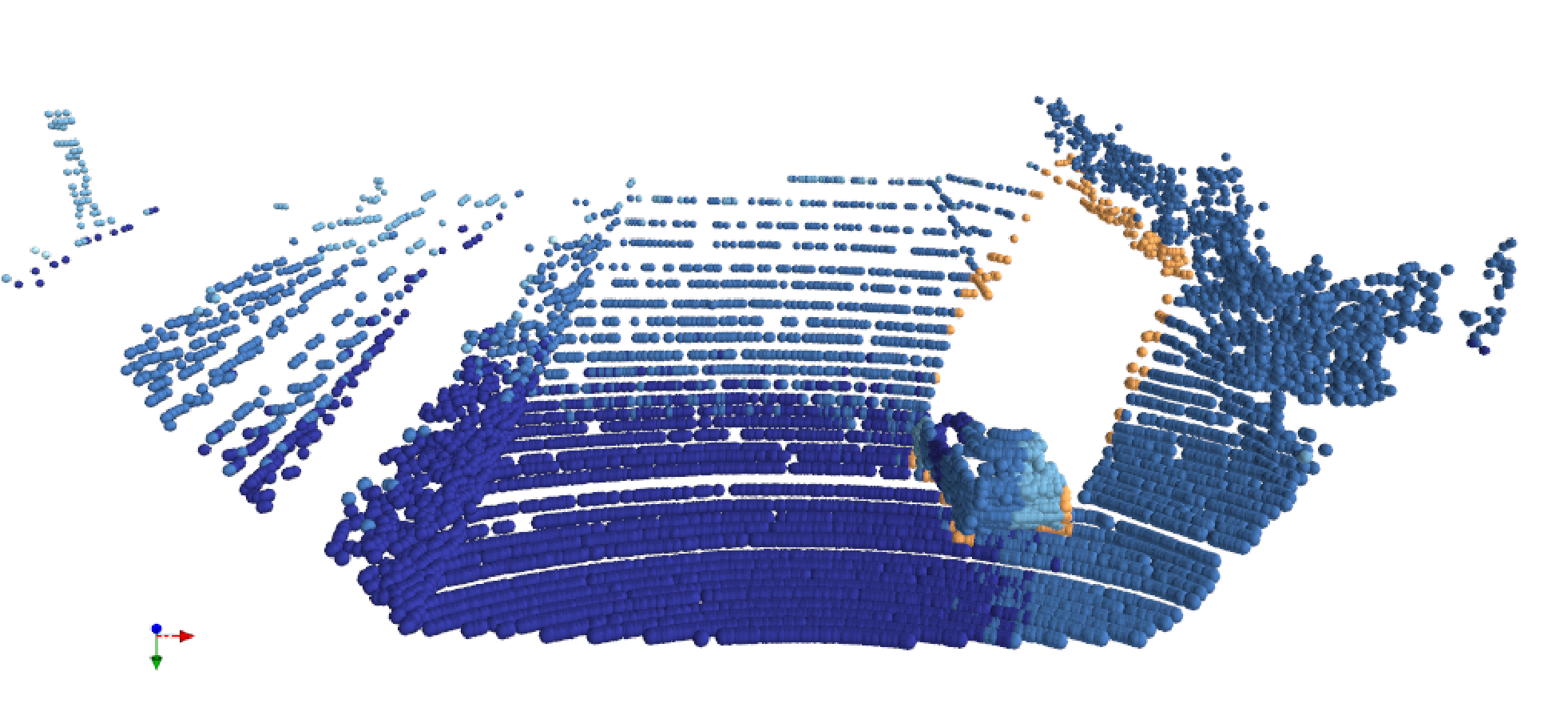}&
		\includegraphics[width=0.29\linewidth,valign=c]{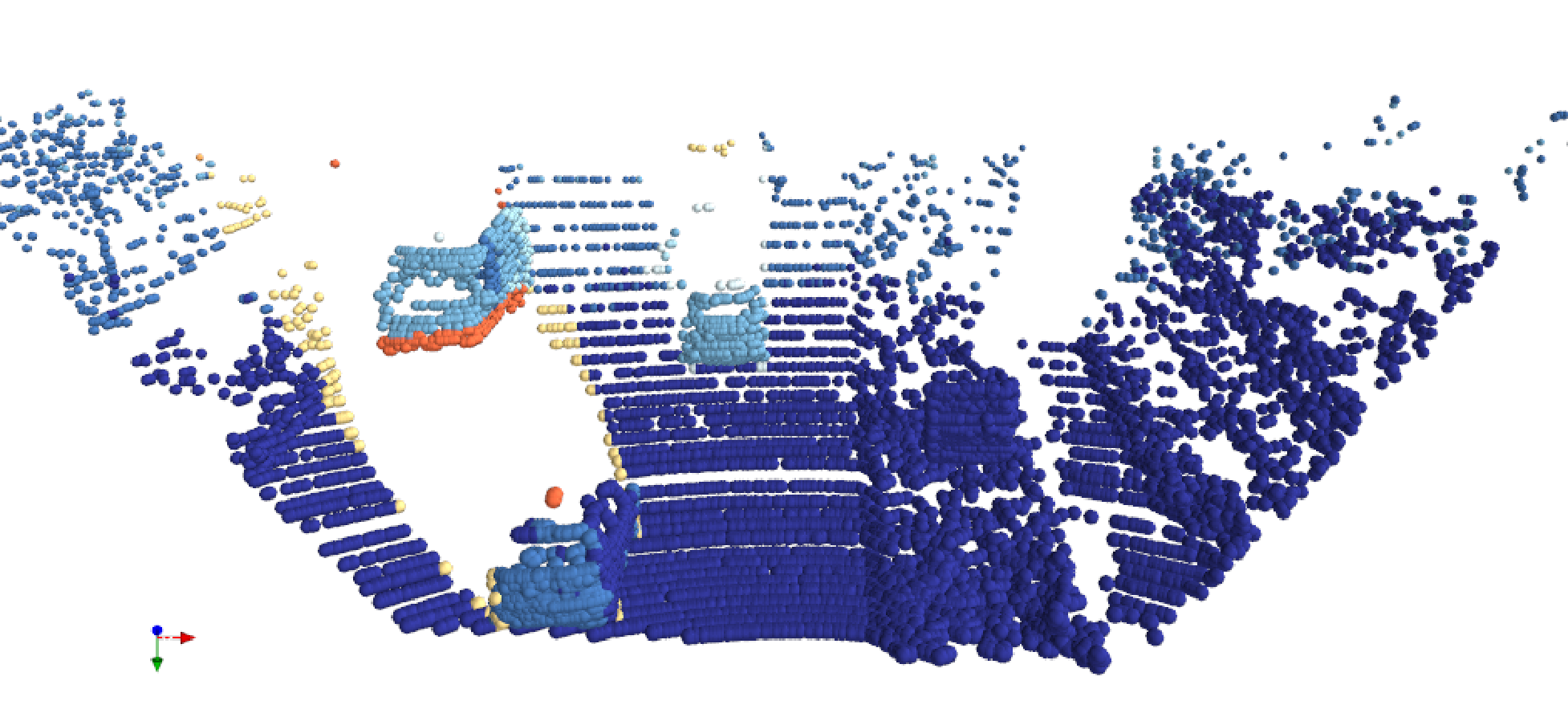}
		\Tstrut\Bstrut\\	
		& \multicolumn{3}{c}{\includegraphics[width=0.9\linewidth,valign=c]{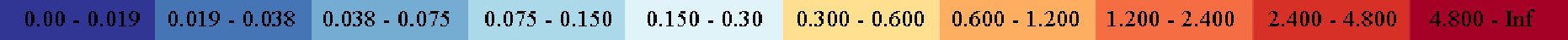}}\Tstrut\Bstrut\\
	\end{tabular}
	\caption{Three examples from $\mathrm{lidarKITTI}$~\cite{geiger2012we} show the qualitative results of our \name{}. For visual enhancement only, we show the RGB images of each scene. We visualize the predicted binary mask, where $BG$ and $FG$ points are encoded by gray and orange colors, respectively. The error map for each scene (third row) shows the end-point error in meters and is colored according to the map shown in the last row. Our \name{} shows low errors (dark blue) over a wide area in each scene, including $FG$ and $BG$ points.}
	\label{Figure10_QualitativeComparison}
	\vspace{-5mm}
\end{figure}

\subsection{Ablation Study} \label{Ablation}
We conduct several experiments by training on $\mathrm{semKITTI}$ and evaluating on $\mathrm{lidarKITTI}$ to verify our design decisions.
To do this, we evaluate $FG$ and $BG$ points separately in $\mathrm{EPE3D_{fg}}$ and $\mathrm{EPE3D_{bg}}$, respectively.
We build our baseline design by computing our features from the encoder of the first feature extraction network ($F_{encoder}$), so that no context encoder module and hybrid features are applied, and we do not consider binary masks for any of our network branches (\ie, ego-motion and scene flow) and our warping layer is computed based on the scene flow branch only.
Then, we check the basic estimates of our model by applying only the proposed losses (\cf \cref{eq_totalloss}), as shown in the $1^{st}$ row in \cref{Table3_Ablation_Components}. 
In the $2^{nd}$ row, we see the positive influence of integrating the predicted background mask $\mathrm{M^P_{bg}}$ into the ego-motion branch. 
In the $3^{rd}$ row, we apply our hybrid warping layer as in \cref{eq_warping}, which further improves the results.
Adding scene flow feature updating and dual attention refinement significantly improves the results for all metrics, as can be clearly seen in rows $4^{th}$ and $5^{th}$.
Without the \textit{stop gradient} in \cref{eq_features}, the results are the same as after using refinements, but applying it shows an improvement in $\mathrm{EPE3D_{fg}}$ and consequently in all other metrics.
We provide further experiments in the supplementary material.
\begin{table}[b]
	\caption{We investigate the contribution of each component in our design on $\mathrm{lidarKITTI}$~\cite{geiger2012we}.}
	\vspace{-6mm}
	\label{Table3_Ablation_Components}
	\begin{center}
		\resizebox{1.0\linewidth}{!}
		{
			\begin{tabular}{cccccc|cccc}				
				 $\mathbf{M^P_{bg}}$ & \textbf{Hyprid}    & \textbf{Feature}   & \textbf{Attention}   & \textbf{Hybrid}   & \textbf{Hybrid}  
				 & $\mathbf{EPE3D}~\downarrow$  & $\mathbf{EPE3D_{fg}}~\downarrow$ & $\mathbf{EPE3D_{bg}}~\downarrow$ & \textbf{Acc3DR~$\uparrow$} 
				 \\         
				                     & \textbf{Warping}   & \textbf{Update}    & \textbf{Refinement}  & \textbf{Features} & \textbf{Features~$\perp{}$}
				& [m]                & [m]                & [m]                 & [\%] 
				\Tstrut\Bstrut\\	
				\hline	
				\xmark                   & \xmark             & \xmark             & \xmark           & \xmark            & \xmark                    
				& 0.168                  & 0.375              & 0.154              & 0.689              
				\Tstrut\Bstrut\\  
				\cmark                   & \xmark             & \xmark             & \xmark           & \xmark            & \xmark         
                & 0.139                  & 0.363              & 0.119              & 0.808    
                \Tstrut\Bstrut\\ 
                \cmark                   & \cmark             & \xmark             & \xmark           & \xmark            & \xmark    
                & 0.085                  & 0.335              & 0.065              & 0.871      
                \Tstrut\Bstrut\\ 
                \cmark                   & \cmark             & \cmark             & \xmark            & \xmark           & \xmark    
                & 0.067                  & 0.326              & 0.048              & 0.931    
                \Tstrut\Bstrut\\
                \cmark                   & \cmark             & \cmark             & \cmark            & \xmark          & \xmark         
                & 0.053                  & 0.287              & 0.035              & 0.952   
                \Tstrut\Bstrut\\
                \cmark                   & \cmark             & \cmark             & \cmark            & \cmark           & \xmark       
                & 0.053                  & 0.282              & 0.036              & 0.957      
                \Tstrut\Bstrut\\
                \cmark                   & \cmark             & \cmark             & \cmark            & \cmark         & \cmark       
                & \textbf{0.049}         & \textbf{0.267}     & \textbf{0.033}     & \textbf{0.964}      
                \\
			\end{tabular}
		}
	\end{center}
    \vspace{-6mm}
\end{table}

\section{Conclusion}
We propose \name{}, which predicts binary segmentation masks for dynamic and static LiDAR-based scenes and jointly estimates hierarchical ego-motion and scene flow. 
Our method works by estimating scene flow at the point-level rather than optimizing it at the object level. 
Our network is free of any clustering and uses point-level refinement, which produces better results than competing methods and allows for non-rigid object motions. 
Our approach outperforms recent approaches that rely on the object-level and shows robust accuracy in the presence of ground points.

\subsection*{Acknowledgment}
This work was partially funded by the Federal Ministry of Education and Research Germany under the project DECODE (01IW21001) and partially in the funding program Photonics Research Germany under the project FUMOS (13N16302).
\newpage
\renewcommand\thesection{\Roman{section}}
\renewcommand{\thetable}{\Roman{table}}
\renewcommand{\thefigure}{\Roman{figure}}
\section*{\centering SUPPLEMENTARY}

In our supplementary material, we explain details of implementation, training and augmentation and we perform further ablation studies to validate our design choices. 
We then add another comparison with a newer method that works under the assumption of rigidity.
Finally, we discuss the possible shortcomings of our approach and show more qualitative results.
\setcounter{section}{0}
\setcounter{table}{0}
\setcounter{figure}{0}
\section{Implementation, Training and Augmentation} \label{sub:Implementation}
Following related approaches~\cite{gojcic2021weakly, dong2022exploiting}, we train our method by considering all frames of the train split of semKITTI \cite{behley2019semantickitti}.
During training, the preprocessed data is randomly sub-sampled to a certain resolution (\ie, $8192$ points), where the order of the points is random and the correlation between consecutive frames is resolved by random selection.
We use the Adam optimizer with default parameters and train our model for $150$ epochs. 
We use an exponentially decaying learning rate, initialized at $0.001$ and then decaying at a rate of $0.7$ every $10$ epochs. 
We apply batch normalization to all layers of our model except the last layer in each head (\ie, segmentation, scene flow, and the layer providing confidence values in the ego-motion branch).
We perform geometric augmentation, which is a random rotation of all points around one randomly chosen axis by a random degree uniformly selected between $-10\degree$ and $+10\degree$. 
Our entire architecture is implemented using TensorFlow.
	
\section{More Experiments}
\subsection{Additional Ablation Studies}
\textbf{Verification of Losses:}
We conduct further experiments to verify our losses. The results are shown in \cref{Table:Ablation_Losses}.
Supervision for all points by the basic self-supervised loss for scene flow (marked with \cmark(*) in the \cref{Table:Ablation_Losses}) and without the losses of segmentation $\mathcal{L}_{seg}$ and ego-motion $\mathcal{L}_{ego}$ results in extremely inaccurate scene flow. 
However, integrating both the additional losses significantly improves the scene flow in all metrics.  
Adding the binary masks to our self-supervised loss, as suggested in the paper, improves the scene flow over $FG$ and $BG$ points even further, as shown in the last row.
\begin{table}[b]
	\caption{We explore the impact of our losses. For this experiment, we train on $\mathrm{semanticKITTI}$~\cite{behley2019semantickitti} and test on $\mathrm{lidarKITTI}$~\cite{geiger2012we} in the presence of ground surface points. The marker (*) indicates that the self-supervised loss of scene flow is applied to all points without considering the segmentation masks.}
	\label{Table:Ablation_Losses}
	\begin{center}
		\begin{tabular}{ccc|cccc}				
			$\mathcal{L}_{sf}$ & $\mathcal{L}_{seg}$ & $\mathcal{L}_{ego}$ & $\mathbf{EPE3D_{all}}$ & $\mathbf{EPE3D_{fg}}$ & $\mathbf{EPE3D_{bg}}$ & \textbf{Acc3DR} \\
			&            &         & [m]                    & [m]                   & [m]                  & [\%] 
			\Tstrut\Bstrut\\	
			\hline	\	
			\cmark(*)            & \xmark     & \xmark       &   0.509             & 0.485                  & 0.501                 & 0.193        \Tstrut\Bstrut\\
			\cmark(*)            & \cmark    & \cmark          &   0.071             & 0.380                  & 0.049                 & 0.920       \Tstrut\Bstrut\\	
			\cmark             & \cmark       & \cmark         & \textbf{0.049}         & \textbf{0.267}     & \textbf{0.033}     & \textbf{0.964}        \Tstrut\Bstrut\\			
		\end{tabular}
	\end{center}
\end{table}

\begin{table}[t]
	\caption{The use of hybrid features with the stop gradient in our \name{} almost matches the results of the task-specific segmentation network and provides the most accurate results for the ego-motion.}
	\label{Table:Hybrid_Features} 
	\vspace{-5mm}
	\begin{center}
		\resizebox{1.0\linewidth}{!}
		{
			\begin{tabular}{c|cccc|cccccc}				
				\multirow{3}{*}{\textbf{Task}} 
				& \multirow{3}{*}{$\mathbf{F_{s,0}}$} & \multirow{3}{*}{$\mathbf{F_{encoder}}$} & \multirow{3}{*}{$\mathbf{HF}$} & \multirow{3}{*}{$\mathbf{HF\perp{}}$}
				& \multicolumn{6}{c}{$\mathbf{lidarKITTI}$~\textbf{\cite{geiger2012we}}} \\ 
				& & & &	
				& \textbf{prec. FG~$\mathbf{\uparrow}$}  & \textbf{rec. FG~$\mathbf{\uparrow}$} & \textbf{prec. BG~$\mathbf{\uparrow}$} & \textbf{rec. BG~$\mathbf{\uparrow}$} 
				& \textbf{RAE~$\mathbf{\downarrow}$}  & \textbf{RTE~$\mathbf{\downarrow}$} 
				\\ 
				& & & &     	
				& [\%]               & [\%]              & [\%]               & [\%] 
				& [$\degree$]        & [m]
				\Tstrut\Bstrut\\
				\hline
				seg. & \cmark & \xmark & \xmark & \xmark
				& \textbf{0.8058}         & \textbf{0.8895}        & \textbf{0.9920}        & \textbf{0.9800}  
				& -               & -           
				\Tstrut\Bstrut\\
				seg. + ego. & \cmark & \cmark & \xmark & \xmark	
				& 0.7083         & 0.8869        & 0.9918        & 0.9691  
				& 0.1143         & \ul{0.0389} 
				\Tstrut\Bstrut\\
				seg. + ego. + sf. & \cmark & \cmark & \xmark & \xmark			
				& 0.7207   & 0.8865     &  0.9913      & 0.9716  
				& \ul{0.1046}   & 0.0398
				\Tstrut\Bstrut\\         
				seg. + ego. + sf. & \cmark & \xmark & \cmark & \xmark	
				& 0.7133   & 0.8800            &  0.9916      & 0.9702 
				& 0.1128   & 0.0422
				\Tstrut\Bstrut\\ 
				\textbf{seg. + ego. + sf.} & \cmark & \xmark & \xmark & \cmark	
				& \ul{0.7958}   & \ul{0.8872}  &  \ul{0.9917}      & \ul{0.9784}
				& \textbf{0.0943}   & \textbf{0.0293}
				
			\end{tabular}
		}
	\end{center}
\end{table}

\textbf{Impact of Hybrid Features with Stop Gradient:}
We verify our decision to develop hybrid features $\mathrm{HF\perp{}}$ with stop gradient by evaluating the segmentation and ego-motion on the $\mathrm{lidarKITTI}$ data set~\cite{geiger2012we} in the presence of the ground surface points in \cref{Table:Hybrid_Features}.	

First, we verify the accuracy of our segmentation without the ego-motion and scene flow branches by training the segmentation task using only the features extracted by the decoder module $F_{s,0}$.
Then, we add the ego-motion branch without scene flow, but using the features from the encoder module of the first feature extraction network $F_{encoder}$. 
The precision of the segmentation at $FG$ points is negatively affected by the addition of the ego-motion branch. 
The addition of the scene flow branch slightly improves the segmentation precision at $FG$ points, and the addition of the context encoder using the hybrid features without stop gradients still shows poor precision at $FG$ points.
With the stop gradient $\mathrm{\perp{}}$, we improve the overall accuracy of the segmentation almost to the results of the specific-segmentation task $1^{st}$ row and we also improve the relative angular error $\mathrm{RAE}$ and the relative translational error $\mathrm{RTE}$.

\subsection{Additional Comparison} \label{sub:no_ground_test}
We compare with the very recent scene flow estimation method, RSF~\cite{deng2023rsf}, which jointly optimizes a global ego-motion and a set of bounding boxes with their own rigid motions, without using any annotated labels.  
The RSF~\cite{deng2023rsf} approach provides a robust scene flow and outperforms most of the recent scene flow approaches when the ground surface is excluded.
However, reliable exclusion of the ground surface is not always possible, may lead to an incomplete representation of the scene.
Therefore, we compare our \name{} with RSF once with excluded ground points, and again when they are present.
The comparison is presented in \cref{Table:no_ground}. 
We consider the default settings of RSF~\cite{deng2023rsf}\footnote{\url{https://github.com/davezdeng8/rsf}} for the evaluation.
For the test without ground points, we feed our network with all points including the ground points, but we evaluate all remaining points after removing the ground points.
The presence of ground points affects the overall accuracy of the RSF~\cite{deng2023rsf} method while our approach still shows a comparable result to RSF~\cite{deng2023rsf} when we evaluate without ground points. 

In terms of efficiency, RSF~\cite{deng2023rsf} takes more than $30$ seconds for each point cloud pair, while our \name{} takes $140ms$ on the same NVIDIA Titan V GPU.
\begin{table}[t]
	\caption{In comparison to RSF \cite{deng2023rsf}, our method shows a consistently high accuracy, independent of the data set or the whether the ground surface is included or excluded.}
	\label{Table:no_ground}
	\begin{center}
		\resizebox{1.0\linewidth}{!}
		{
			\begin{tabular}{c|lcc|cccc|cccc}				
				& 
				\multirow{3}{*}{\textbf{Method}} & 
				\multirow{3}{*}{\textbf{Sup.}} & 
				\multirow{3}{*}{\textbf{Rigid.}} &
				\multicolumn{4}{c|}{$\mathbf{stereoKITTI}$~\textbf{\cite{menze2015object}}} & 
				\multicolumn{4}{c}{$\mathbf{lidarKITTI}$~\textbf{\cite{geiger2012we}}} \\ 
				& & & 
				& \textbf{EPE3D}~$\mathbf{\downarrow}$  & \textbf{Out3D}~$\mathbf{\downarrow}$ & \textbf{Acc3DS}~$\mathbf{\uparrow}$ & \textbf{Acc3DR}~$\mathbf{\uparrow}$ 
				& \textbf{EPE3D}~$\mathbf{\downarrow}$  & \textbf{Out3D}~$\mathbf{\downarrow}$ & \textbf{Acc3DS}~$\mathbf{\uparrow}$ & \textbf{Acc3DR}~$\mathbf{\uparrow}$ \\
				& & & 
				& [m]              & [\%]              & [\%]            & [\%]              
				& [m]              & [\%]              & [\%]            & [\%] 
				\Tstrut\Bstrut\\
				
				\hline
				\textbf{without}
				& RSF~\cite{deng2023rsf}  & \textit{None} & \cmark
				& \textbf{0.035}         & \textbf{0.146}          & \textbf{0.932}          & \textbf{0.971}
				& 0.085         & \textbf{0.239}         & \textbf{0.883}          & 0.929
				\Tstrut\Bstrut\\
				\textbf{ground} & \textbf{Ours} & \textbf{\textit{Weak}}   & \textbf{\xmark}
				& 0.042         & 0.190         & 0.874          & 0.969             
				& \textbf{0.069}       & 0.257        & 0.857        & \textbf{0.932}          
				\Tstrut\Bstrut\\
				\hline
				\textbf{with} 
				& RSF~\cite{deng2023rsf}  & \textit{None} & \cmark
				& 0.205         & 0.387         & 0.735          & 0.802
				& 0.416         & 0.767         & 0.308         & 0.498
				\Tstrut\Bstrut\\
				\textbf{ground} & \textbf{Ours} & \textbf{\textit{Weak}}   & \textbf{\xmark}
				& \textbf{0.039}         & \textbf{0.212}        & \textbf{0.922}        & \textbf{0.966}            
				& \textbf{0.049}         & \textbf{0.267}        & \textbf{0.918}        & \textbf{0.964}  
				
			\end{tabular}
		}
	\end{center}
\end{table}

\subsection{Limitations} \label{sub:limitations}
\begin{figure}[t]
	\centering
	\begin{tabular}{p{0.05cm}cccc}
		& \textit{Example 1} & \textit{Example 2} & \textit{Example 3} 
		\Tstrut\Bstrut\\ 
		
		\rotatebox[origin=c]{90}{\textit{Scene}} &
		\includegraphics[width=0.29\linewidth,valign=c]{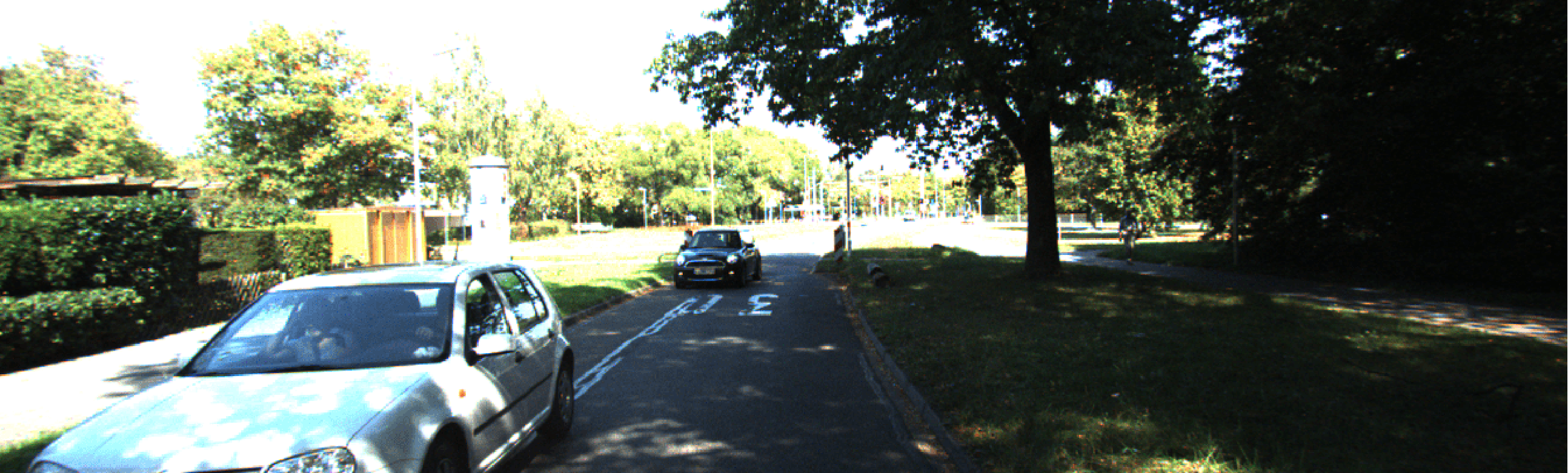}&
		\includegraphics[width=0.29\linewidth,valign=c]{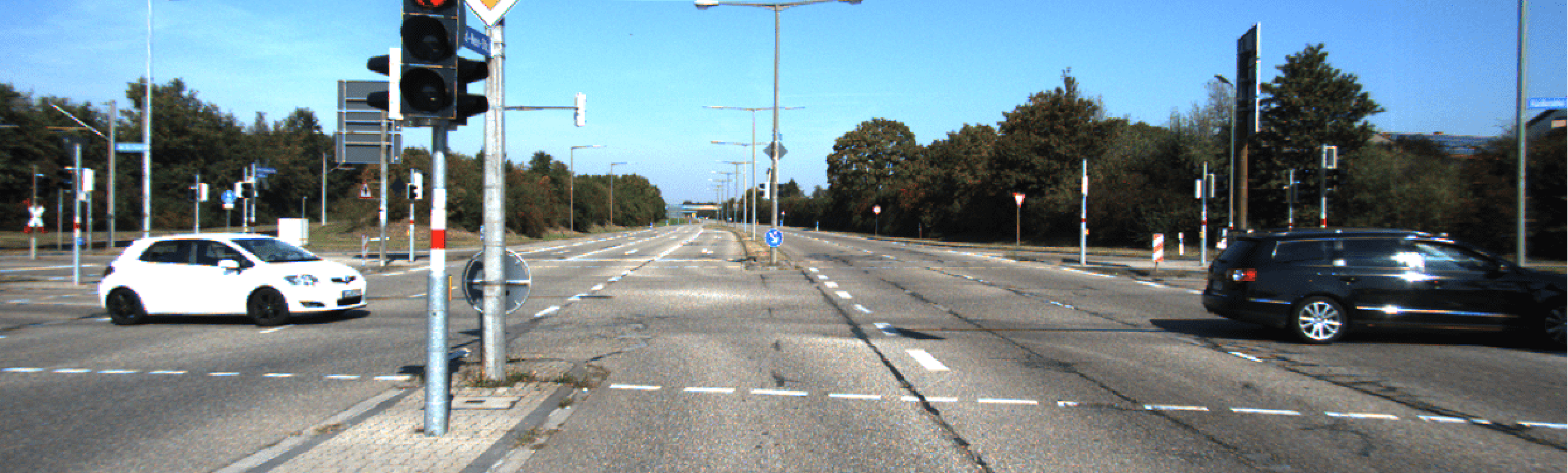}& 
		\includegraphics[width=0.29\linewidth,valign=c]{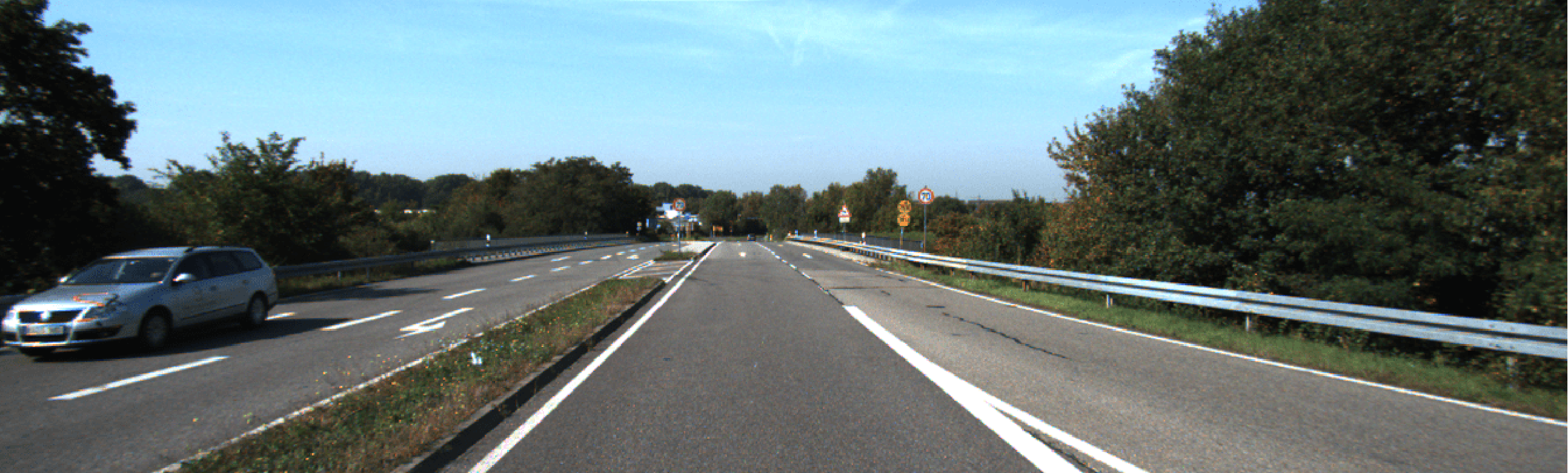}  
		\Tstrut\Bstrut\\	
			
		\rotatebox[origin=c]{90}{\textit{$M^P_{fg}$}} &
		\includegraphics[width=0.29\linewidth,valign=c]{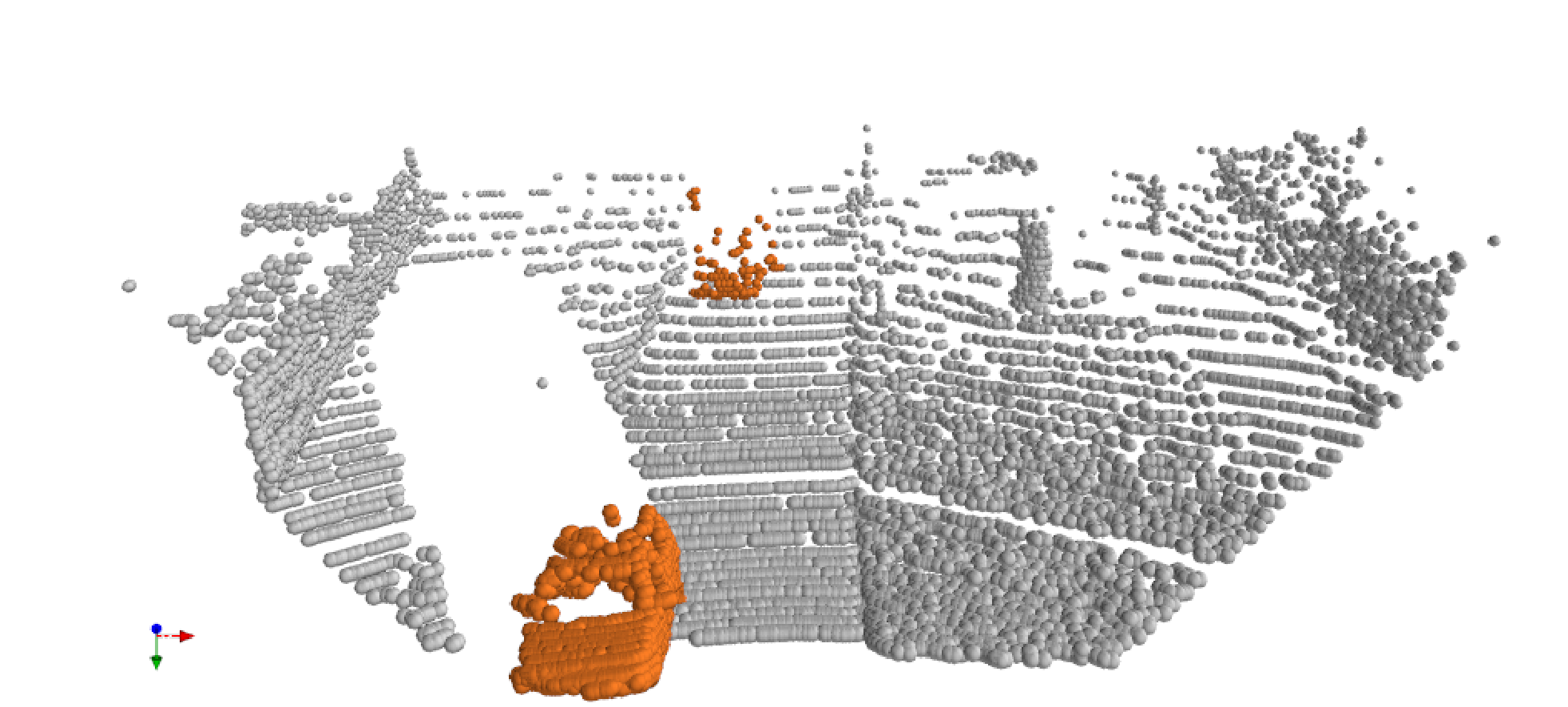}&
		\includegraphics[width=0.29\linewidth,valign=c]{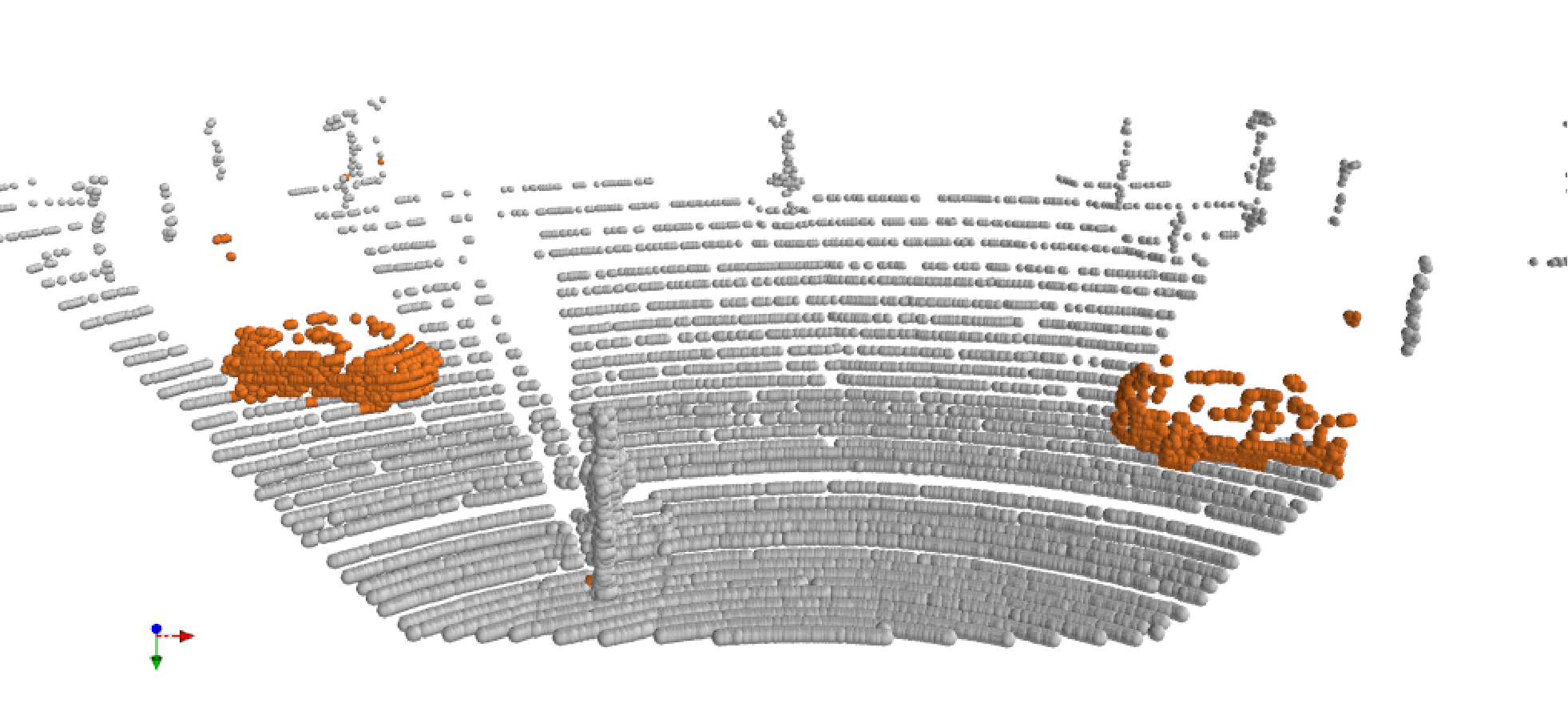}&
		\includegraphics[width=0.29\linewidth,valign=c]{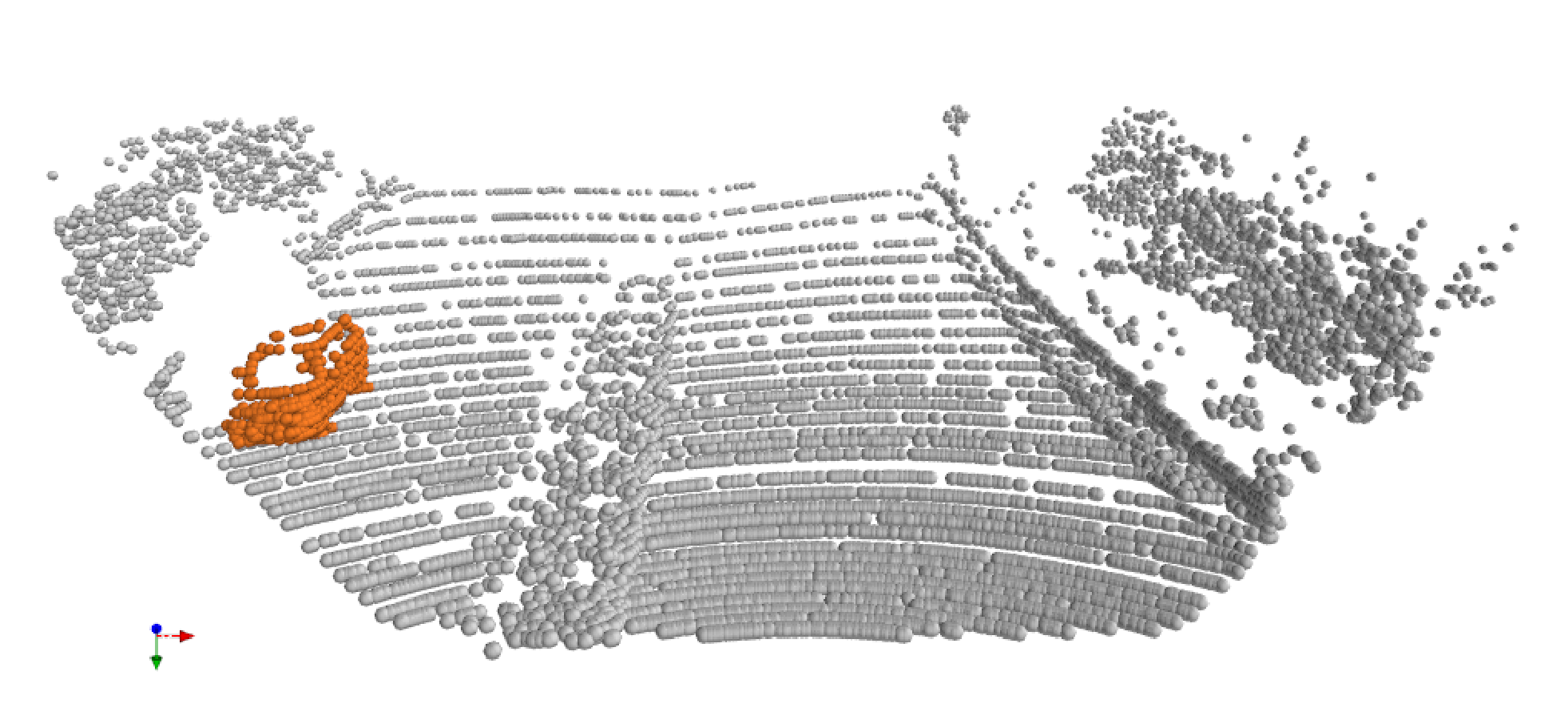}
		\Tstrut\Bstrut\\
		
		\rotatebox[origin=c]{90}{\textit{$M^Q_{fg}$}} &
		\includegraphics[width=0.29\linewidth,valign=c]{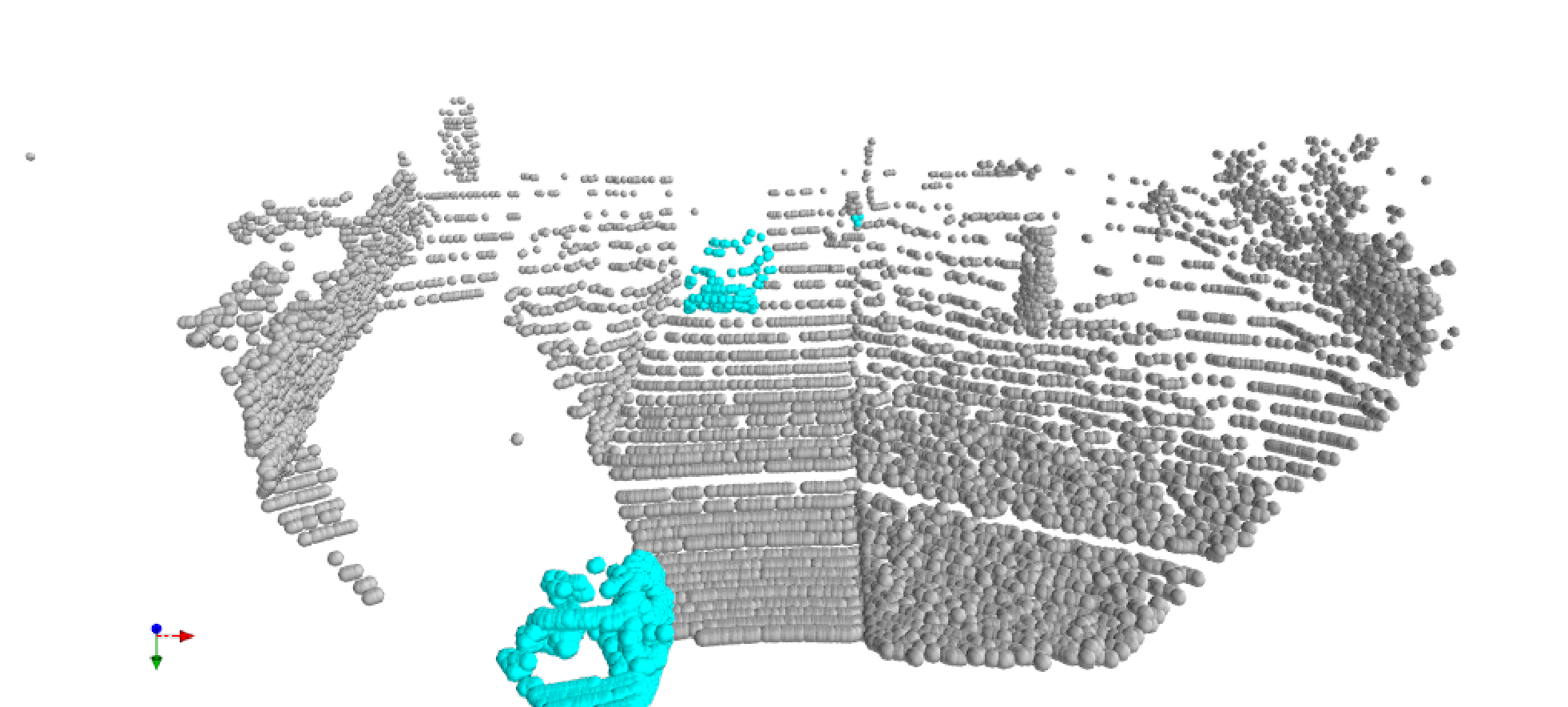}&
		\includegraphics[width=0.29\linewidth,valign=c]{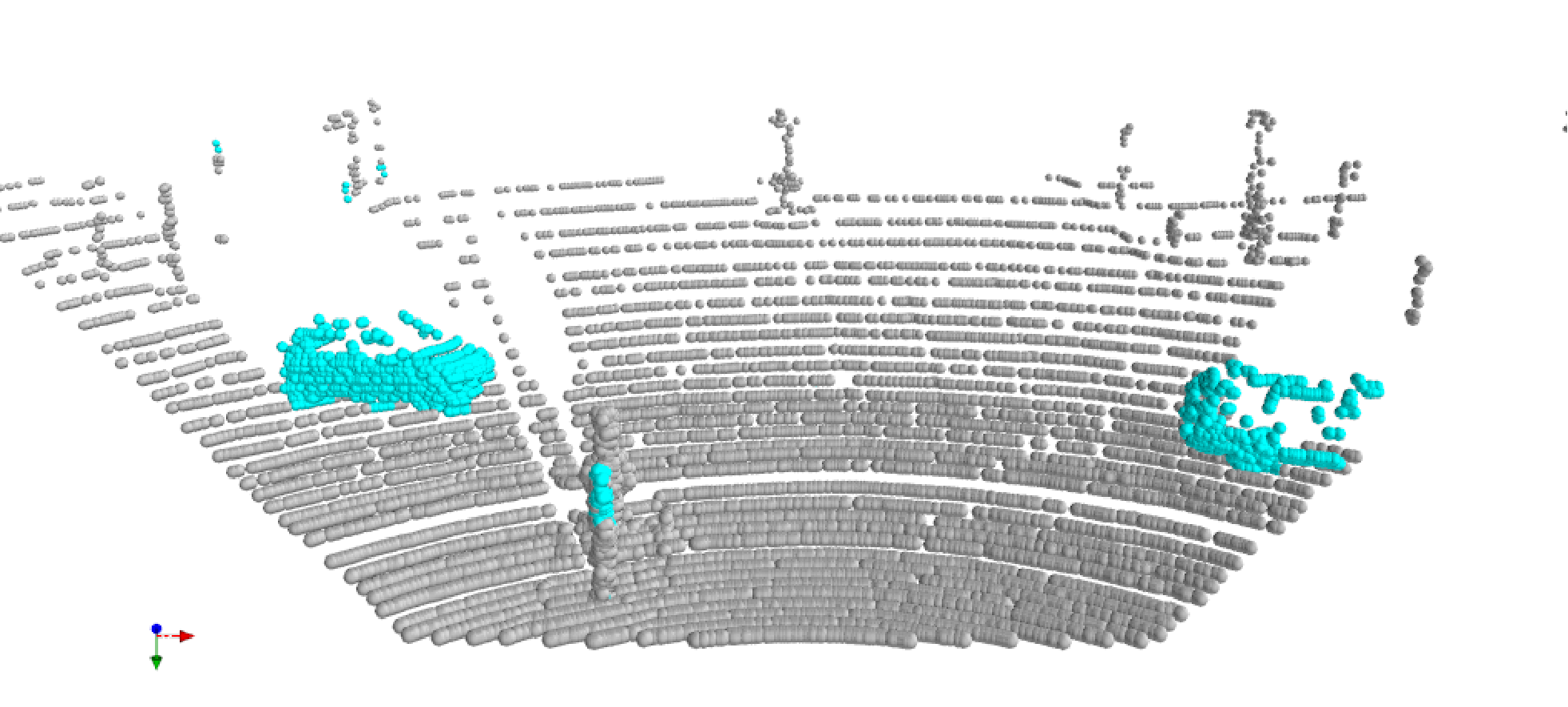}&
		\includegraphics[width=0.29\linewidth,valign=c]{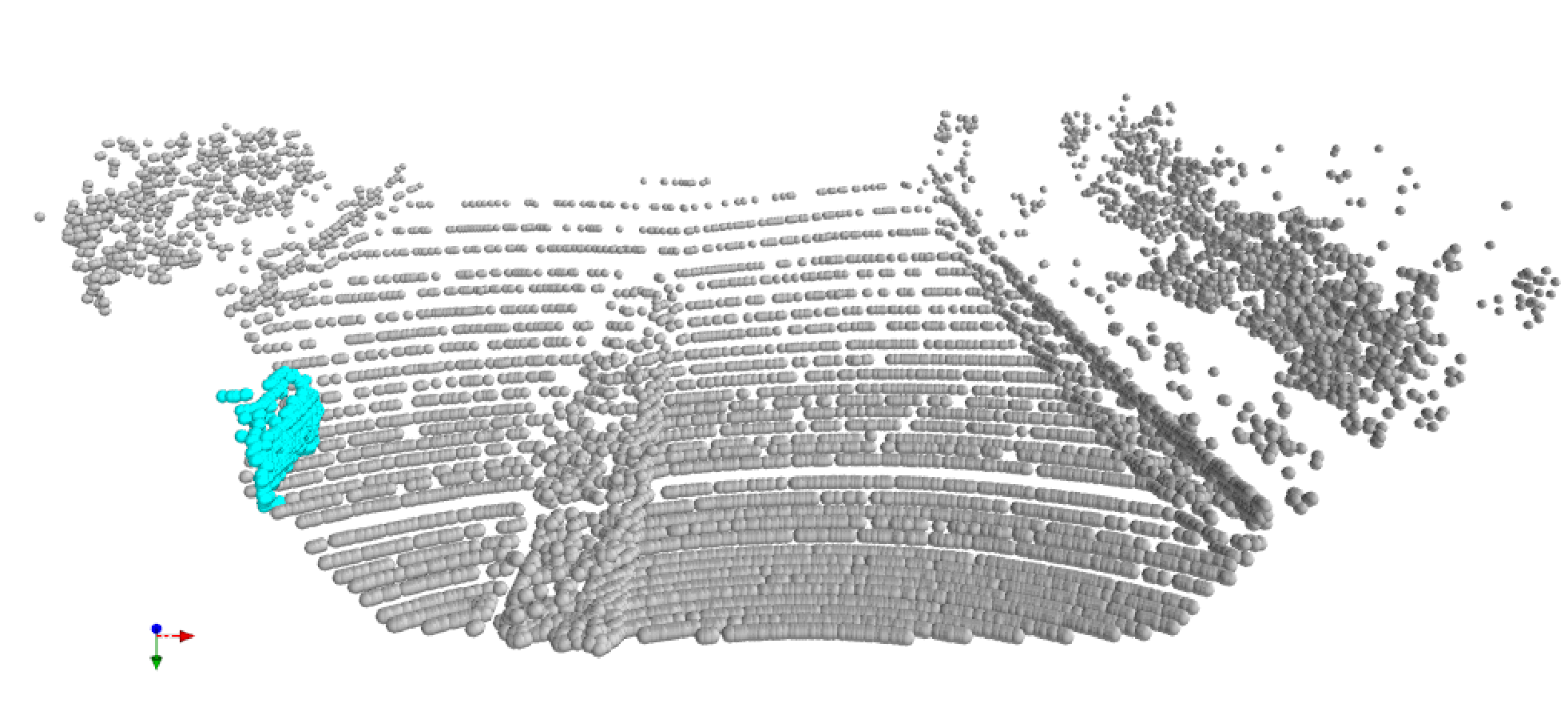}
		\Tstrut\Bstrut\\
		
		\rotatebox[origin=c]{90}{\textit{Error Map}} &
		\includegraphics[width=0.29\linewidth,valign=c]{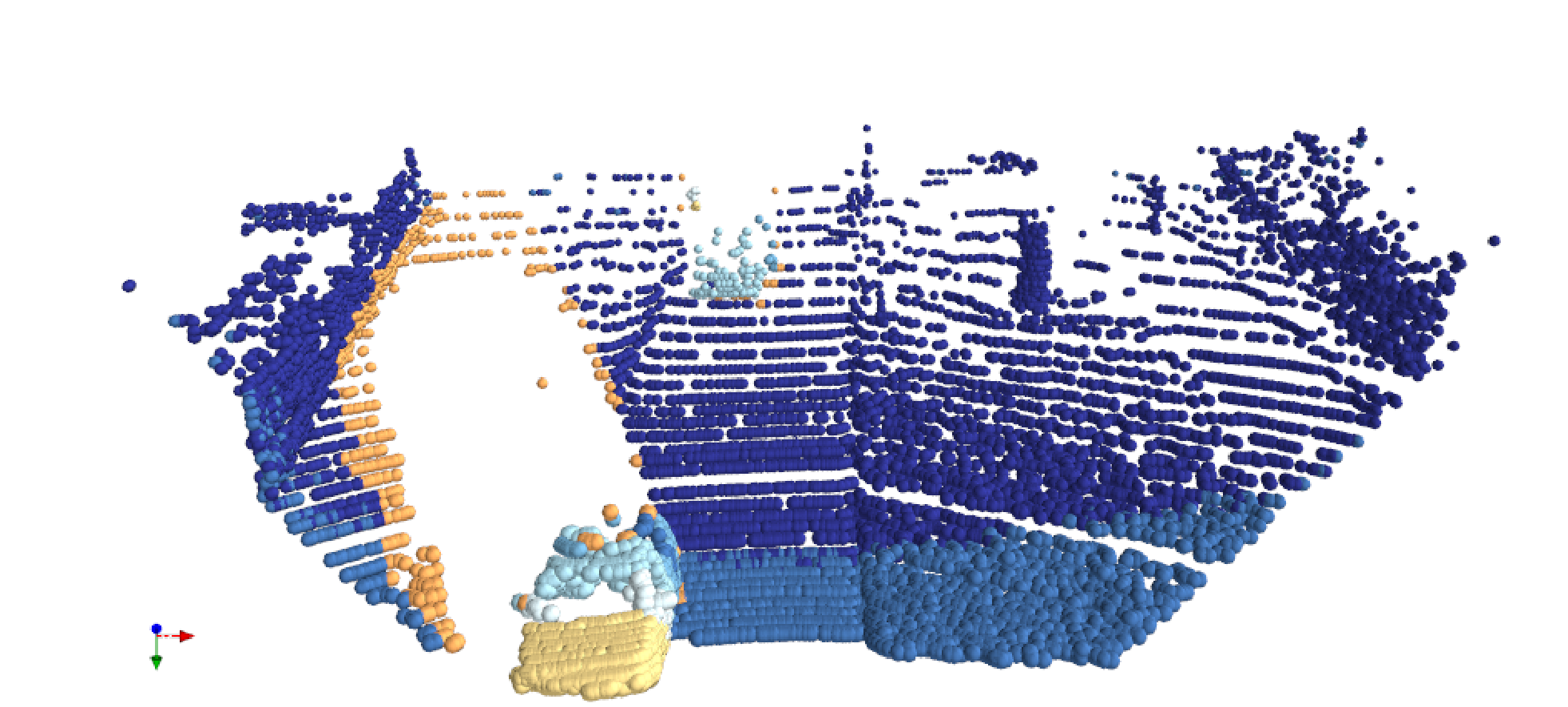}&
		\includegraphics[width=0.29\linewidth,valign=c]{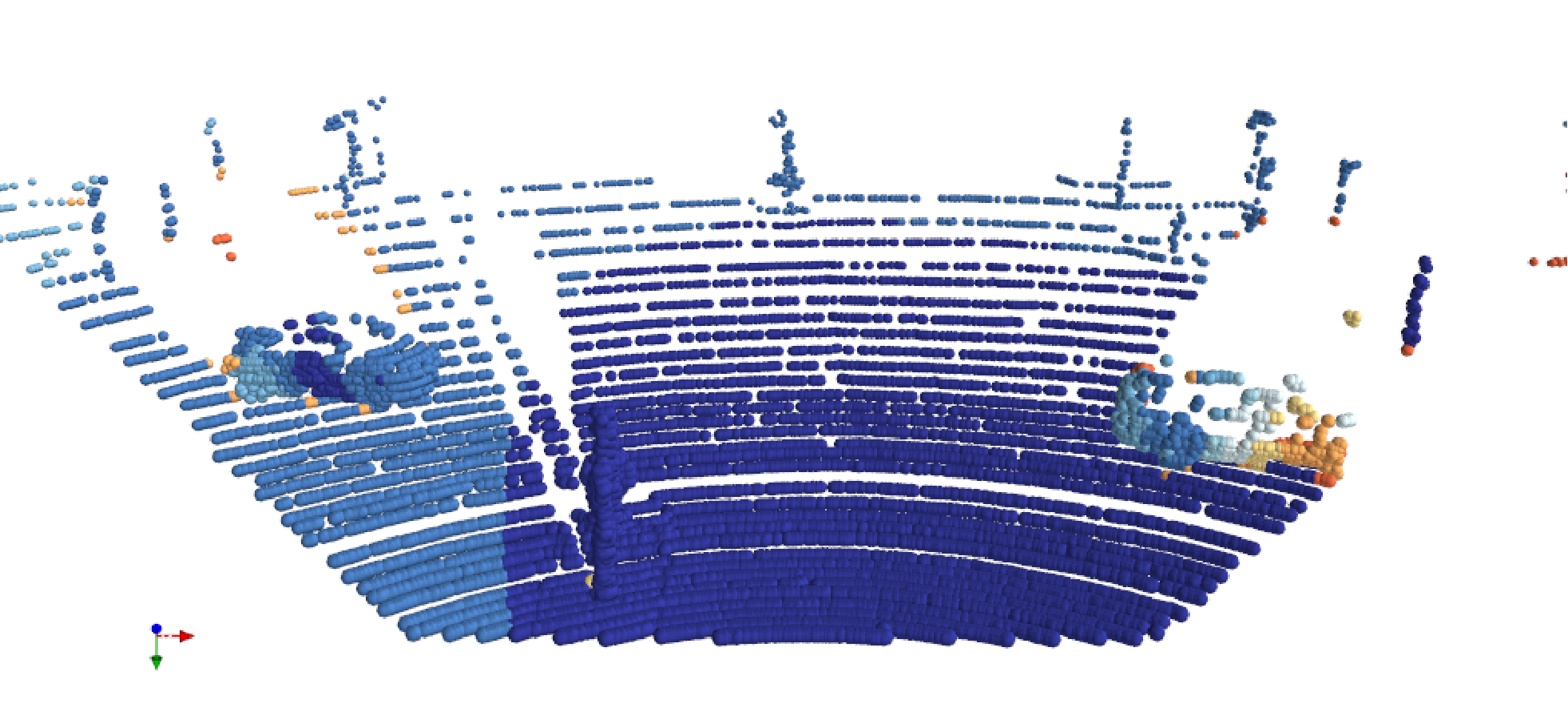}&
		\includegraphics[width=0.29\linewidth,valign=c]{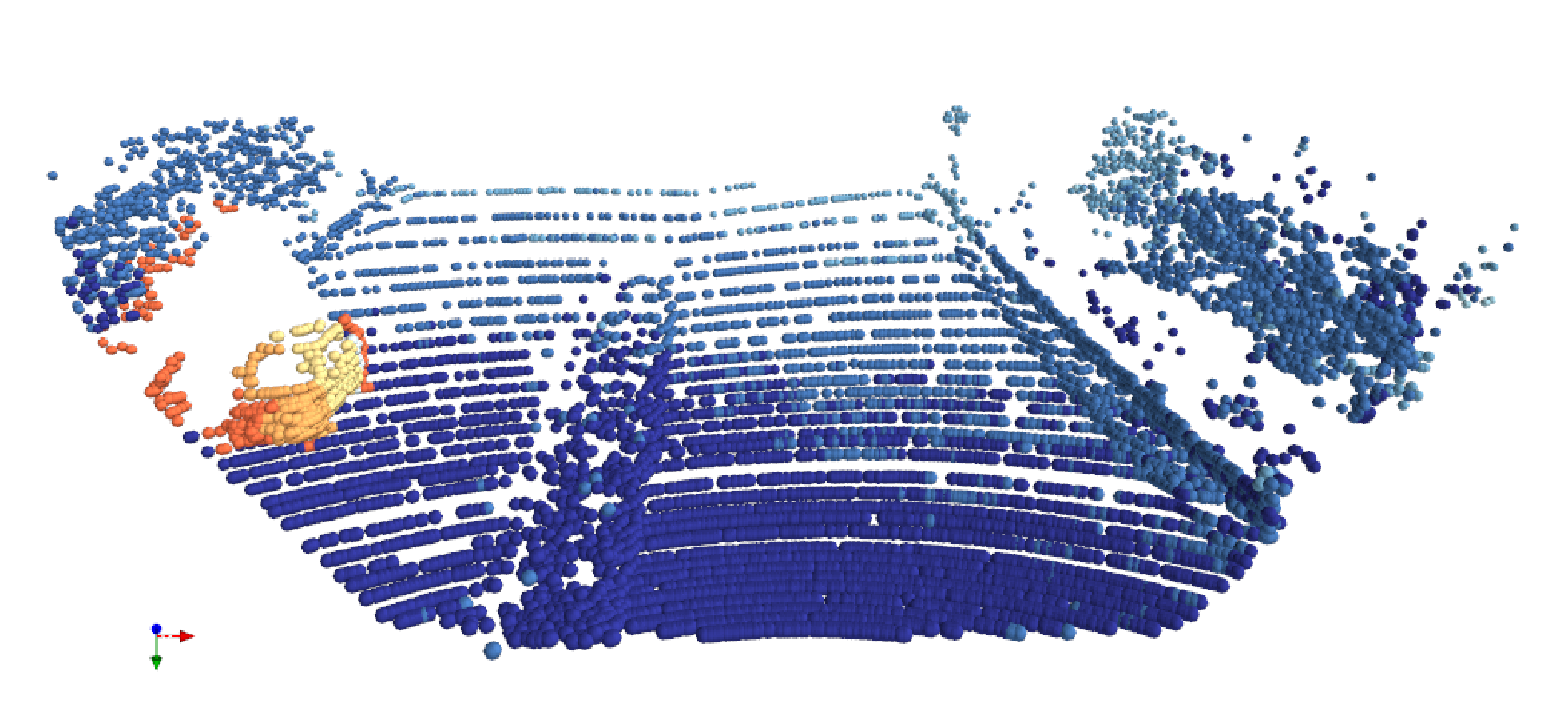}
		\Tstrut\Bstrut\\	
		
		& \multicolumn{3}{c}{\includegraphics[width=0.9\linewidth,valign=c]{eps_min/KITTI_errorcolors_3D-min.eps}}\Tstrut\Bstrut\\
	\end{tabular}
	\caption{Three examples from $\mathrm{lidarKITTI}$~\cite{geiger2012we} show the cases where cars are not fully sensed in the second frame $Q$ and our scene flow prediction partially fails. For visual enhancement only, we show the RGB images of each scene. We visualize the predicted binary mask, where $BG$ and $FG$ points are encoded by gray and orange or cyan colors, respectively. The error map for each scene (third row) shows the end-point error in meters and is colored according to the map shown in the last row.}
	\label{Figure:qual_failures}
\end{figure}

In terms of accuracy, we find that our \name{} can fail for moving objects that leave the field of view, so that they are partially occluded or disappear in the second LiDAR frame $Q$. 
In this case, the scene flow prediction for these areas is often partially or completely wrong.
We illustrate such cases in \Cref{Figure:qual_failures}. 

Adding robustness against occlusions remains a challenge for future work.

\subsection{Additional Qualitative Results} \label{sub:qual_results}
\begin{figure}[t]
	\centering
	\begin{tabular}{p{0.05cm}cccc}
		& \textit{Example 1} & \textit{Example 2} & \textit{Example 3} 
		\Tstrut\Bstrut\\ 
		
		\rotatebox[origin=c]{90}{\textit{Scene}} &
		\includegraphics[width=0.29\linewidth,valign=c]{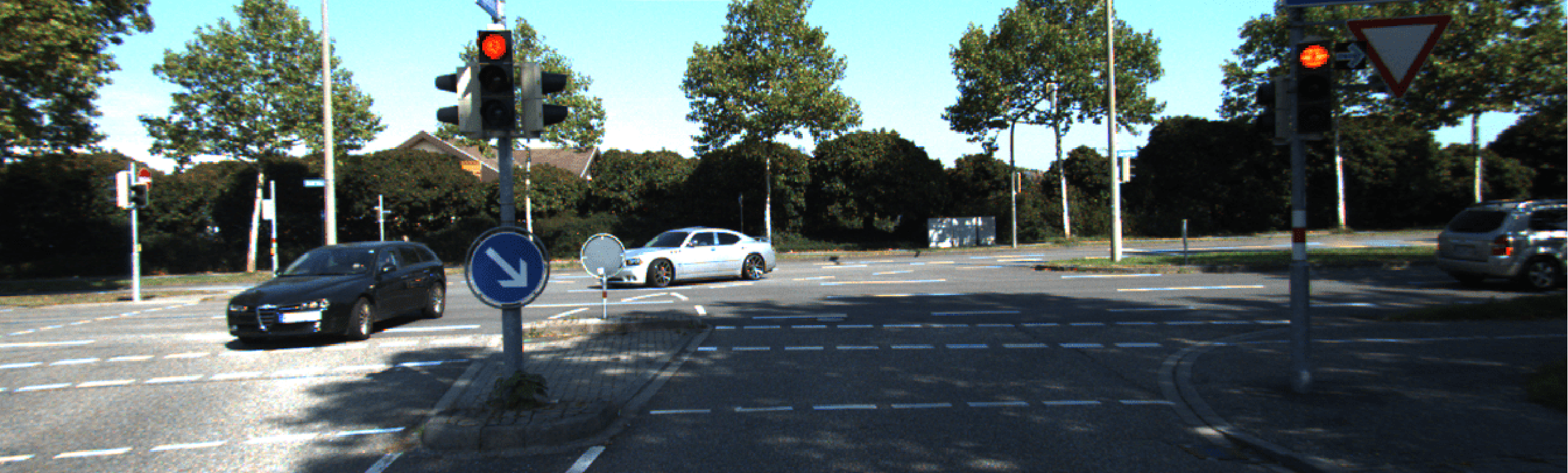}& 
		\includegraphics[width=0.29\linewidth,valign=c]{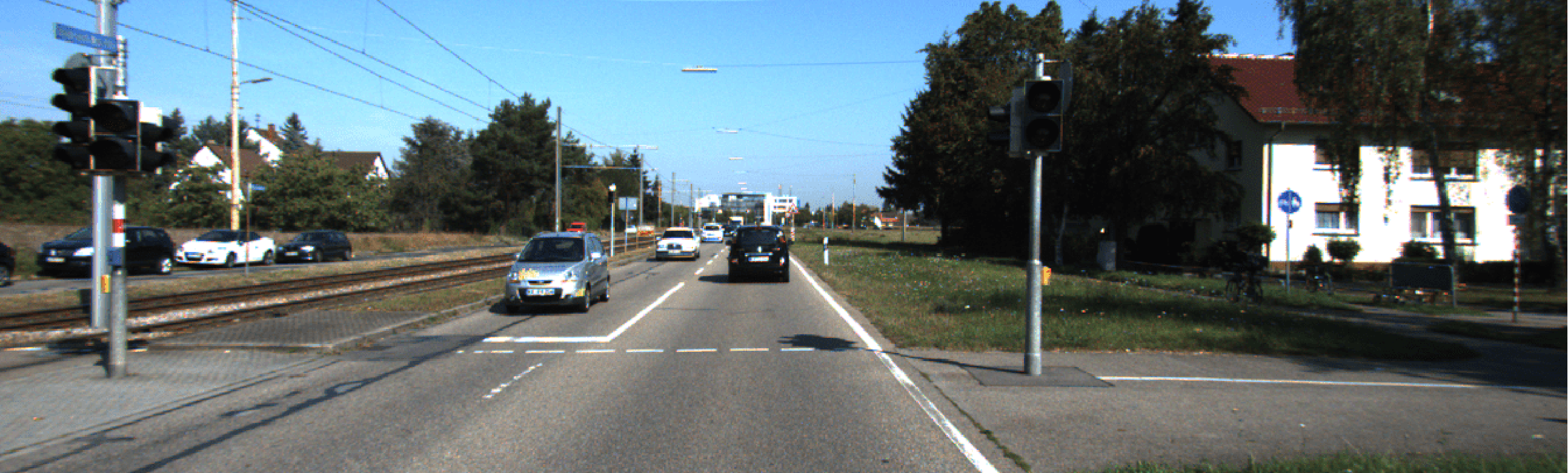}& 
		\includegraphics[width=0.29\linewidth,valign=c]{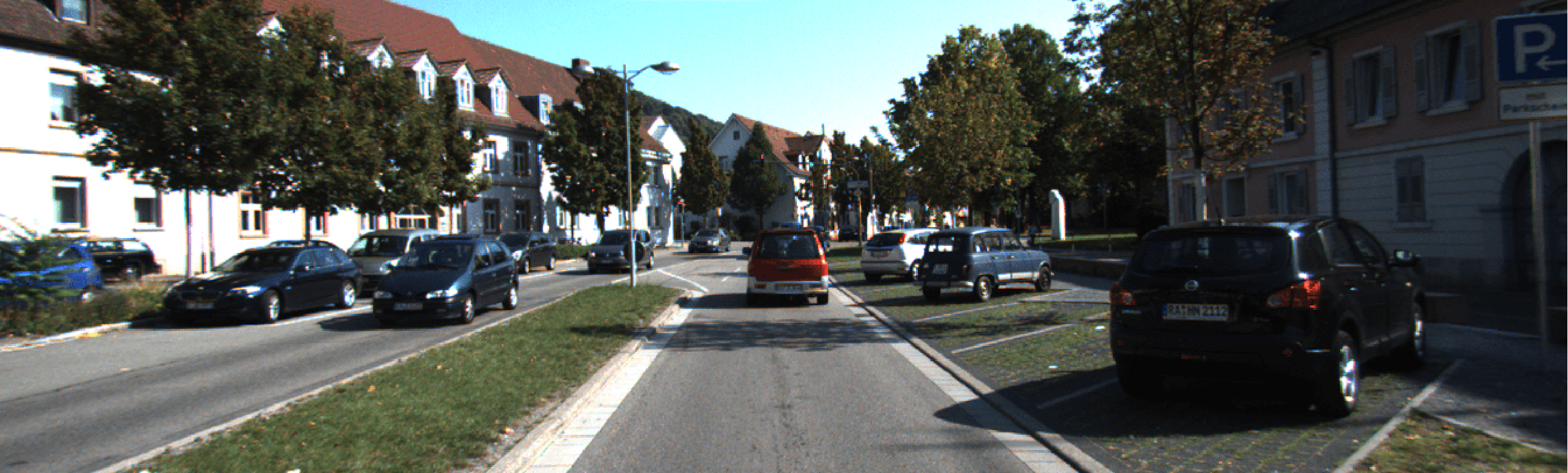} 
		\Tstrut\Bstrut\\	
				
		\rotatebox[origin=c]{90}{\textit{$M^P_{fg}$}} &
		\includegraphics[width=0.29\linewidth,valign=c]{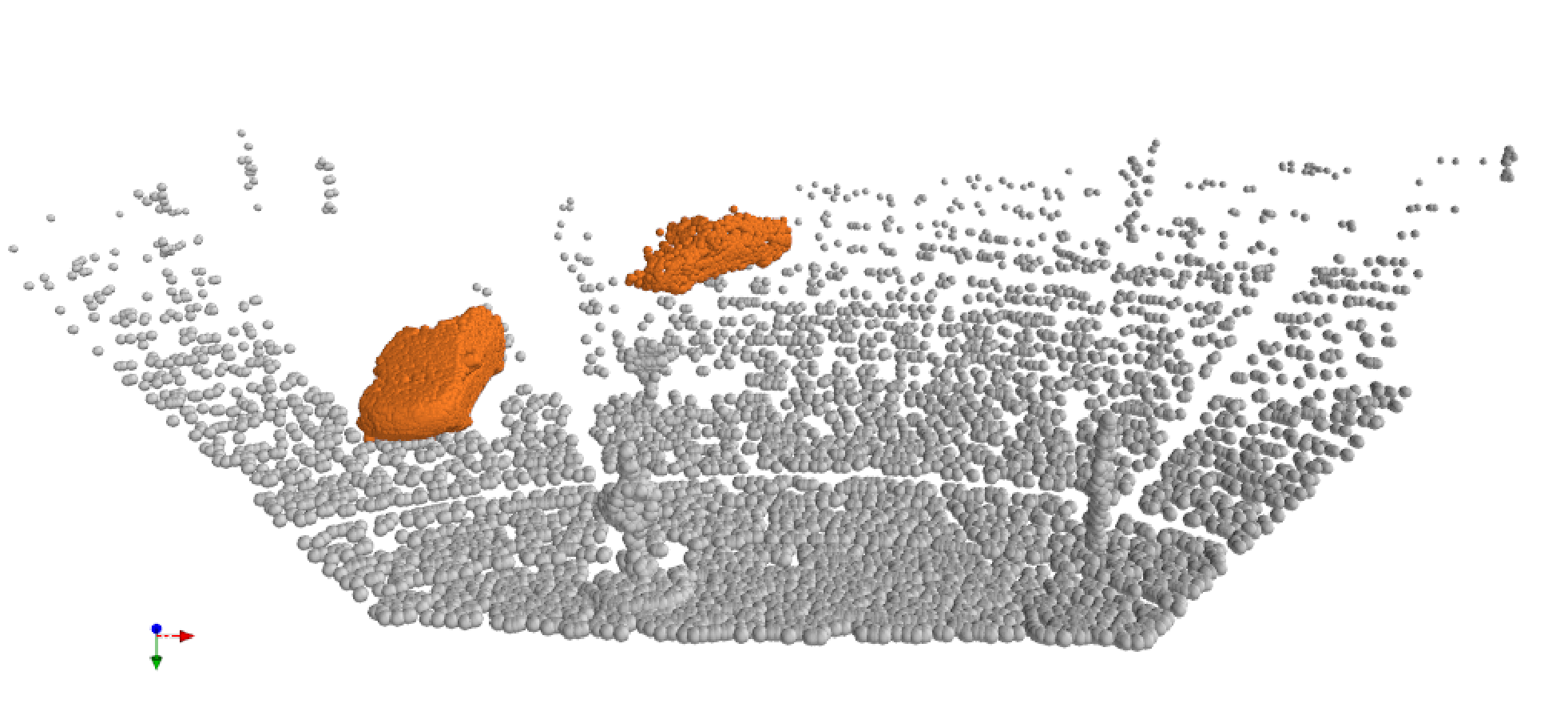}&
		\includegraphics[width=0.29\linewidth,valign=c]{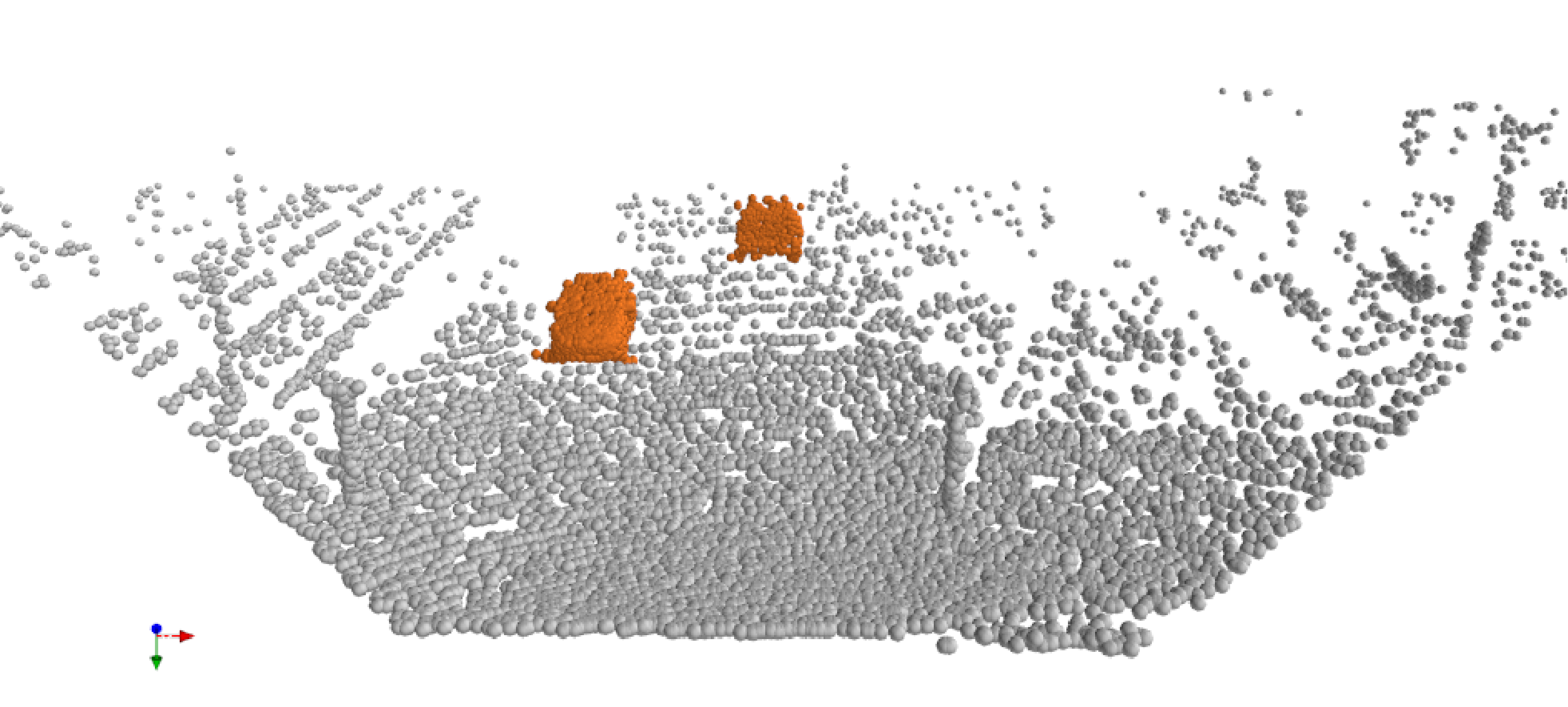}&
		\includegraphics[width=0.29\linewidth,valign=c]{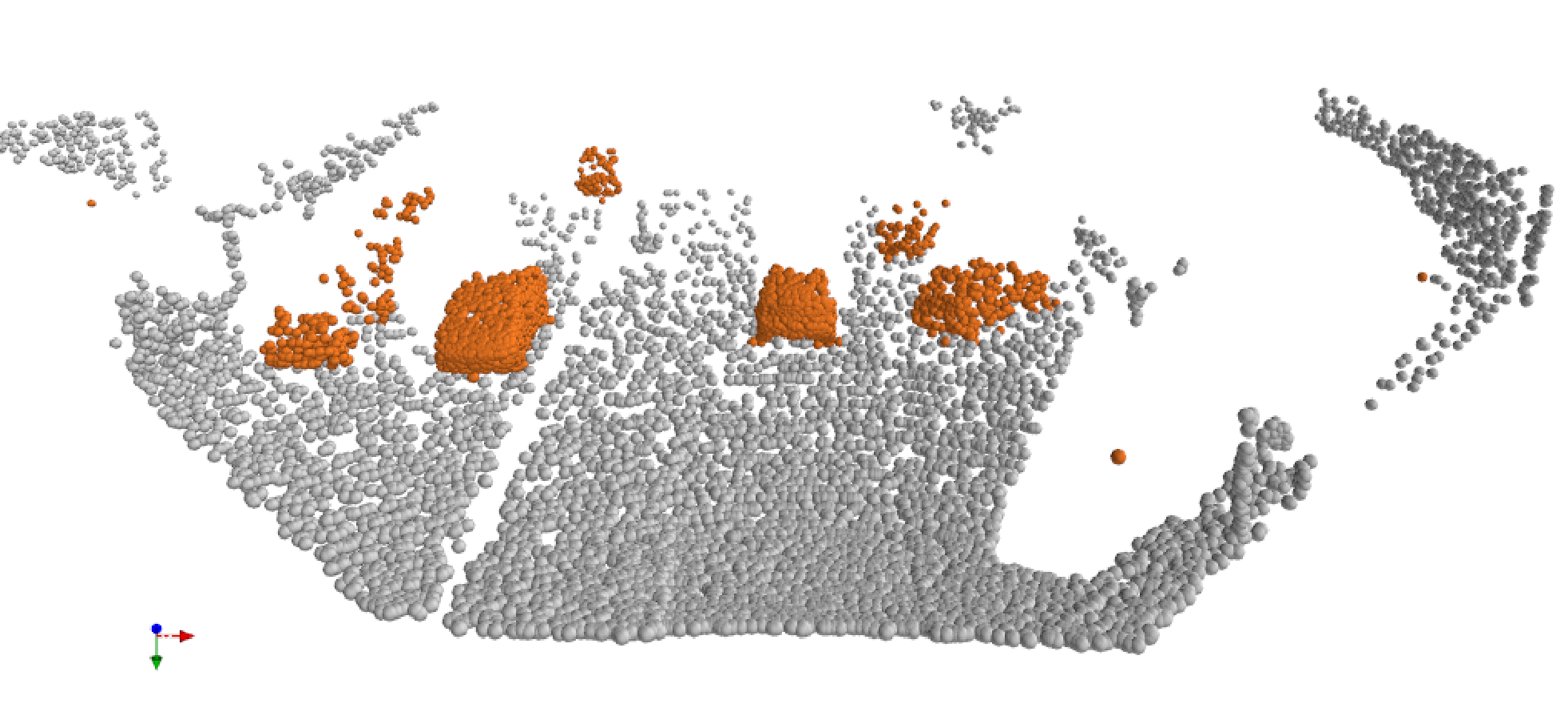}
		\Tstrut\Bstrut\\
		
		\rotatebox[origin=c]{90}{\textit{$M^Q_{fg}$}} &
		\includegraphics[width=0.29\linewidth,valign=c]{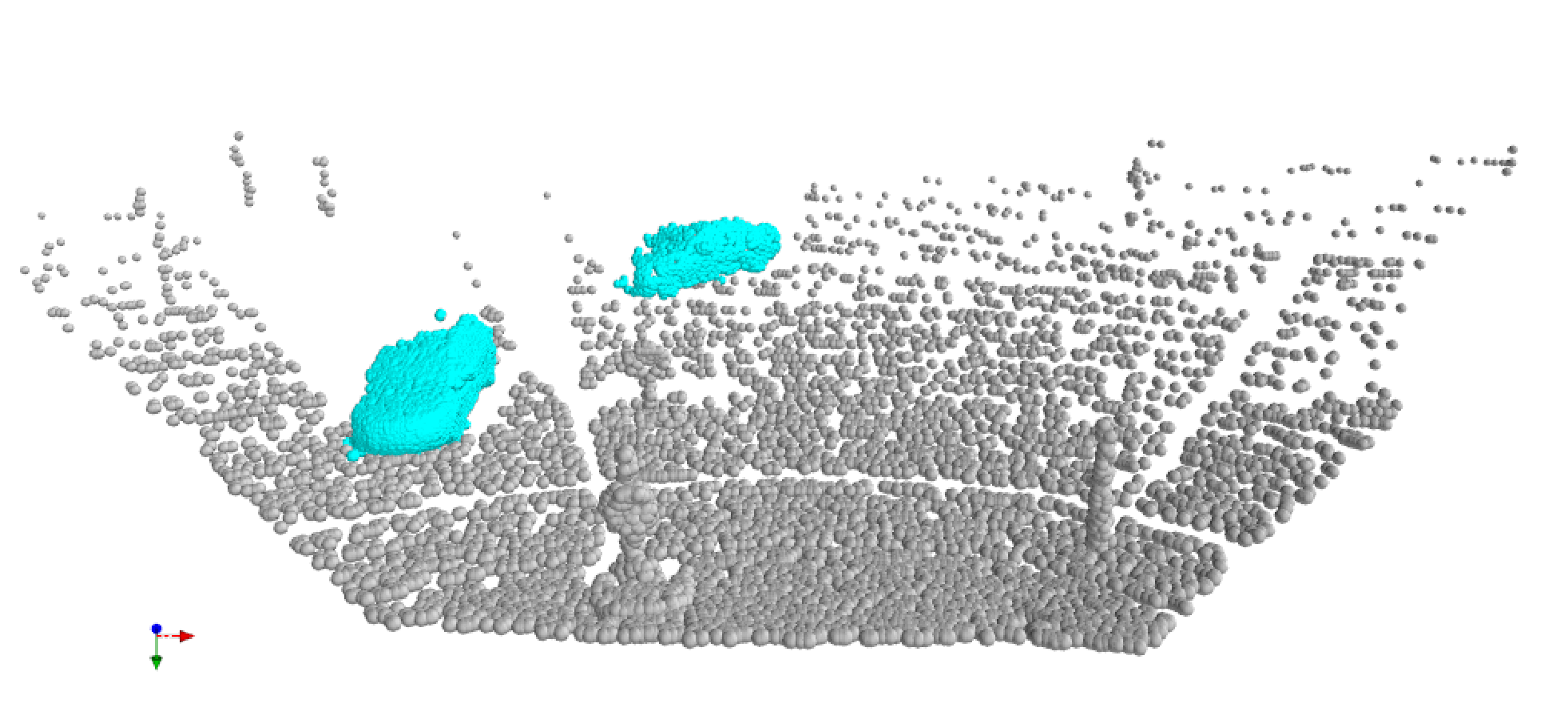}&
		\includegraphics[width=0.29\linewidth,valign=c]{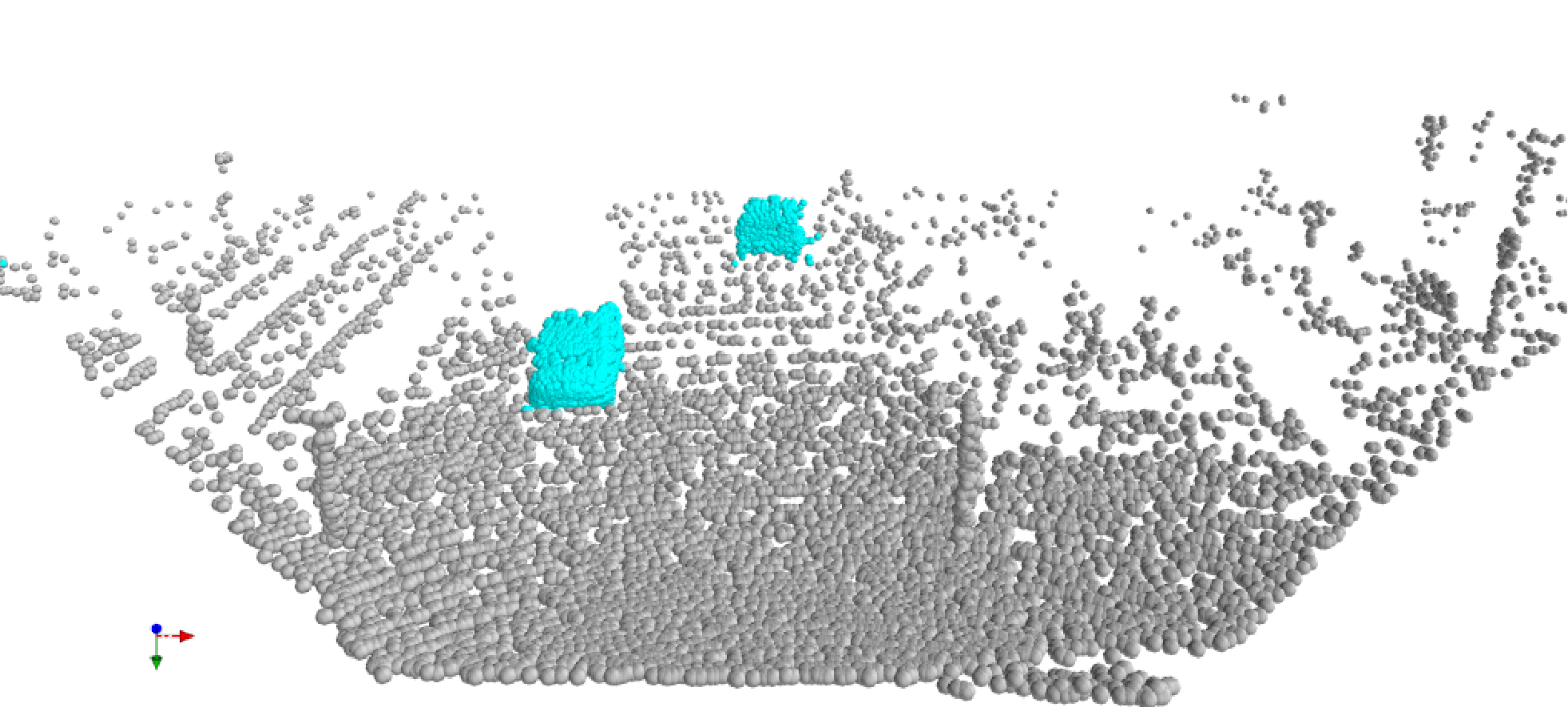}&
		\includegraphics[width=0.29\linewidth,valign=c]{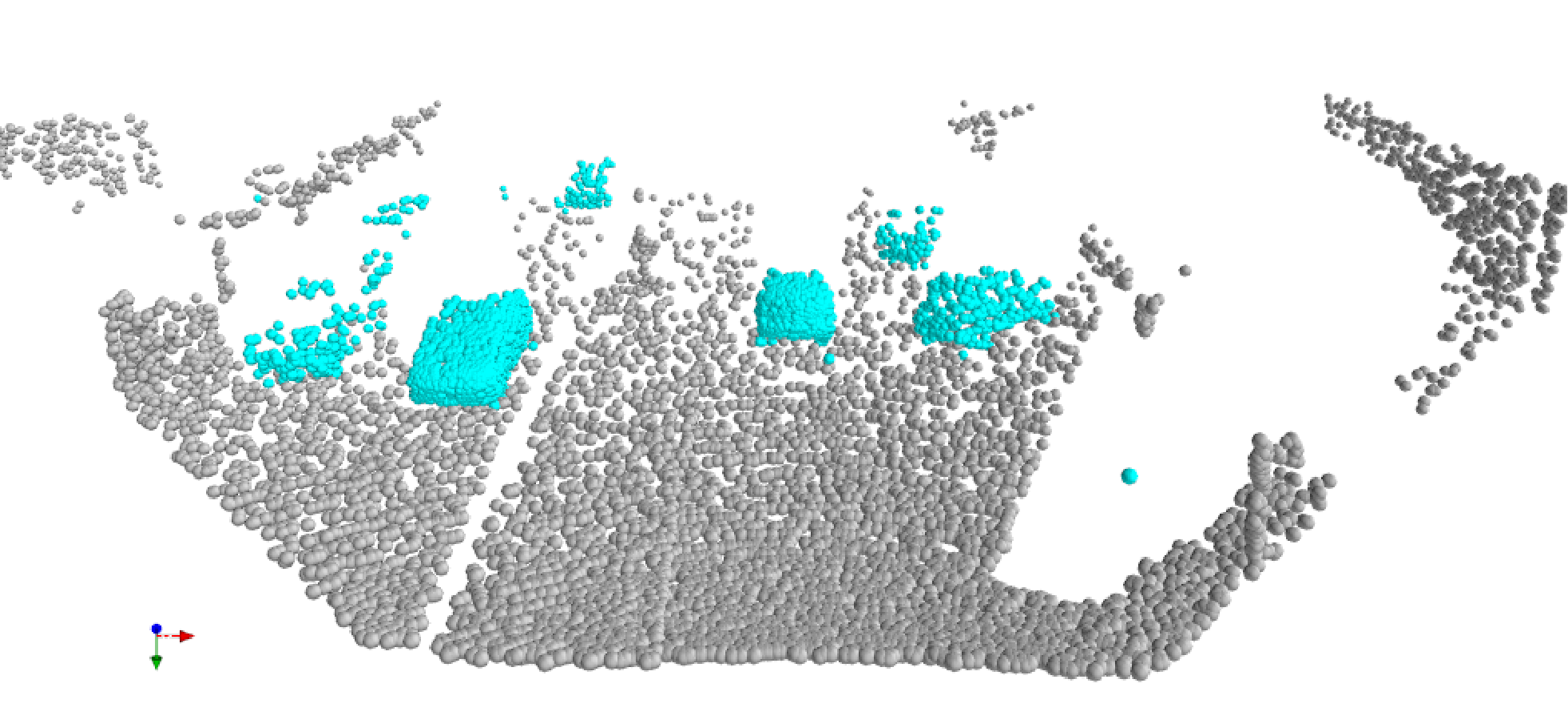}
		\Tstrut\Bstrut\\
		
		\rotatebox[origin=c]{90}{\textit{Error Map}} &
		\includegraphics[width=0.29\linewidth,valign=c]{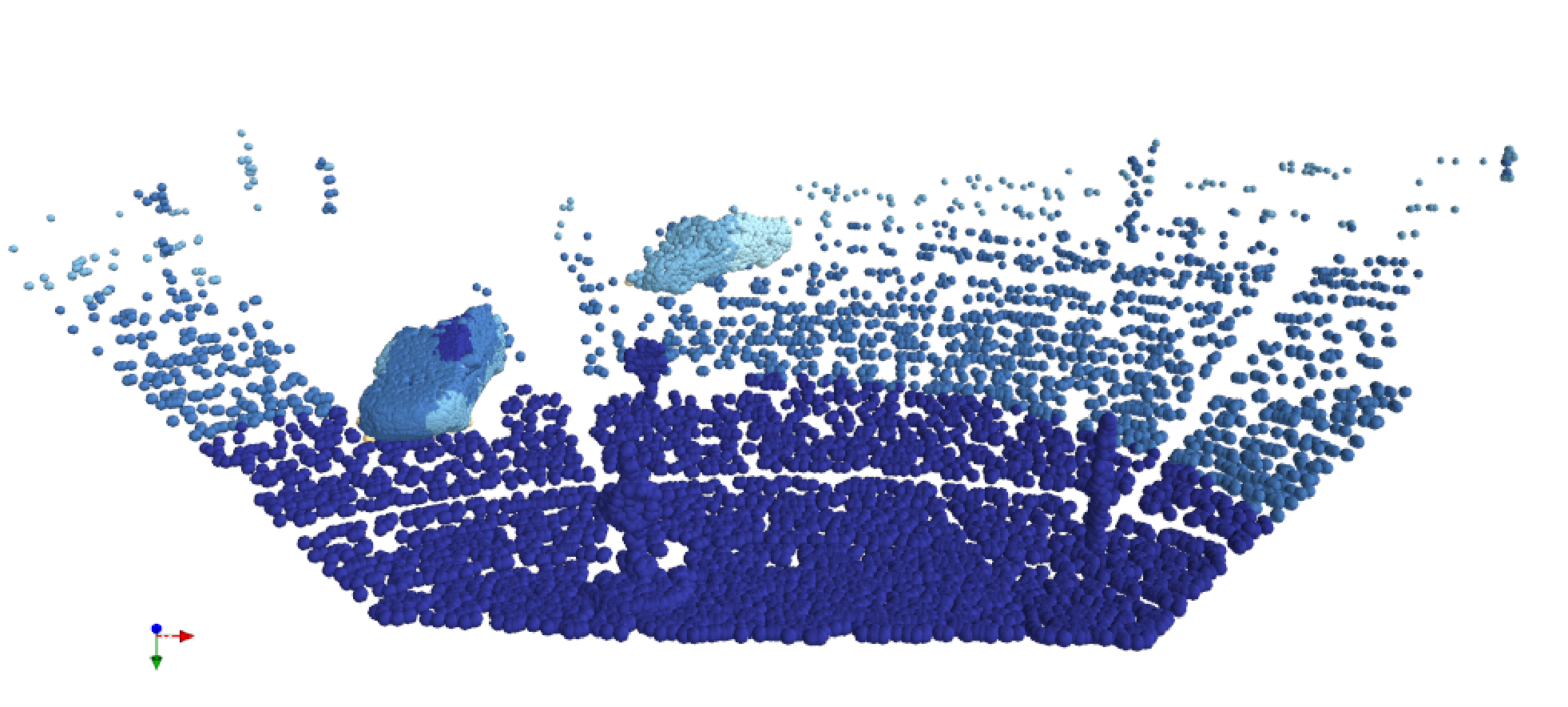}&
		\includegraphics[width=0.29\linewidth,valign=c]{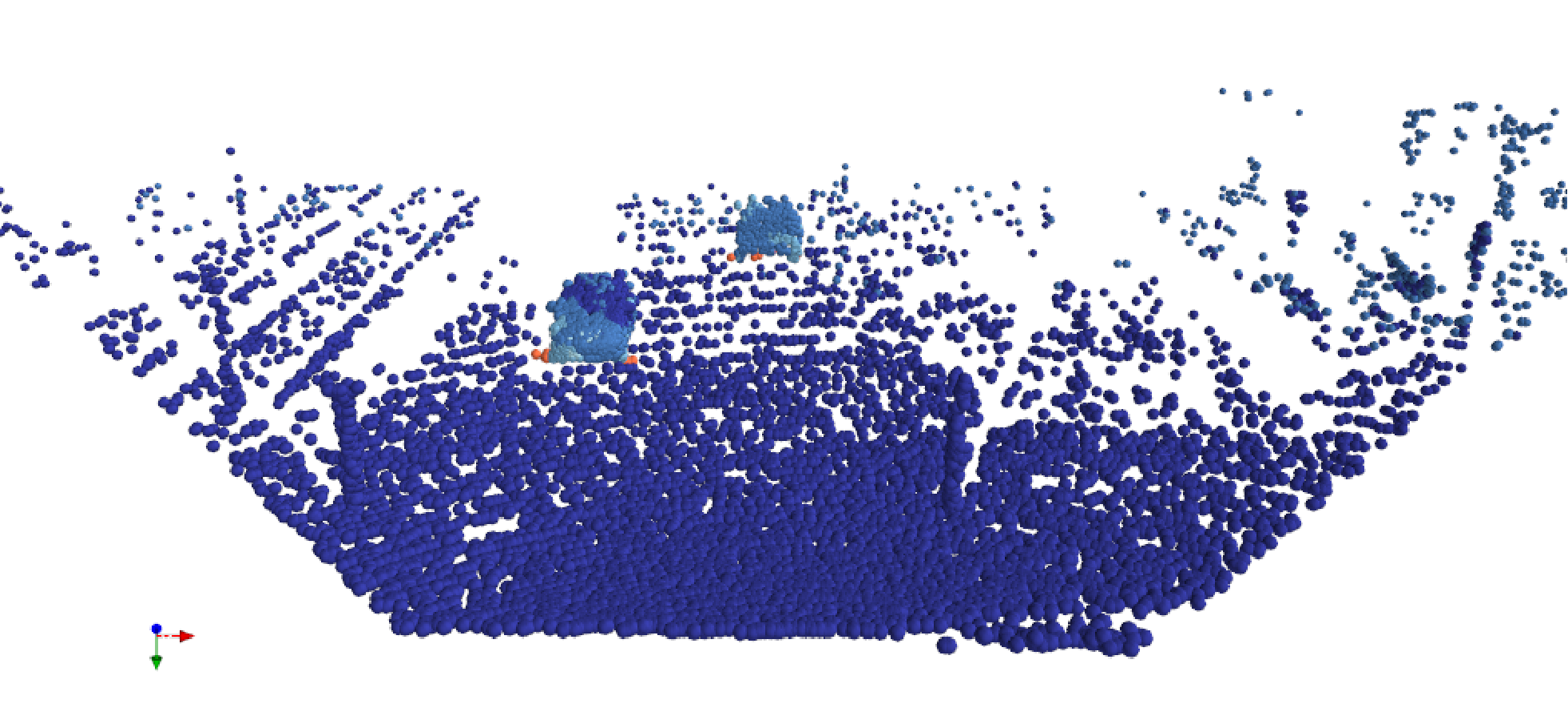}&
		\includegraphics[width=0.29\linewidth,valign=c]{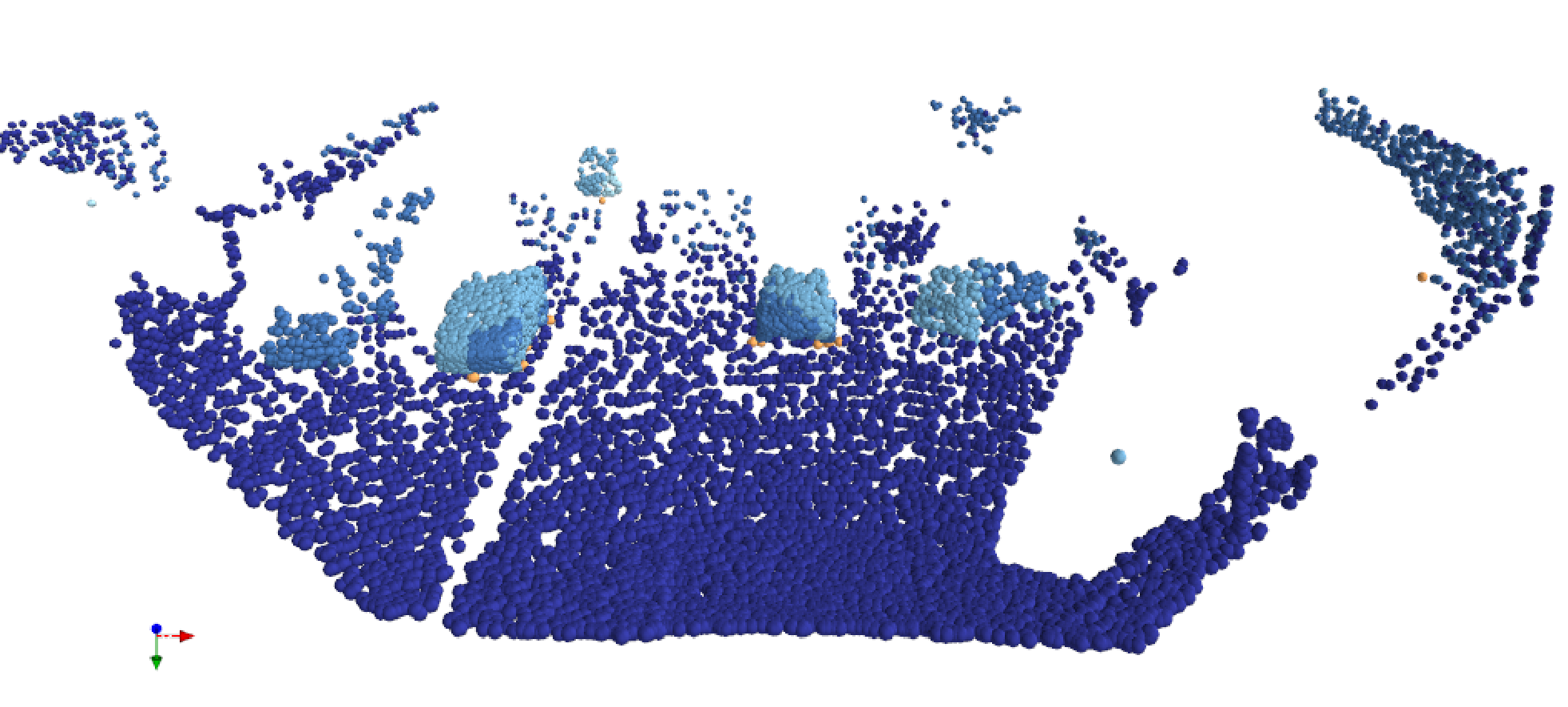}
		\Tstrut\Bstrut\\	
		
		\hline

		& \textit{Example 4} & \textit{Example 5} & \textit{Example 6} 
		\Tstrut\Bstrut\\ 
		\rotatebox[origin=c]{90}{\textit{Scene}} &
		\includegraphics[width=0.29\linewidth,valign=c]{eps_min/Raw_000130_10_img-min.eps}&
		\includegraphics[width=0.29\linewidth,valign=c]{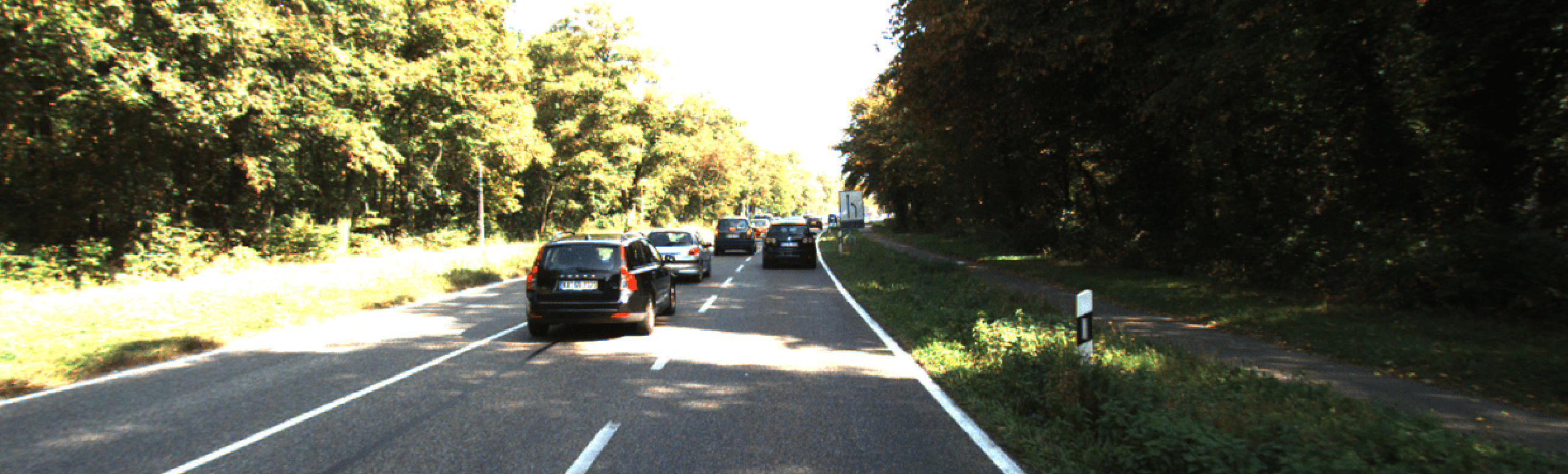}&
		\includegraphics[width=0.29\linewidth,valign=c]{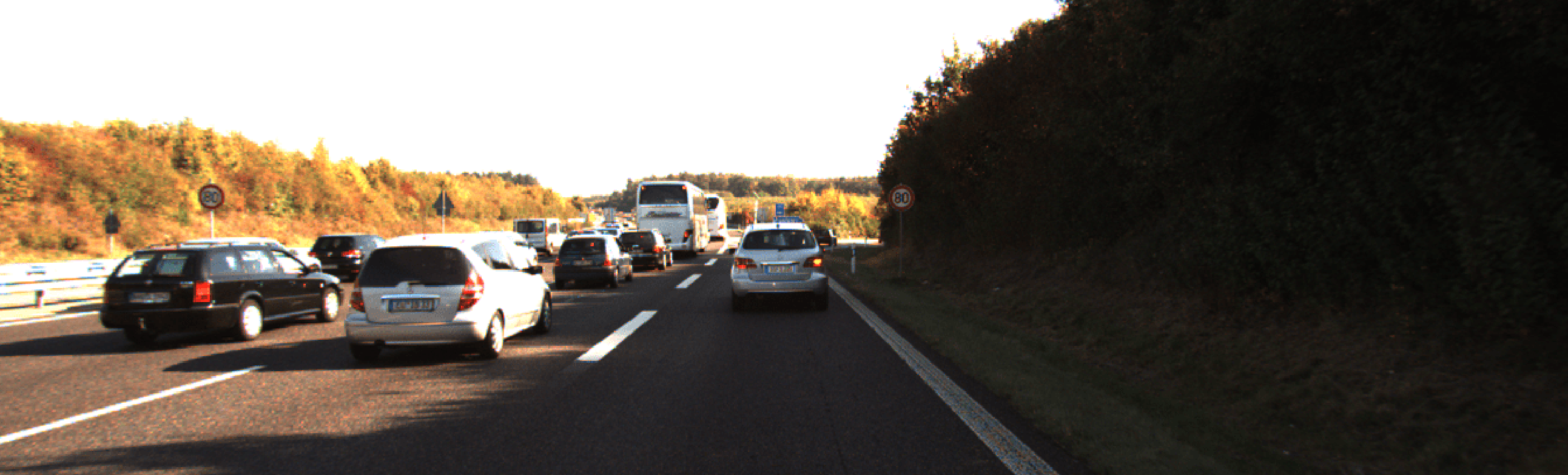}
		\Tstrut\Bstrut\\		

		\rotatebox[origin=c]{90}{\textit{$M^P_{fg}$}} &
		\includegraphics[width=0.29\linewidth,valign=c]{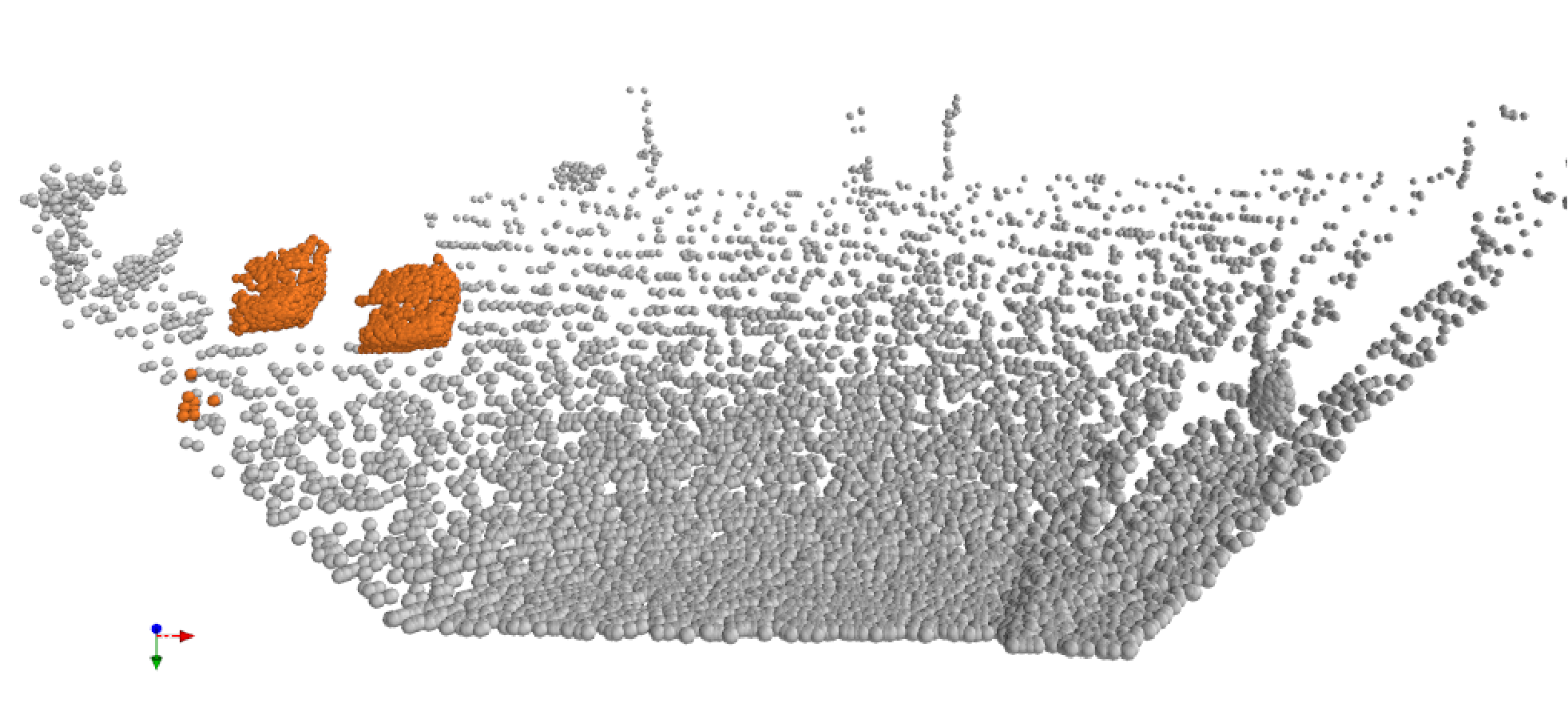}&
		\includegraphics[width=0.29\linewidth,valign=c]{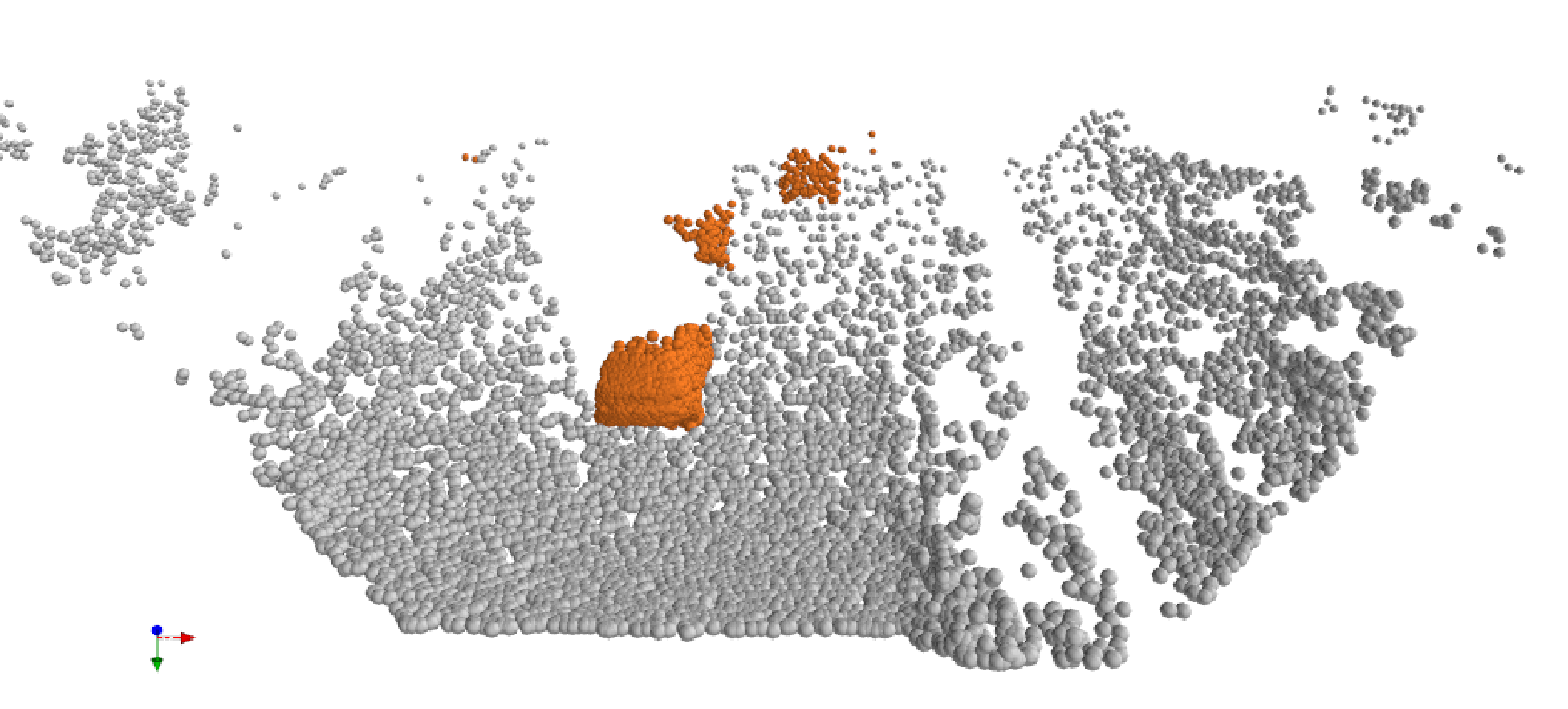}&
		\includegraphics[width=0.29\linewidth,valign=c]{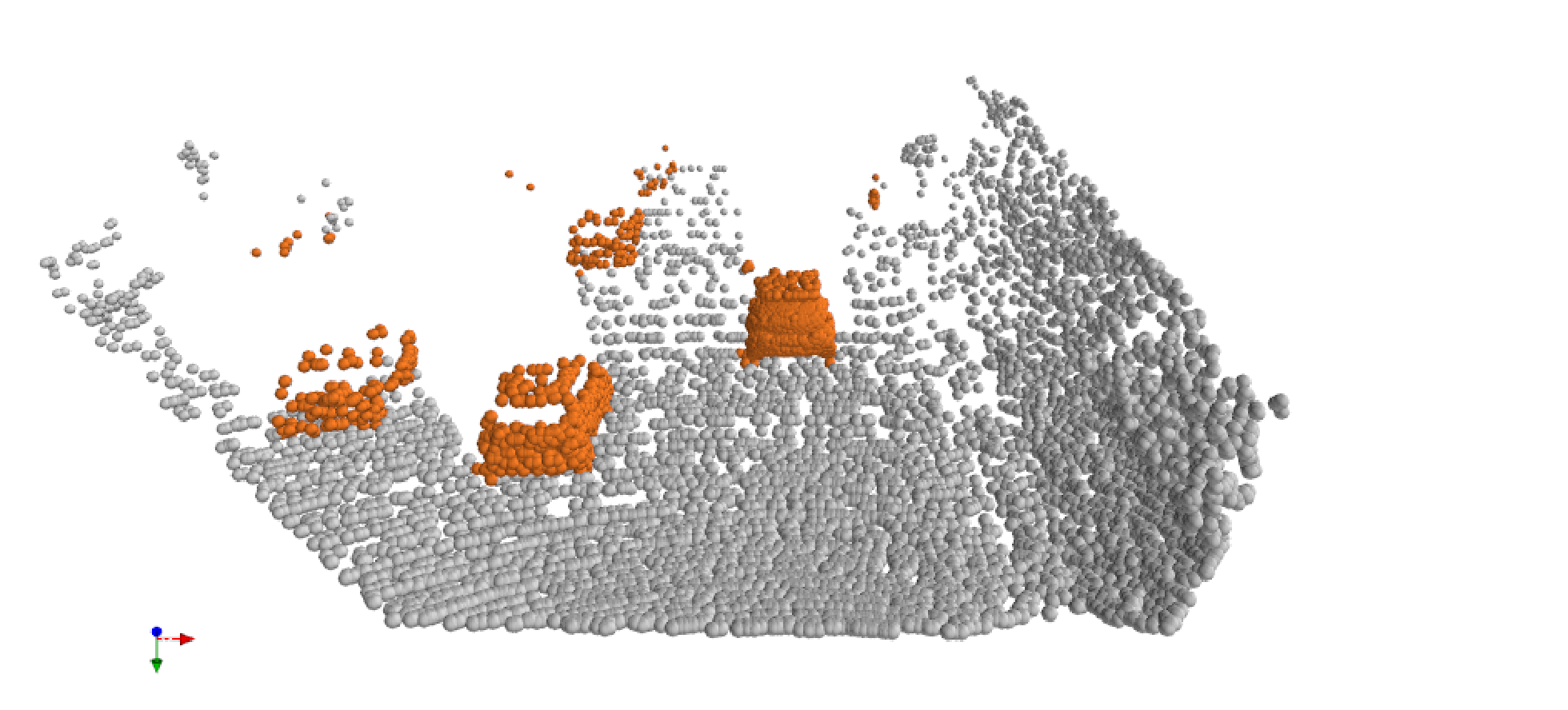}
		\Tstrut\Bstrut\\
		
		\rotatebox[origin=c]{90}{\textit{$M^Q_{fg}$}} &
		\includegraphics[width=0.29\linewidth,valign=c]{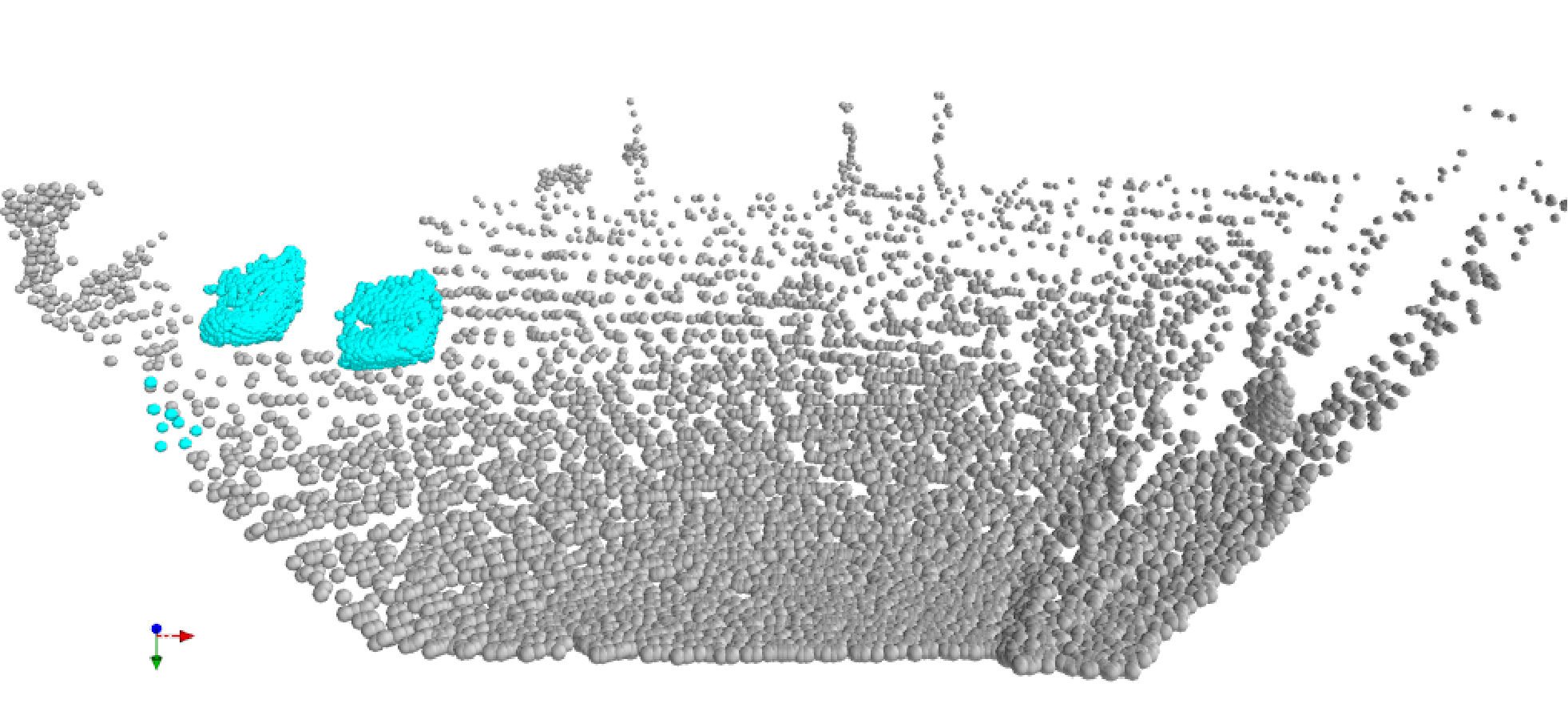}&
		\includegraphics[width=0.29\linewidth,valign=c]{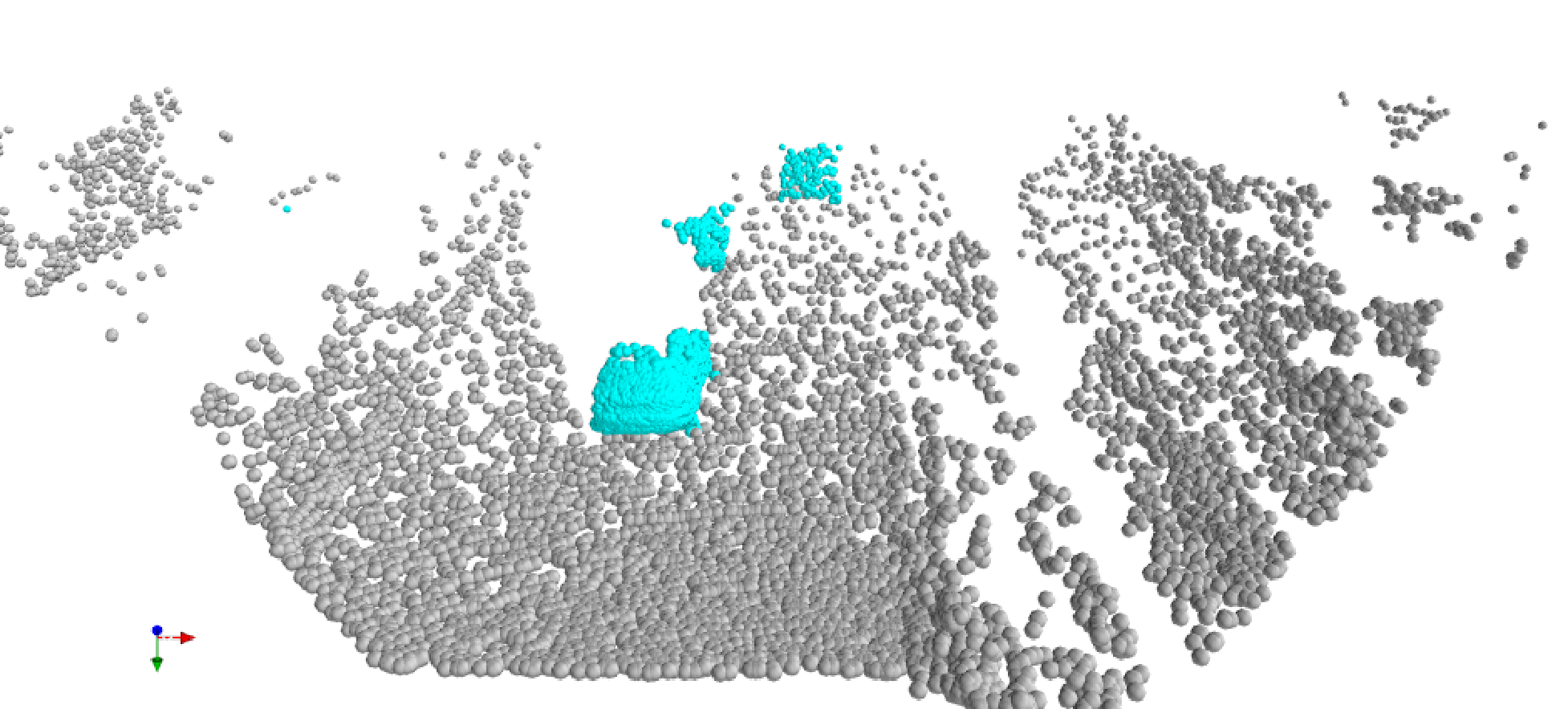}&
		\includegraphics[width=0.29\linewidth,valign=c]{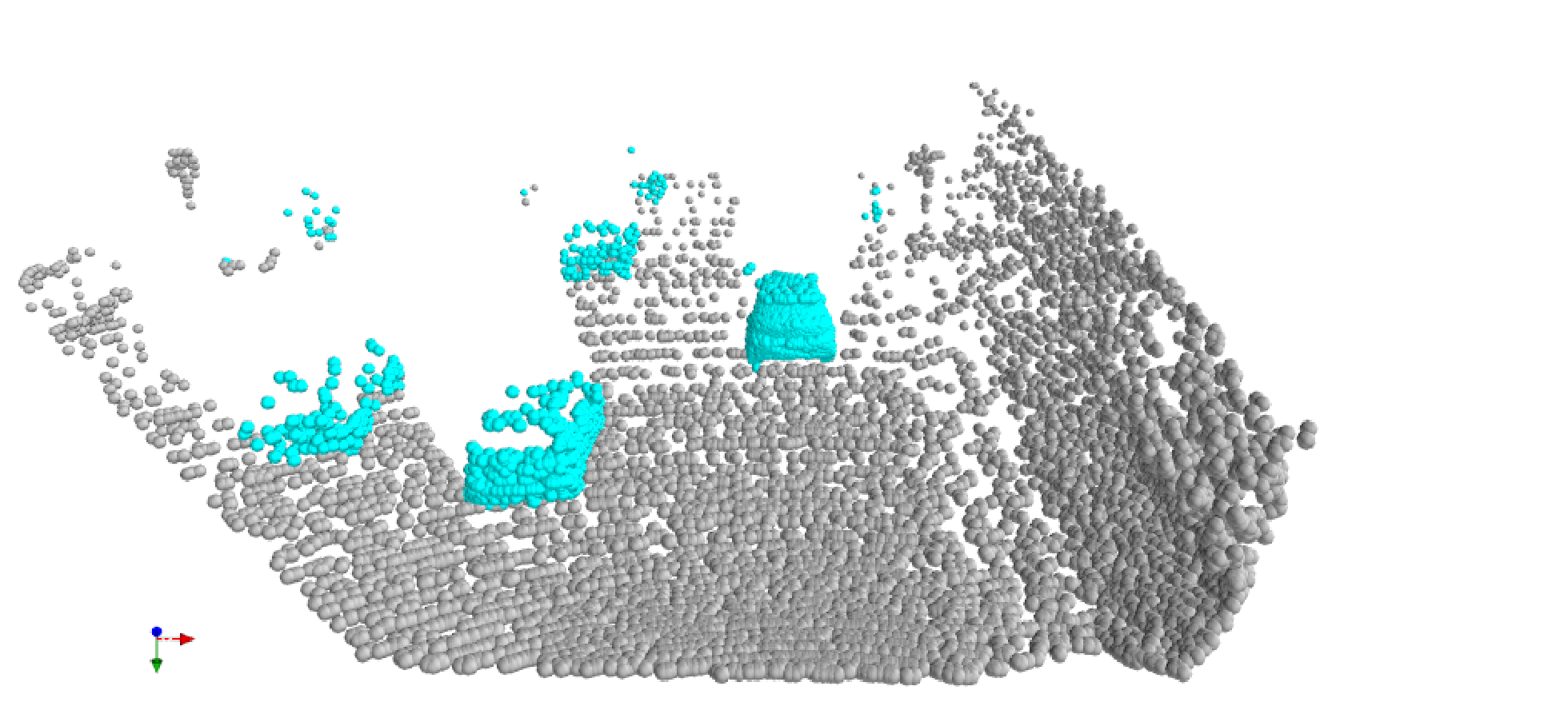}
		\Tstrut\Bstrut\\
		
		\rotatebox[origin=c]{90}{\textit{Error Map}} &
		\includegraphics[width=0.29\linewidth,valign=c]{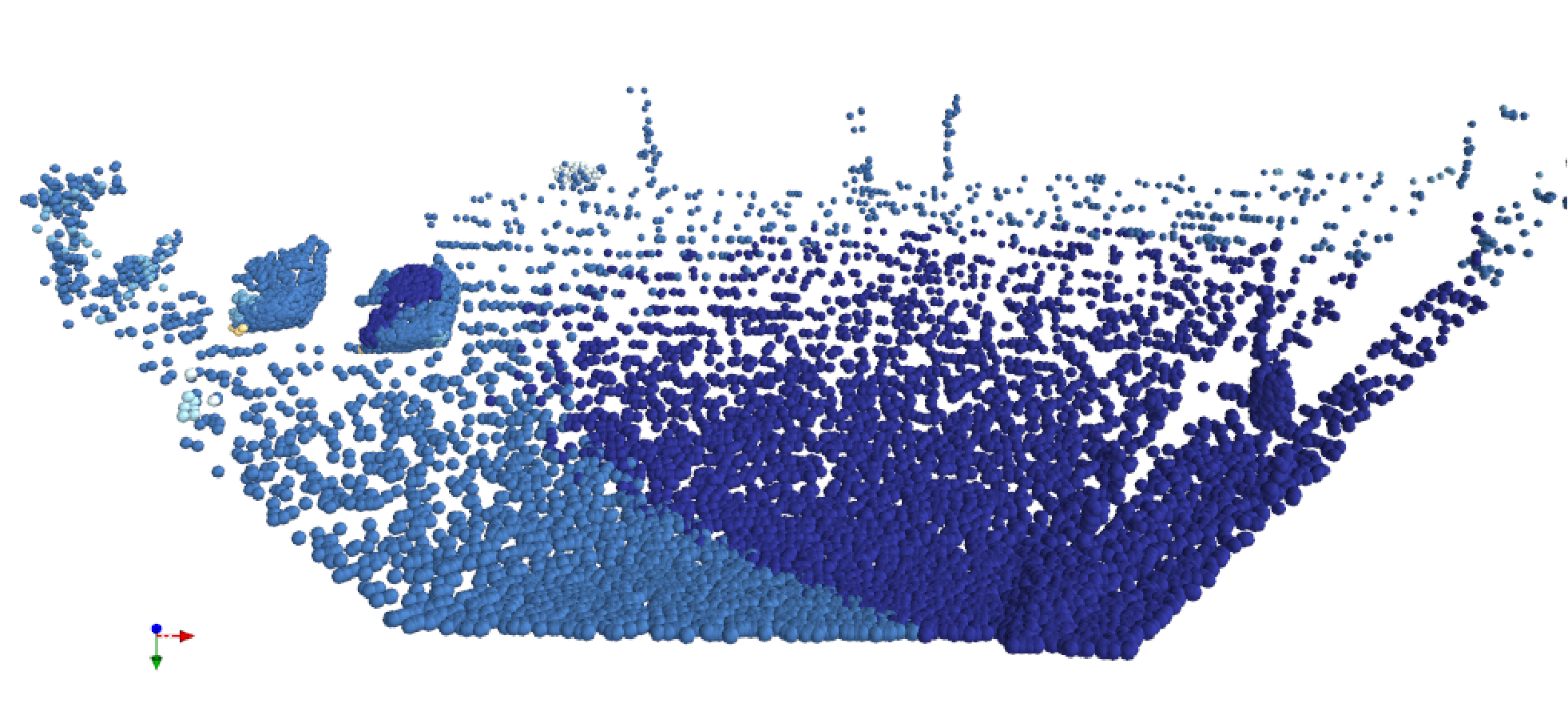}&
		\includegraphics[width=0.29\linewidth,valign=c]{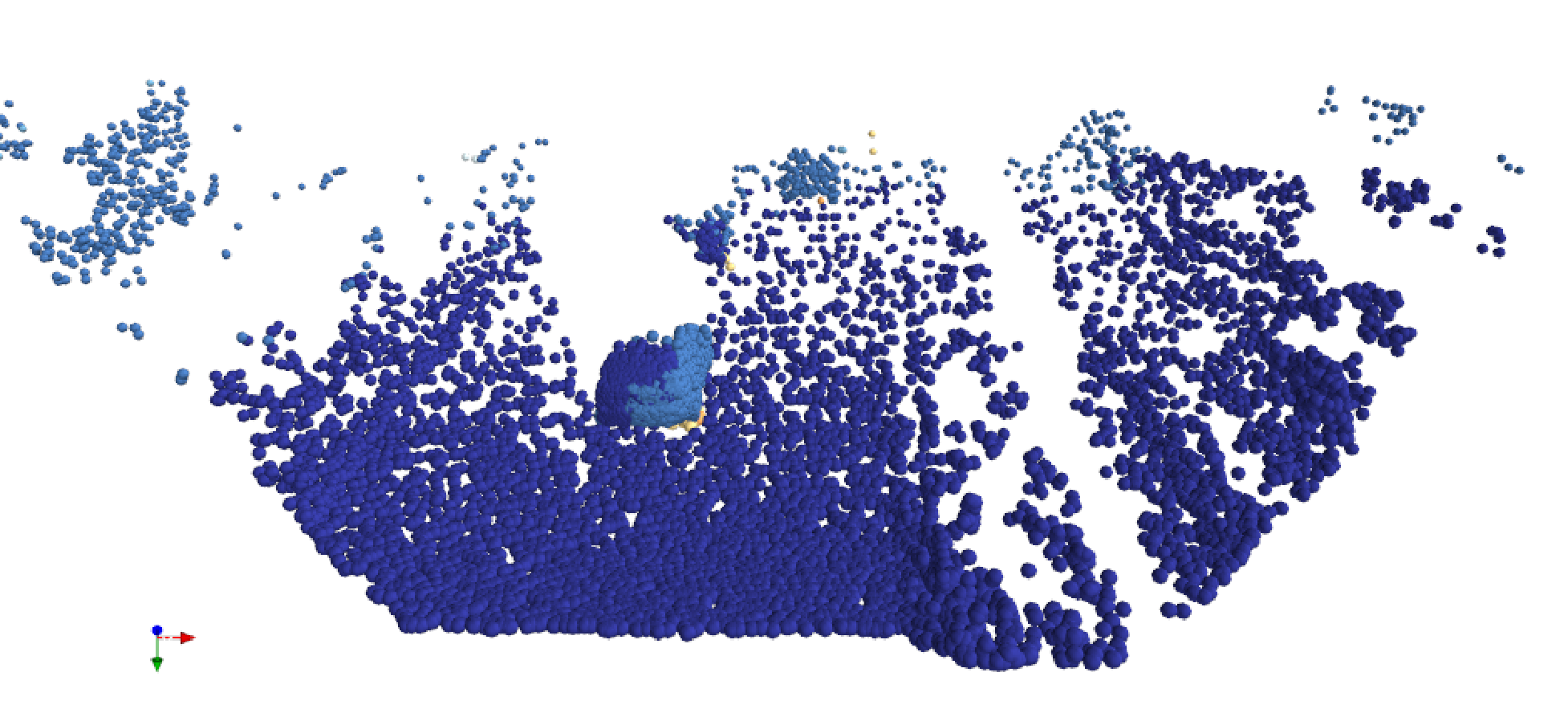}&
		\includegraphics[width=0.29\linewidth,valign=c]{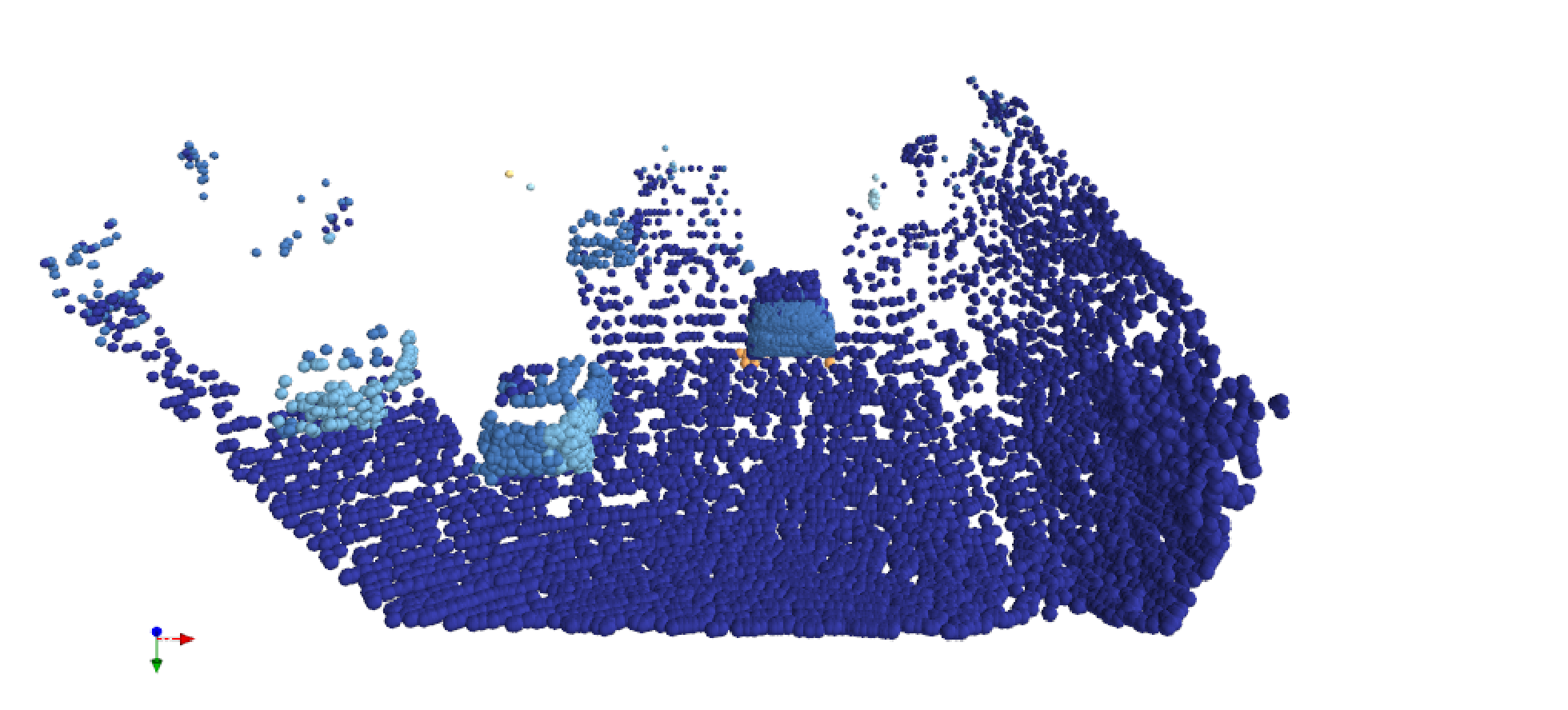}
		\Tstrut\Bstrut\\
		
		& \multicolumn{3}{c}{\includegraphics[width=0.9\linewidth,valign=c]{eps_min/KITTI_errorcolors_3D-min.eps}}\Tstrut\Bstrut\\
	\end{tabular}
	\caption{Six examples from $\mathrm{stereoKITTI}$~\cite{menze2015object} show the qualitative results of our \name{}.}
	\label{Figure:qual_stereo}
\end{figure}

\begin{figure}[t]
	\centering
	\begin{tabular}{p{0.05cm}cccc}
		& \textit{Example 1} & \textit{Example 2} & \textit{Example 3} 
		\Tstrut\Bstrut\\ 
		
		\rotatebox[origin=c]{90}{\textit{Scene}} &
		\includegraphics[width=0.29\linewidth,valign=c]{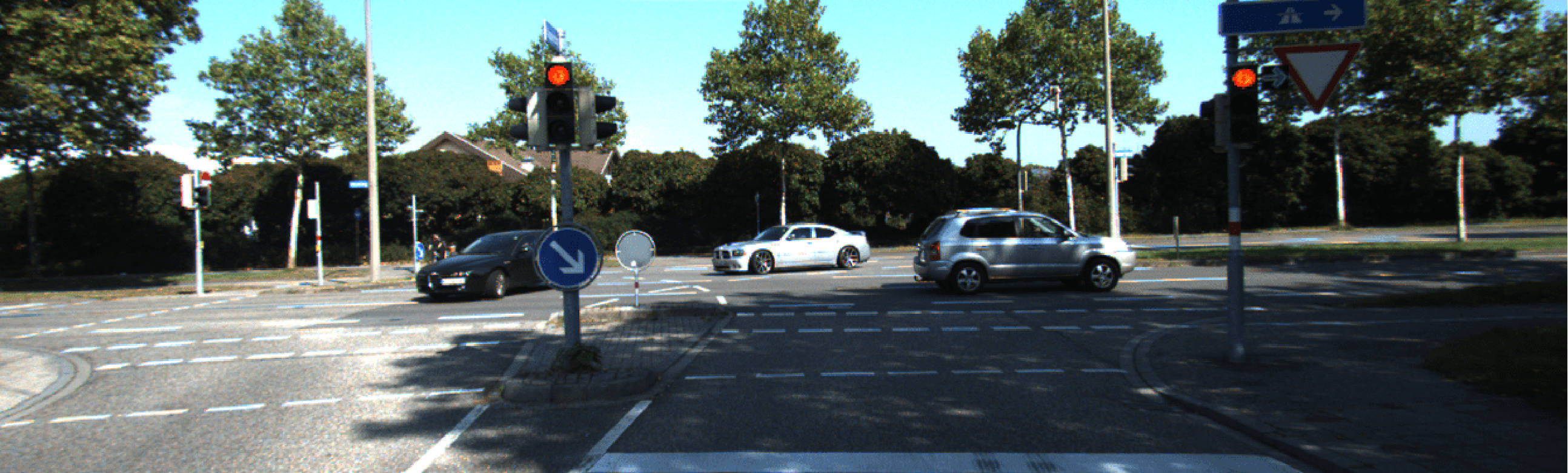}& 
		\includegraphics[width=0.29\linewidth,valign=c]{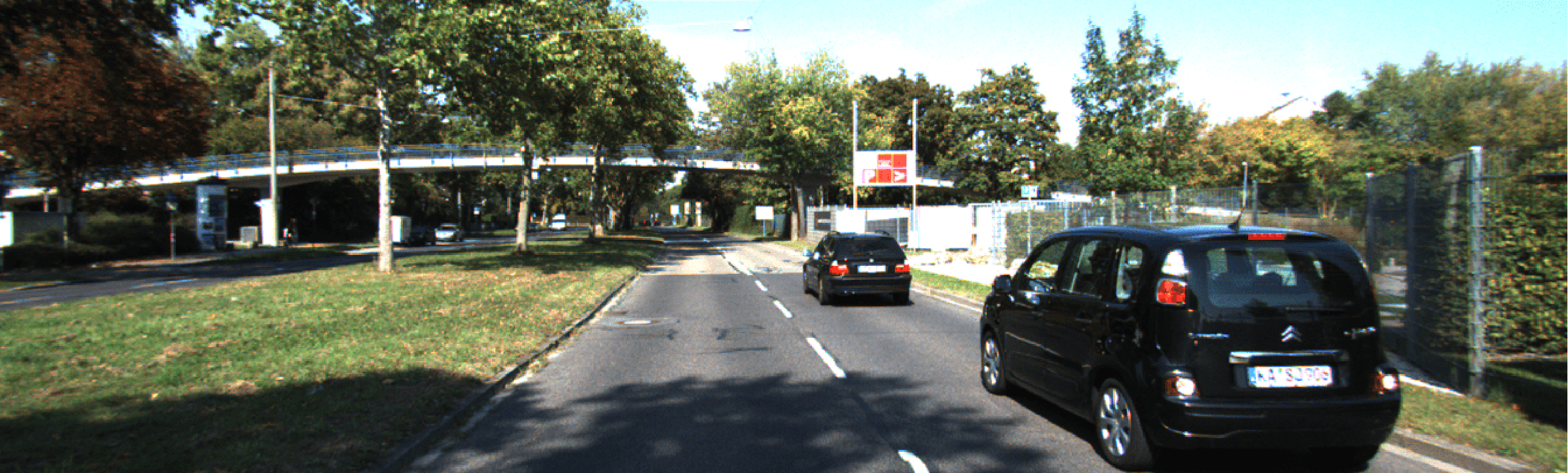}& 
ng		\includegraphics[width=0.29\linewidth,valign=c]{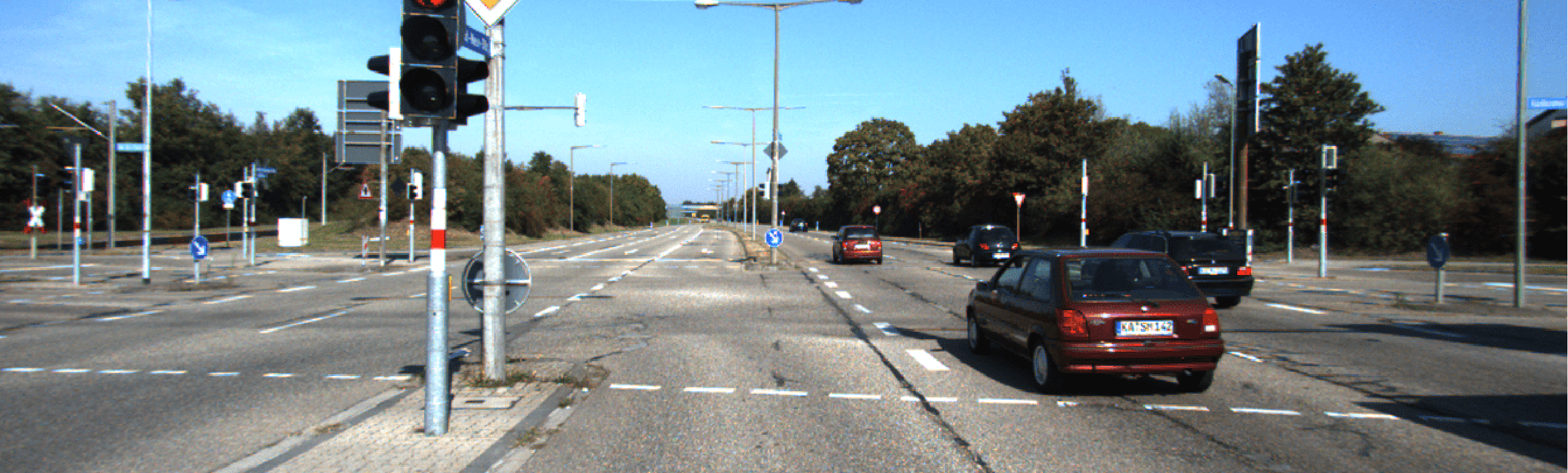} 
		\Tstrut\Bstrut\\	
		
		\rotatebox[origin=c]{90}{\textit{$M^P_{fg}$}} &
		\includegraphics[width=0.29\linewidth,valign=c]{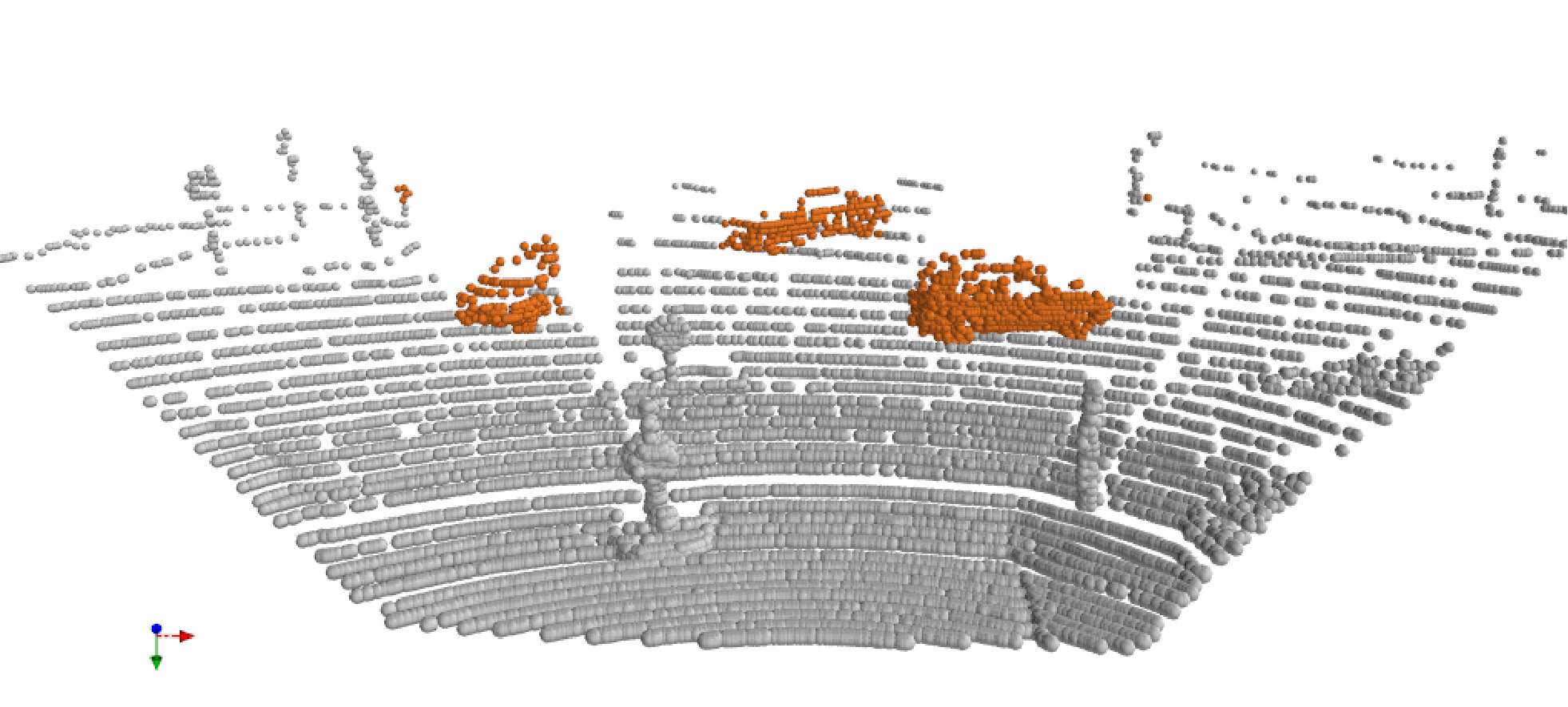}&
		\includegraphics[width=0.29\linewidth,valign=c]{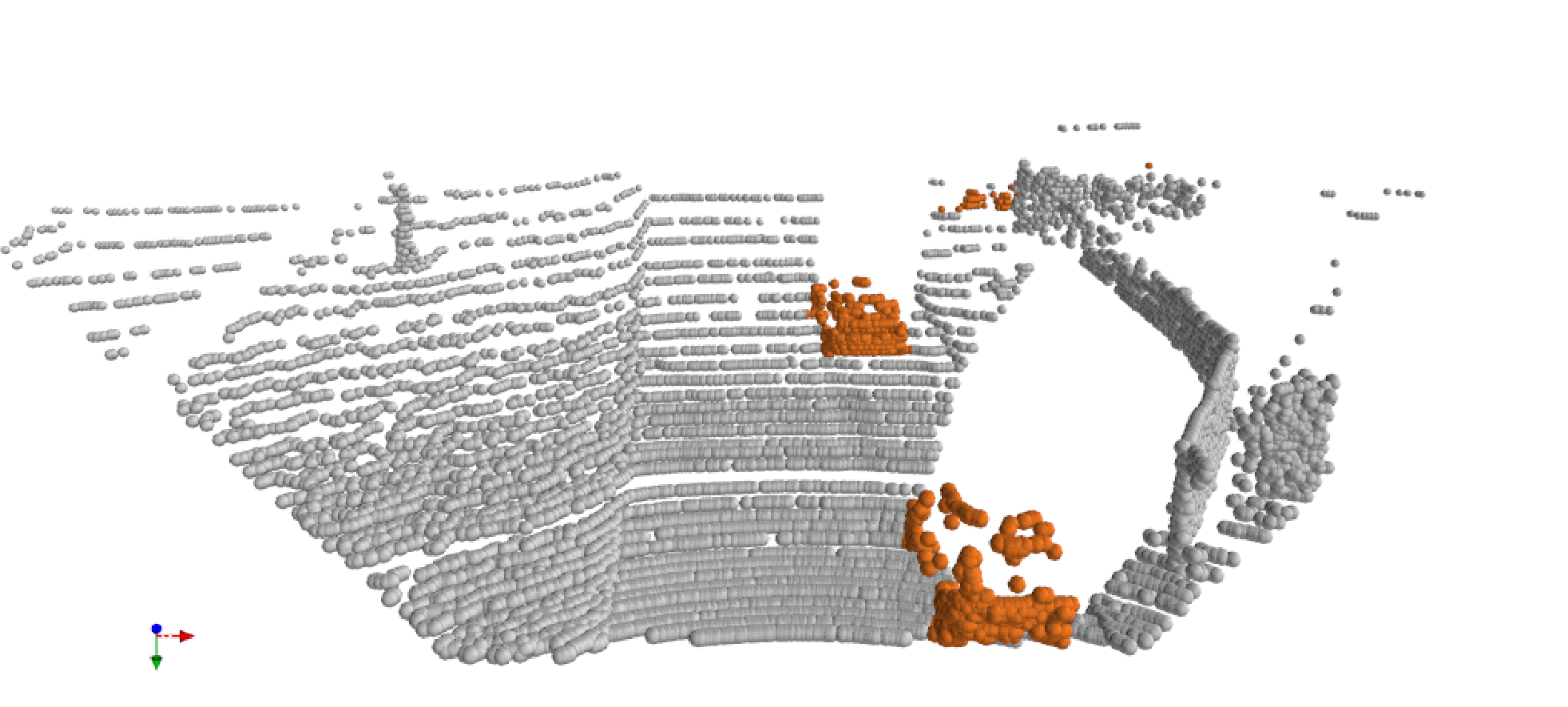}&
		\includegraphics[width=0.29\linewidth,valign=c]{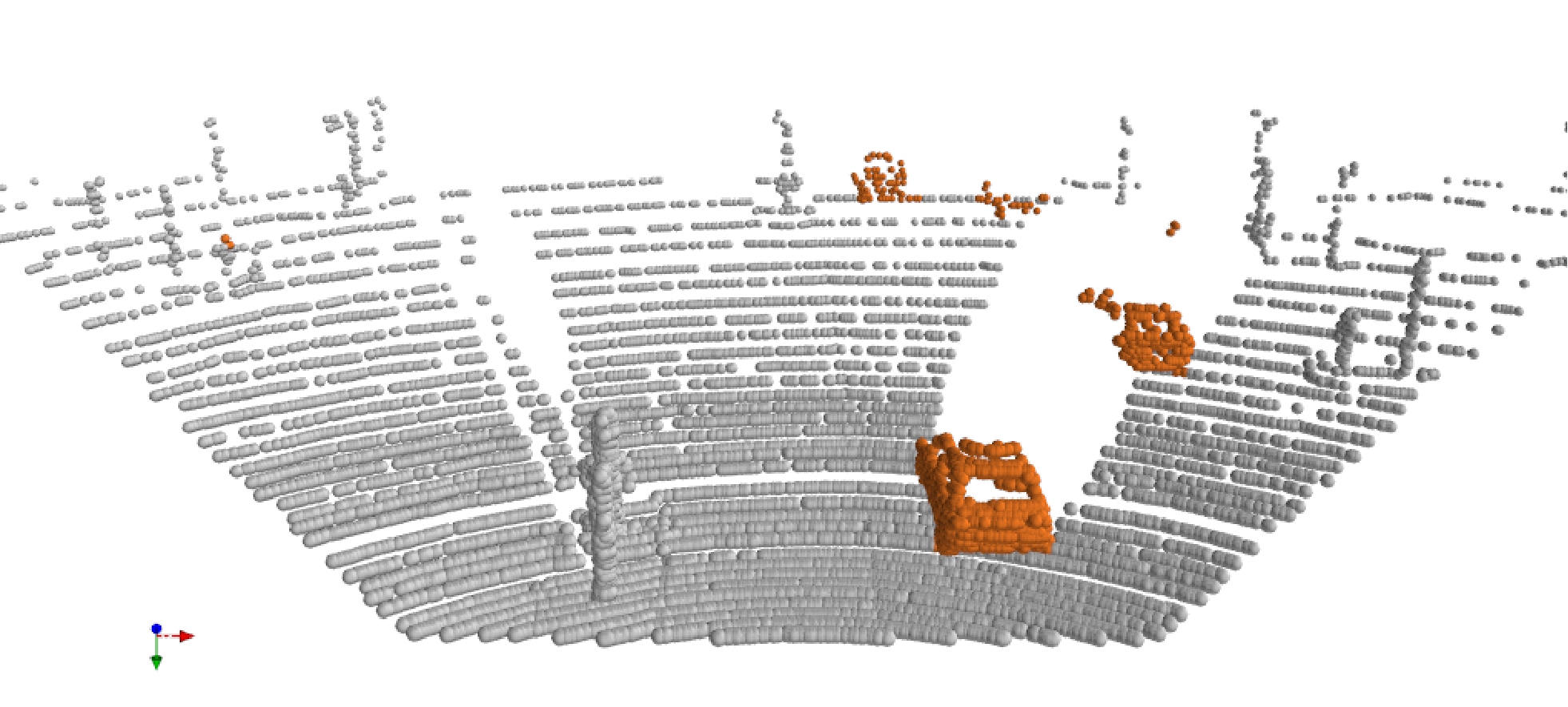}
		\Tstrut\Bstrut\\
		
		\rotatebox[origin=c]{90}{\textit{$M^Q_{fg}$}} &
		\includegraphics[width=0.29\linewidth,valign=c]{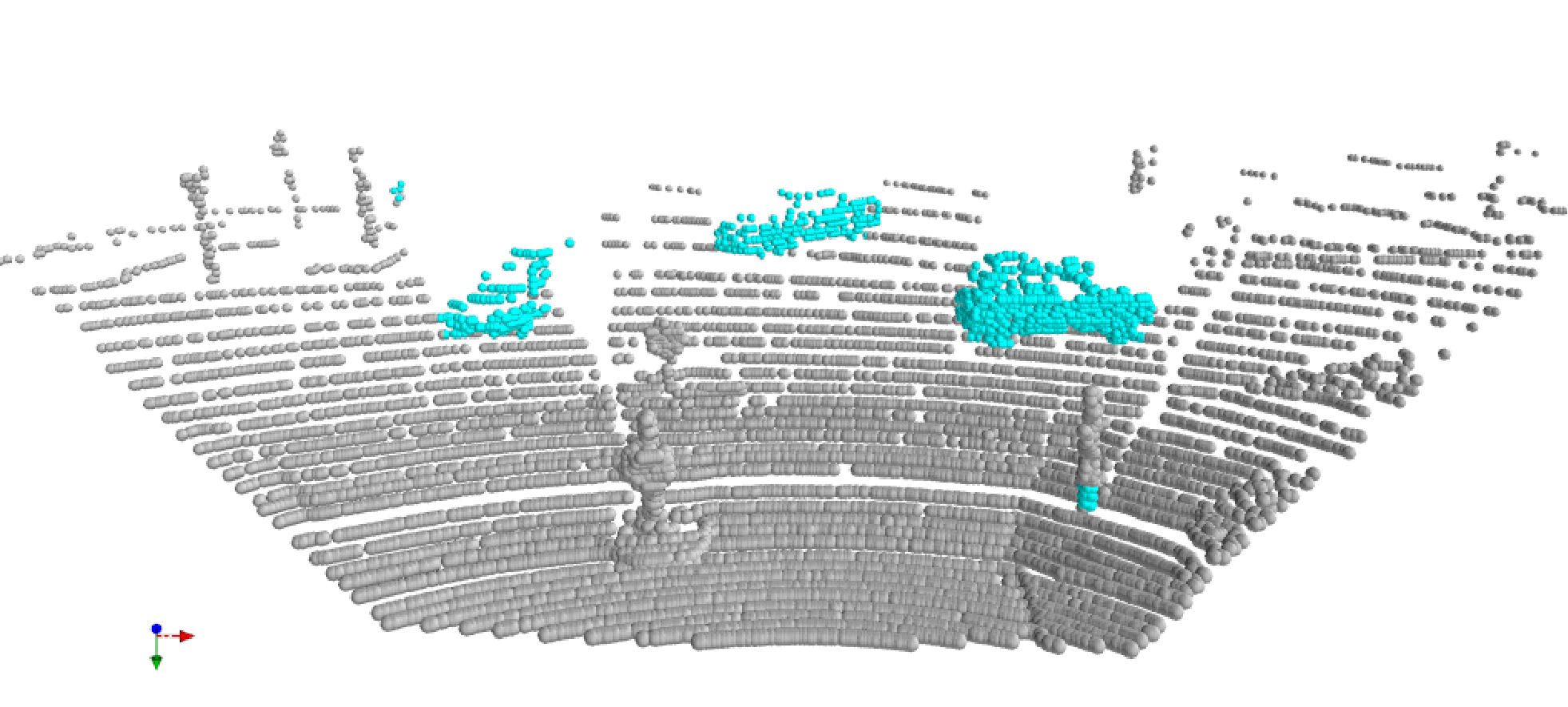}&
		\includegraphics[width=0.29\linewidth,valign=c]{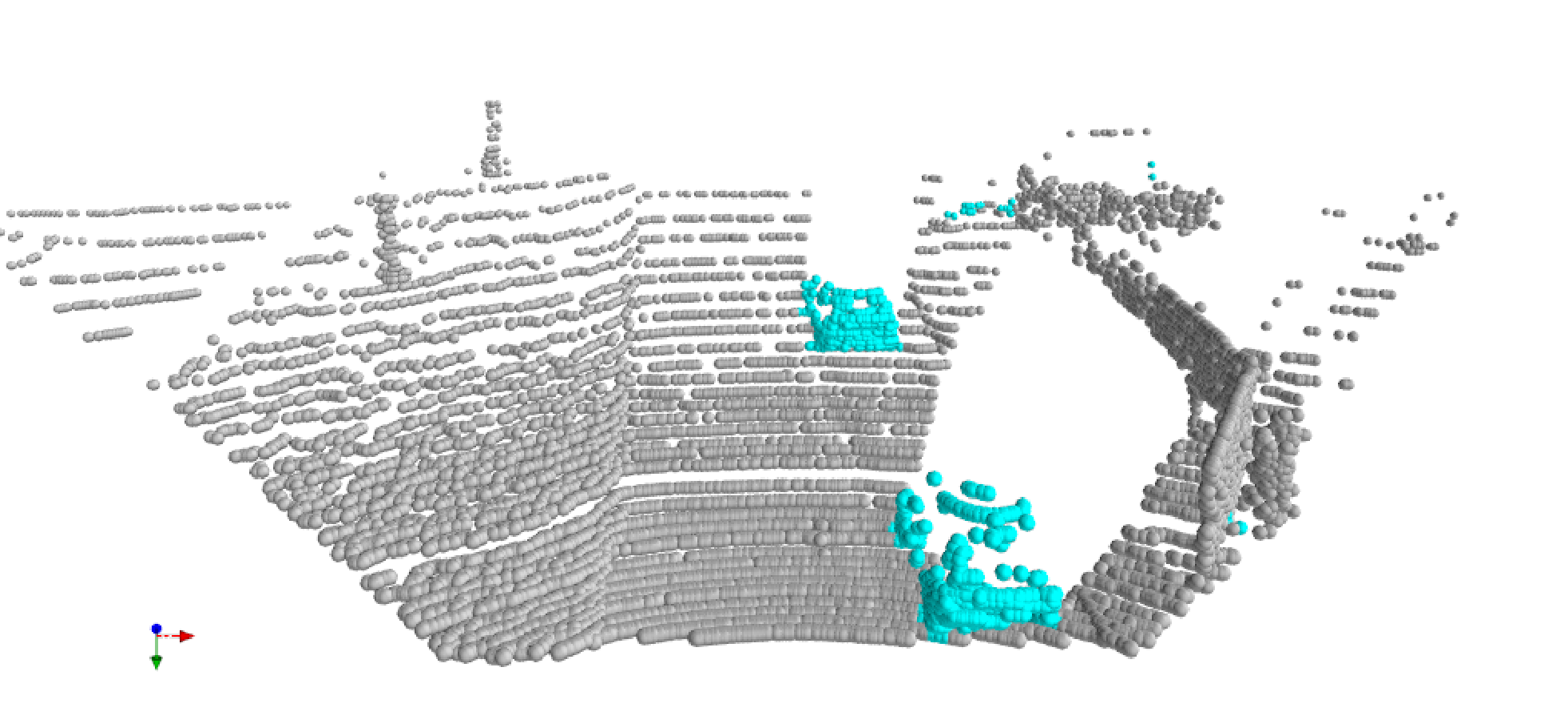}&
		\includegraphics[width=0.29\linewidth,valign=c]{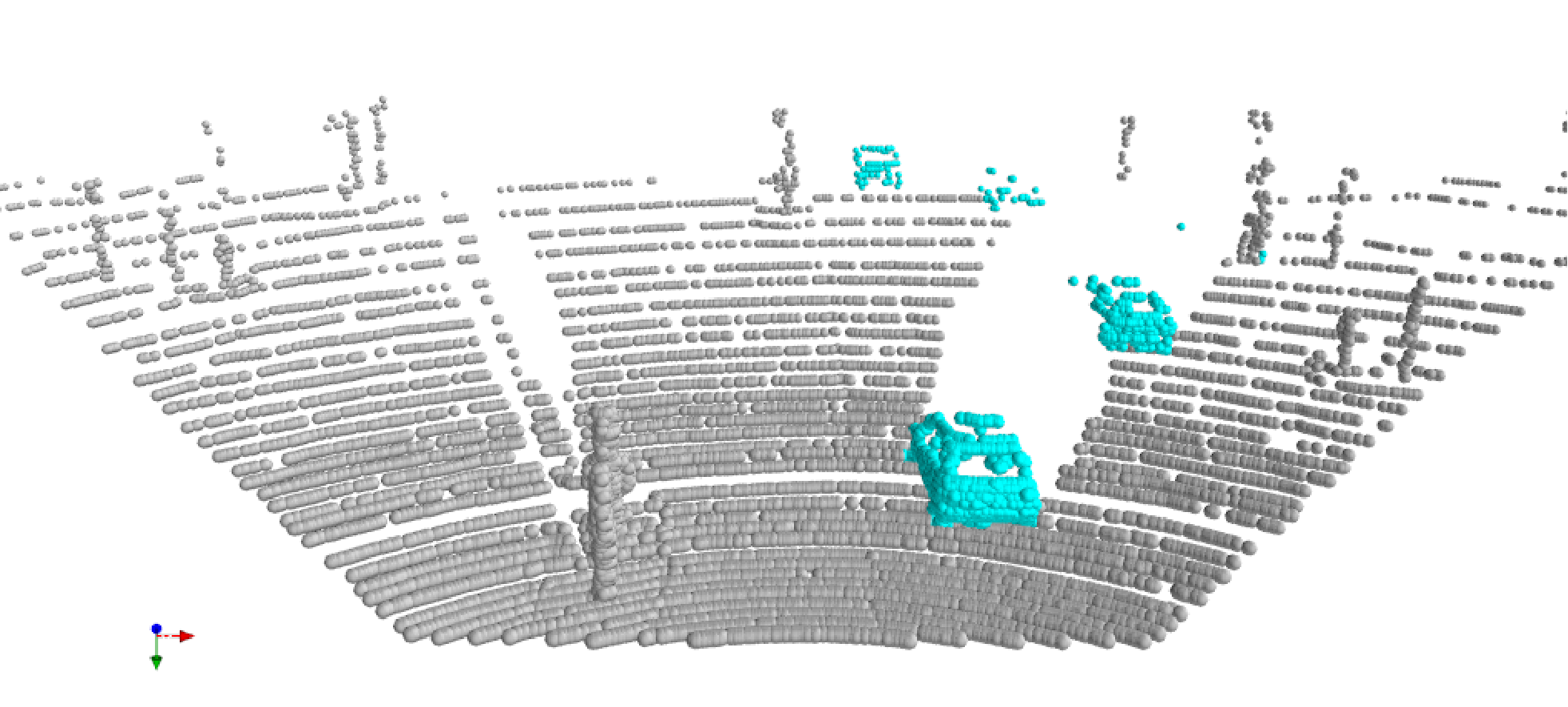}
		\Tstrut\Bstrut\\
		
		\rotatebox[origin=c]{90}{\textit{Error Map}} &
		\includegraphics[width=0.29\linewidth,valign=c]{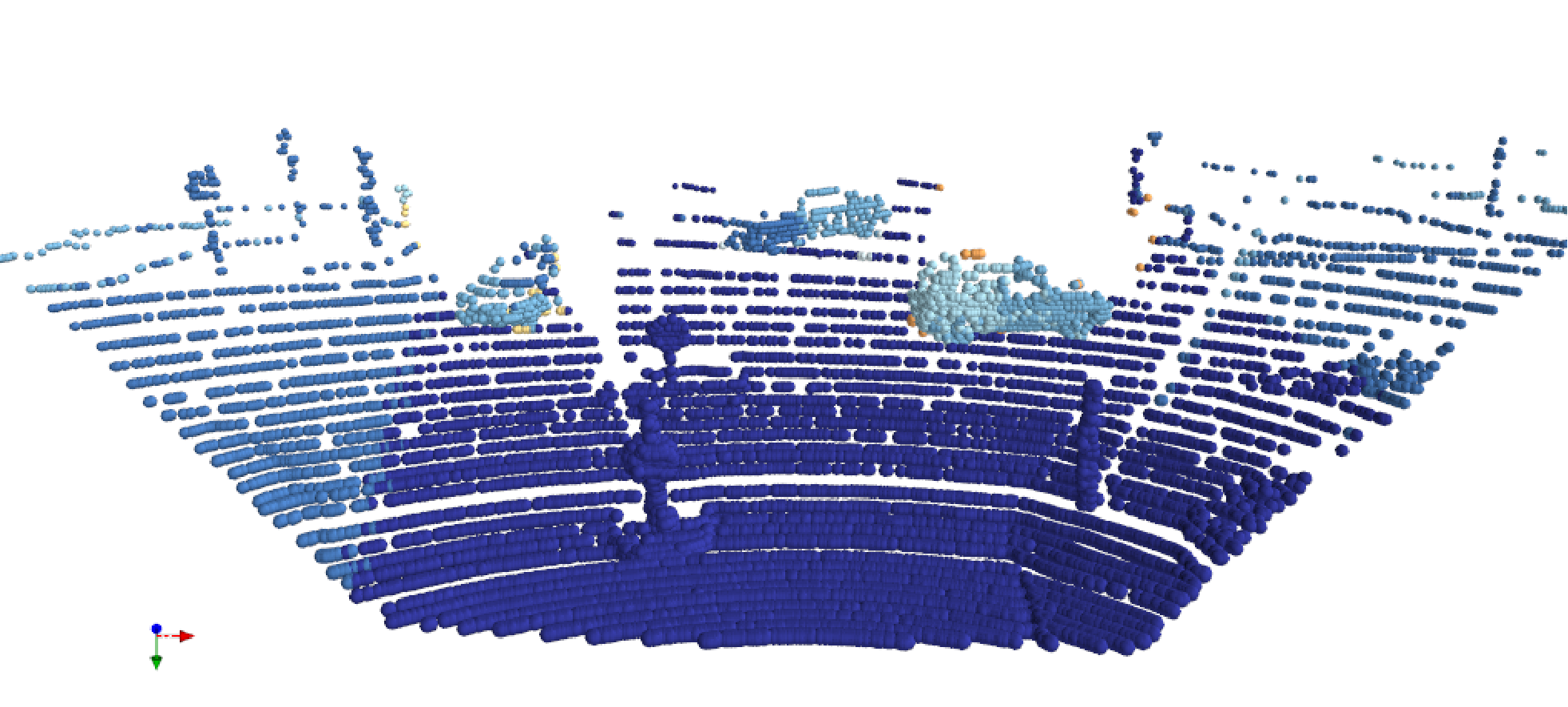}&
		\includegraphics[width=0.29\linewidth,valign=c]{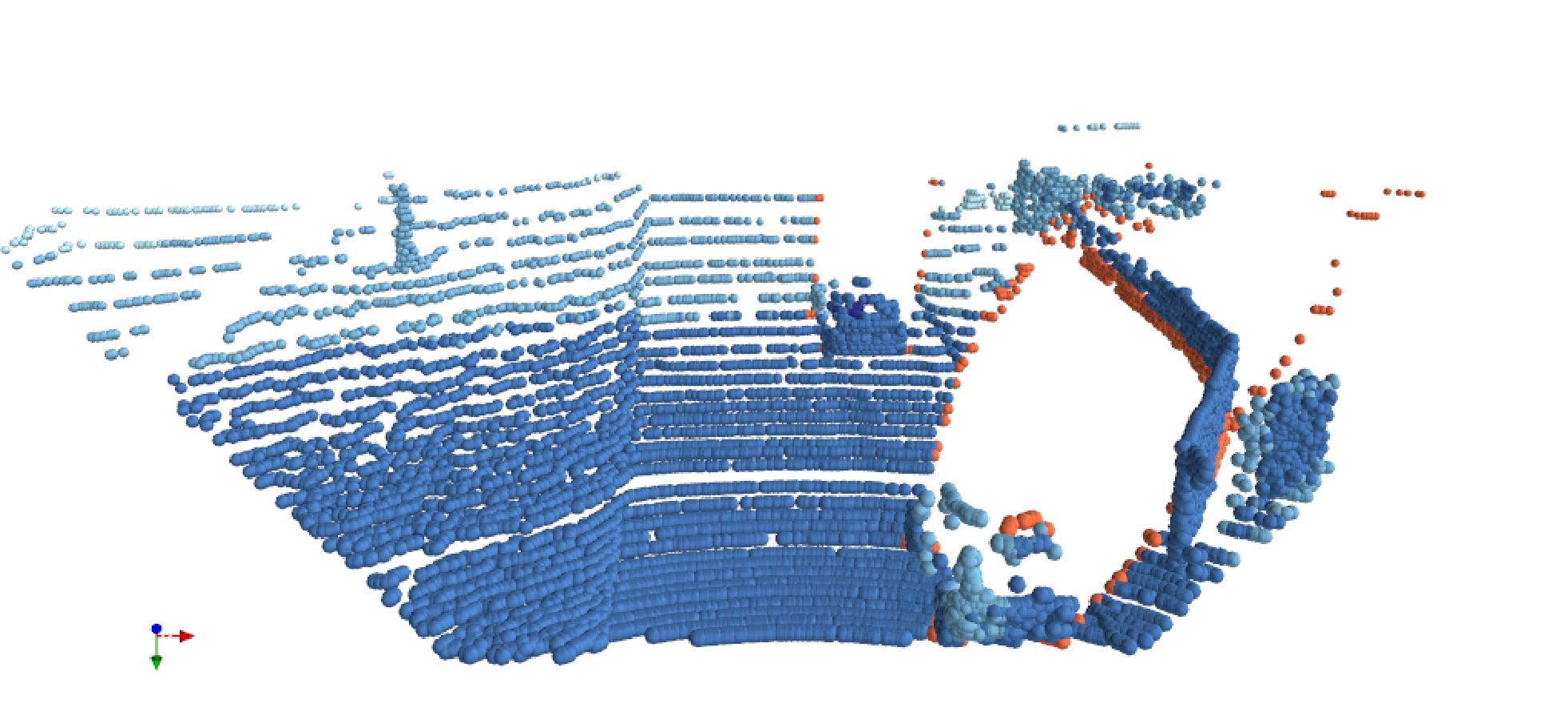}&
		\includegraphics[width=0.29\linewidth,valign=c]{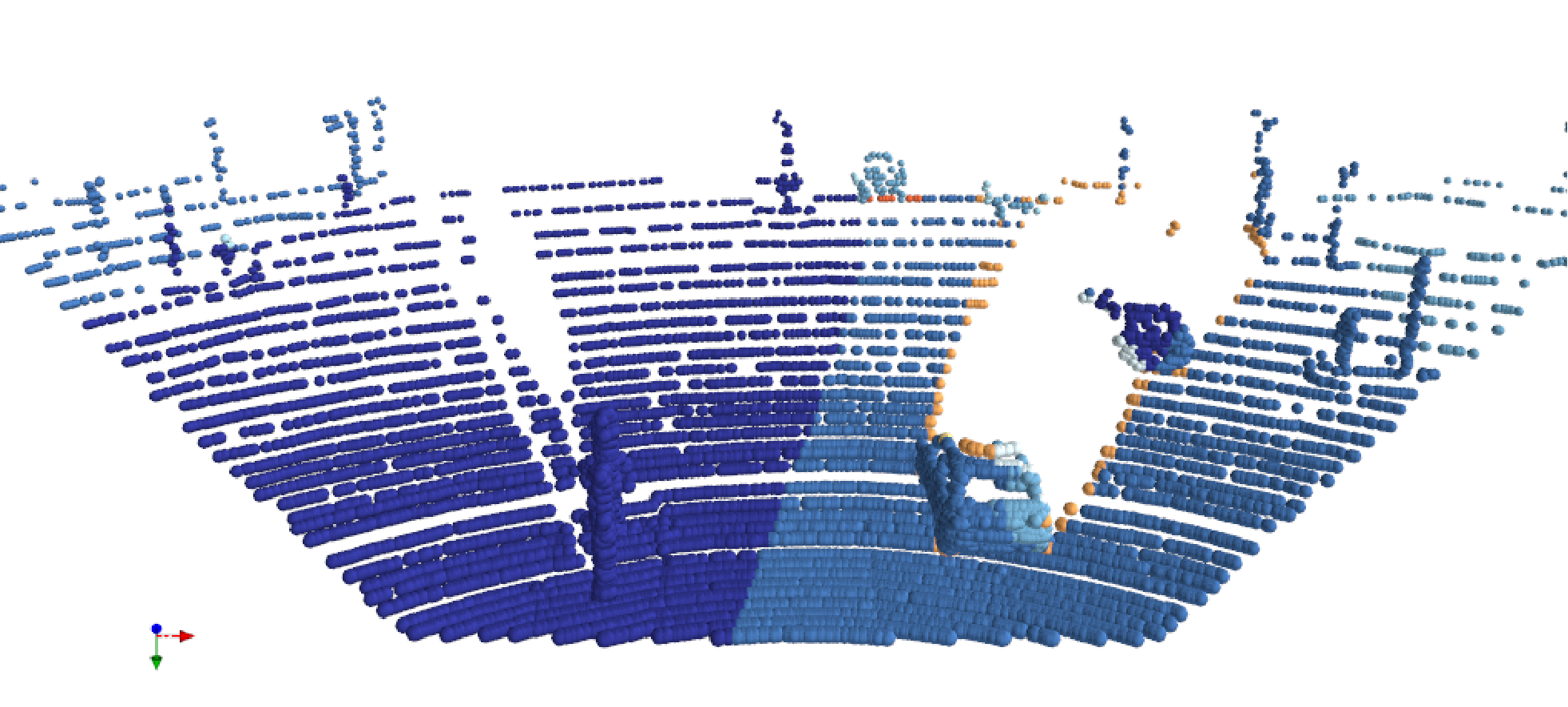}
		\Tstrut\Bstrut\\	
		
		\hline

		& \textit{Example 4} & \textit{Example 5} & \textit{Example 6} 
		\Tstrut\Bstrut\\ 		
		\rotatebox[origin=c]{90}{\textit{Scene}} &
		\includegraphics[width=0.29\linewidth,valign=c]{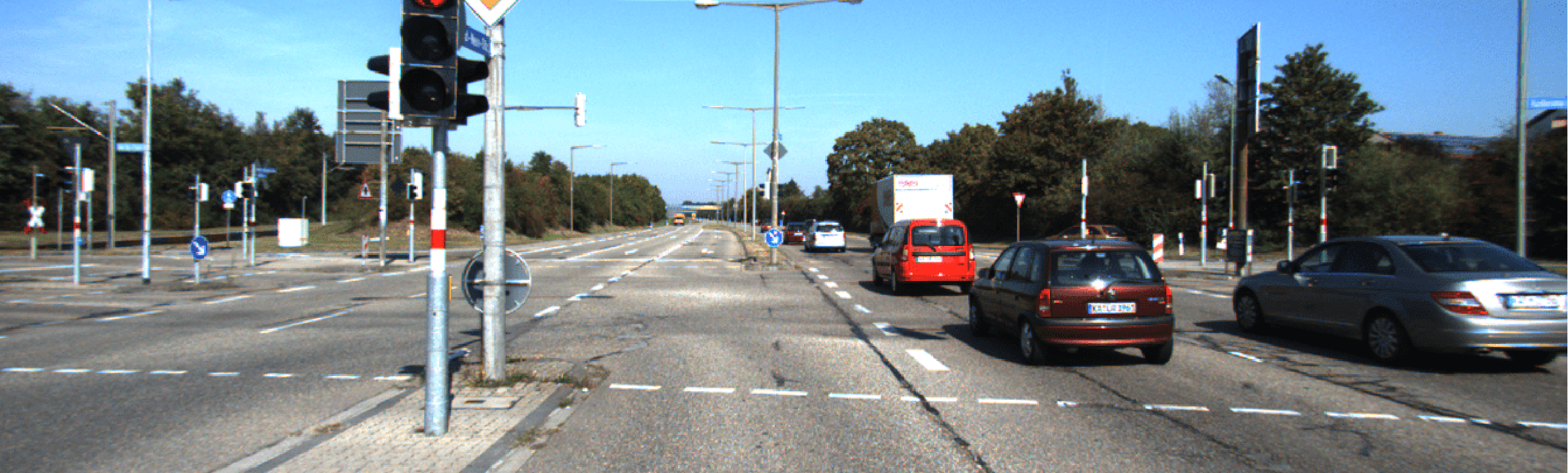}&
		\includegraphics[width=0.29\linewidth,valign=c]{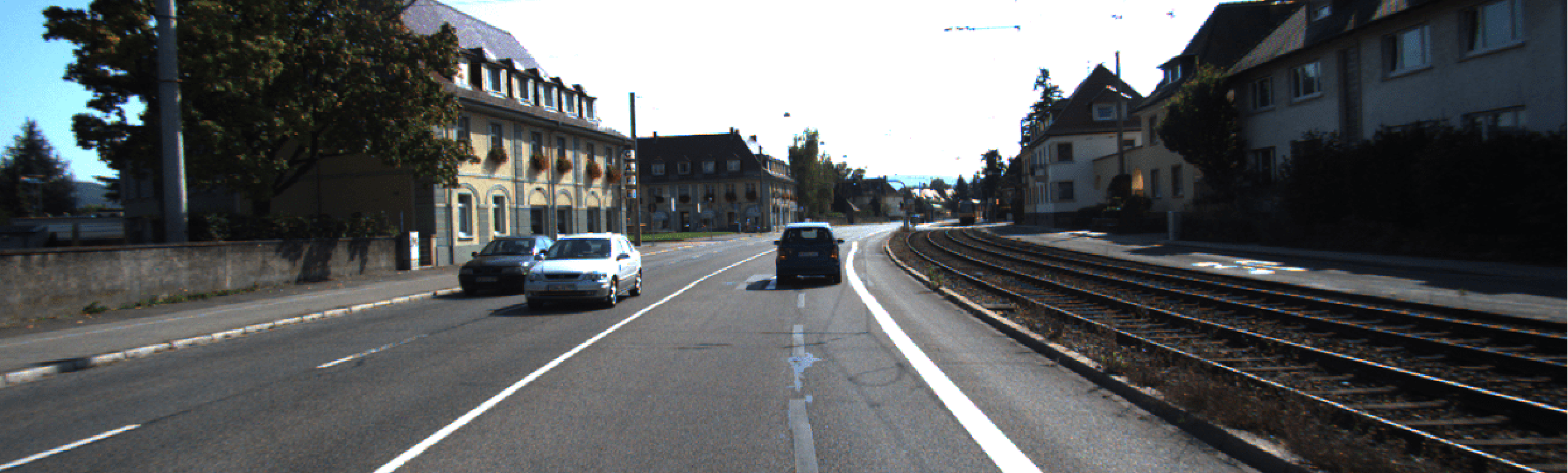}&
		\includegraphics[width=0.29\linewidth,valign=c]{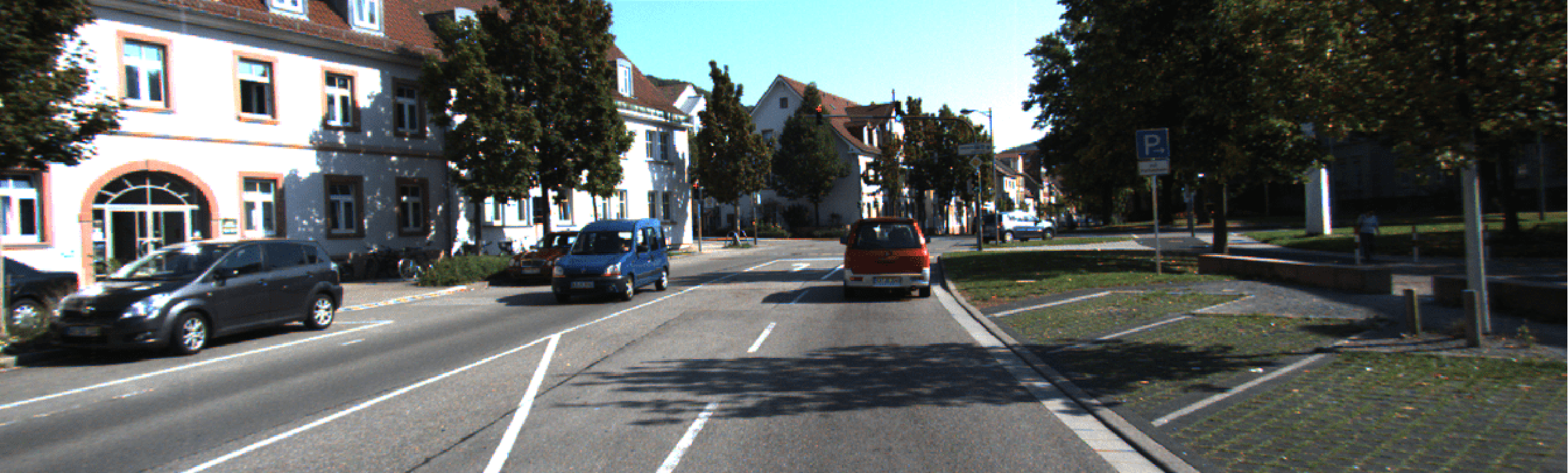}
		\Tstrut\Bstrut\\			
		\rotatebox[origin=c]{90}{\textit{$M^P_{fg}$}} &
		\includegraphics[width=0.29\linewidth,valign=c]{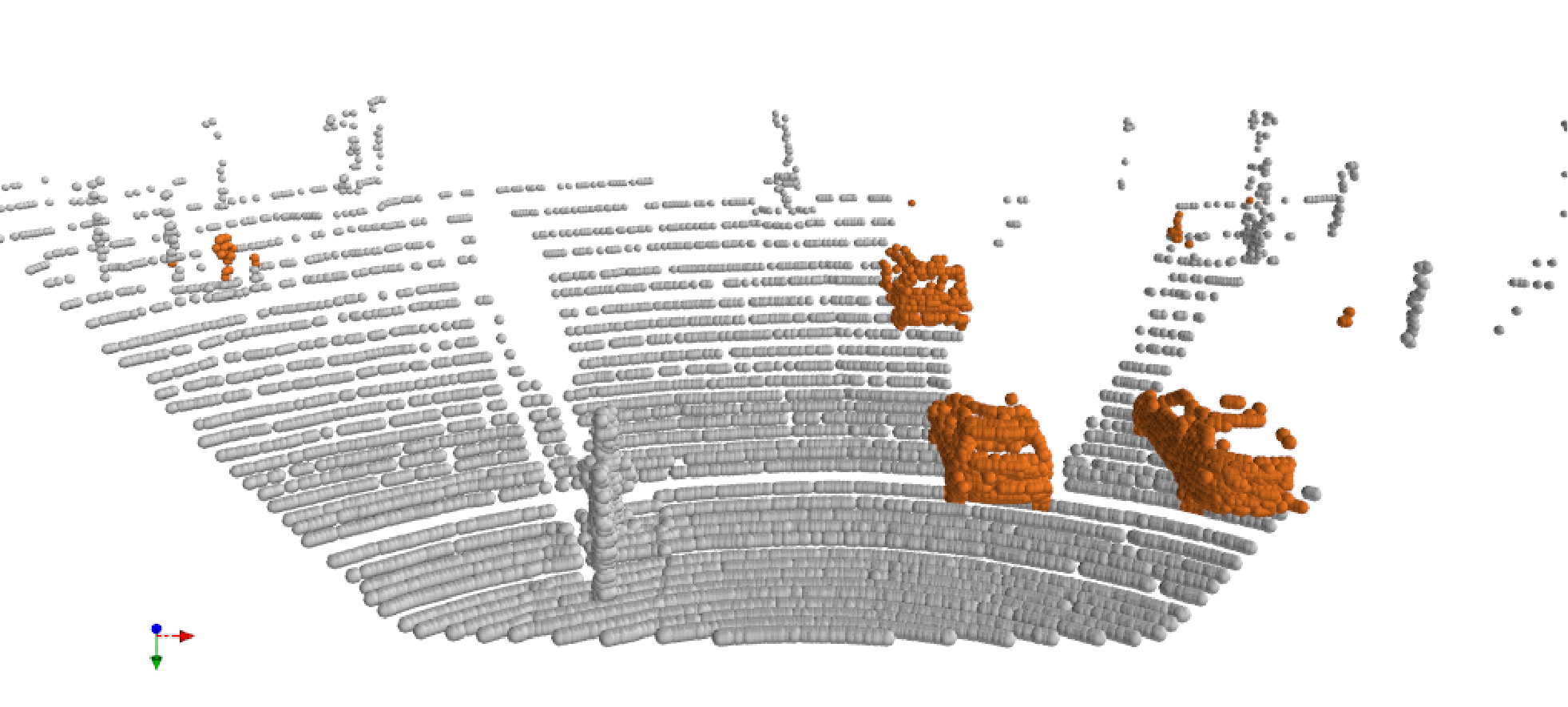}&
		\includegraphics[width=0.29\linewidth,valign=c]{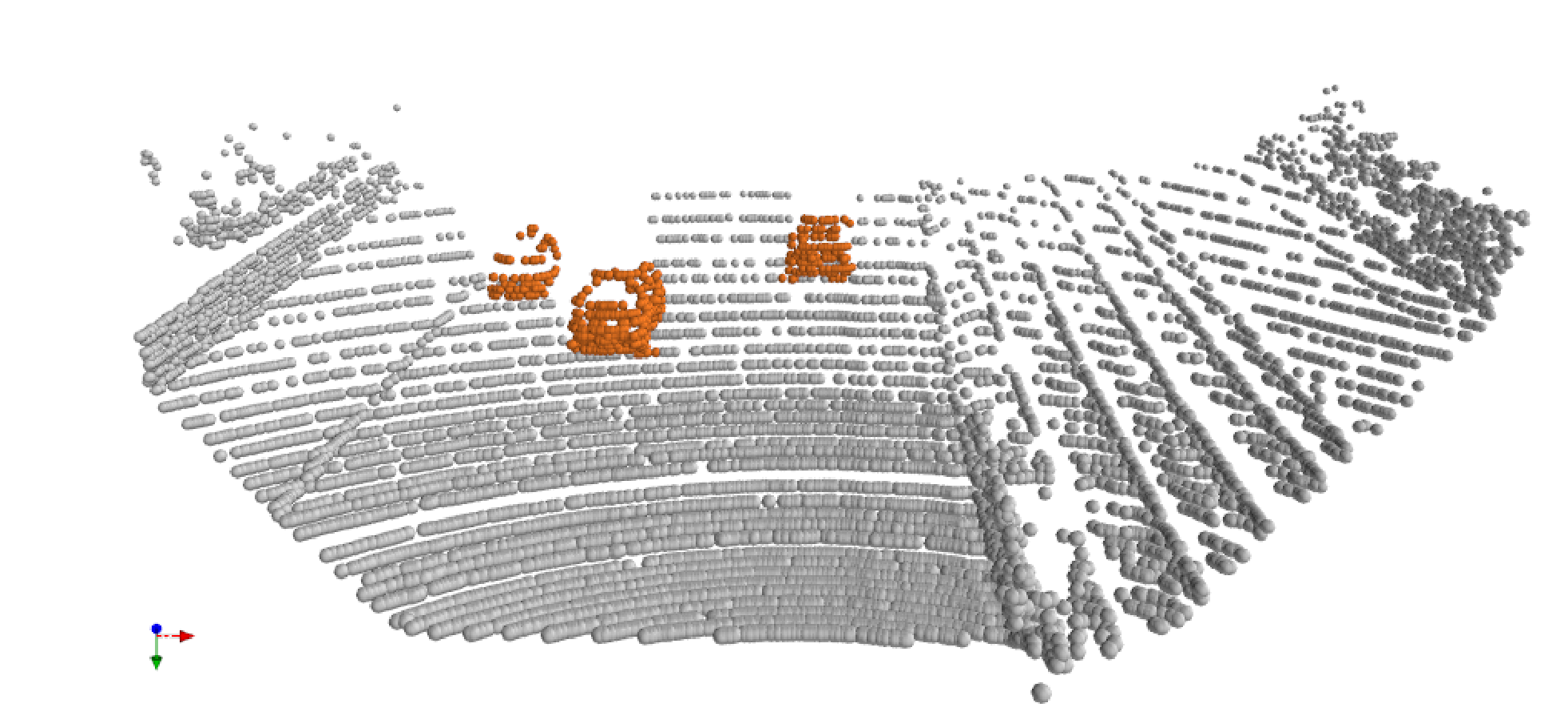}&
		\includegraphics[width=0.29\linewidth,valign=c]{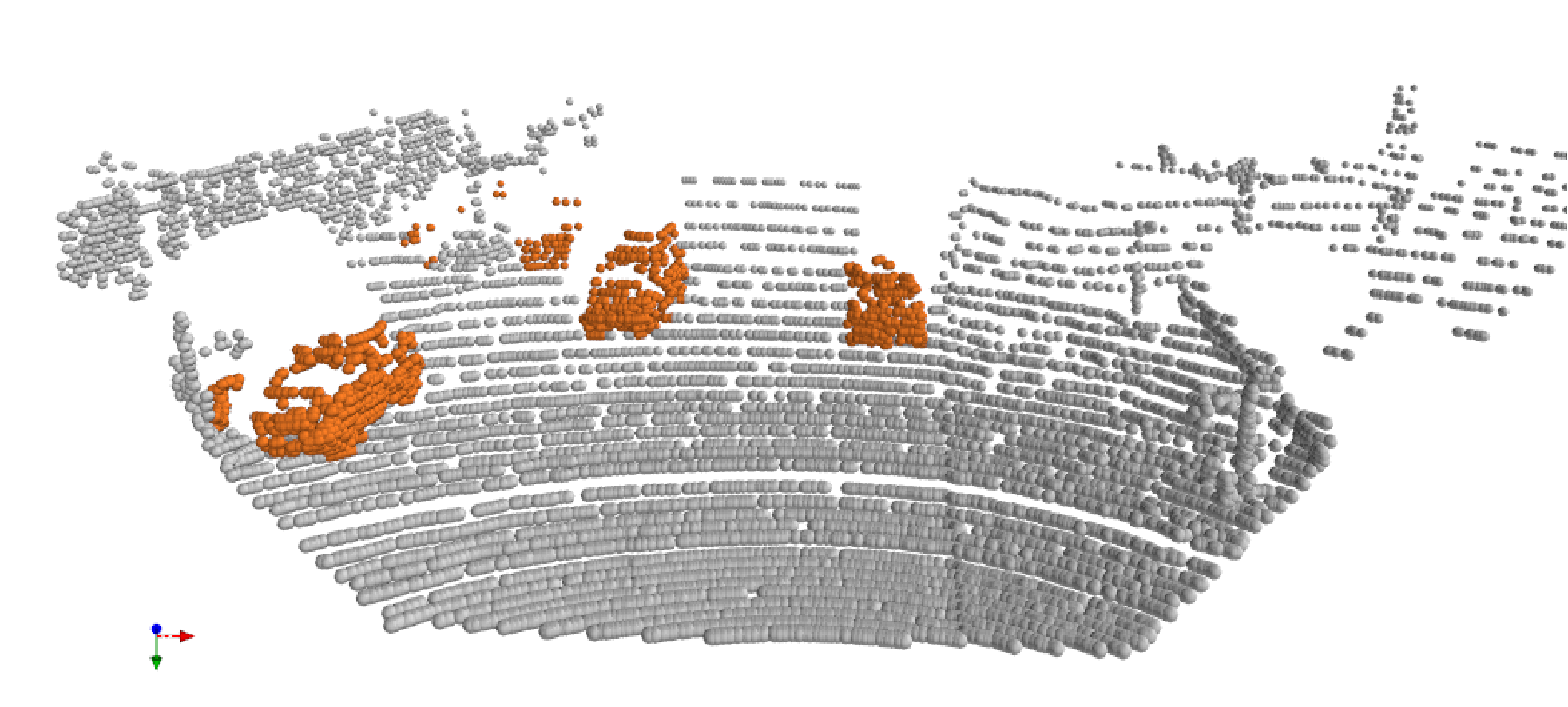}
		\Tstrut\Bstrut\\
		
		\rotatebox[origin=c]{90}{\textit{$M^Q_{fg}$}} &
		\includegraphics[width=0.29\linewidth,valign=c]{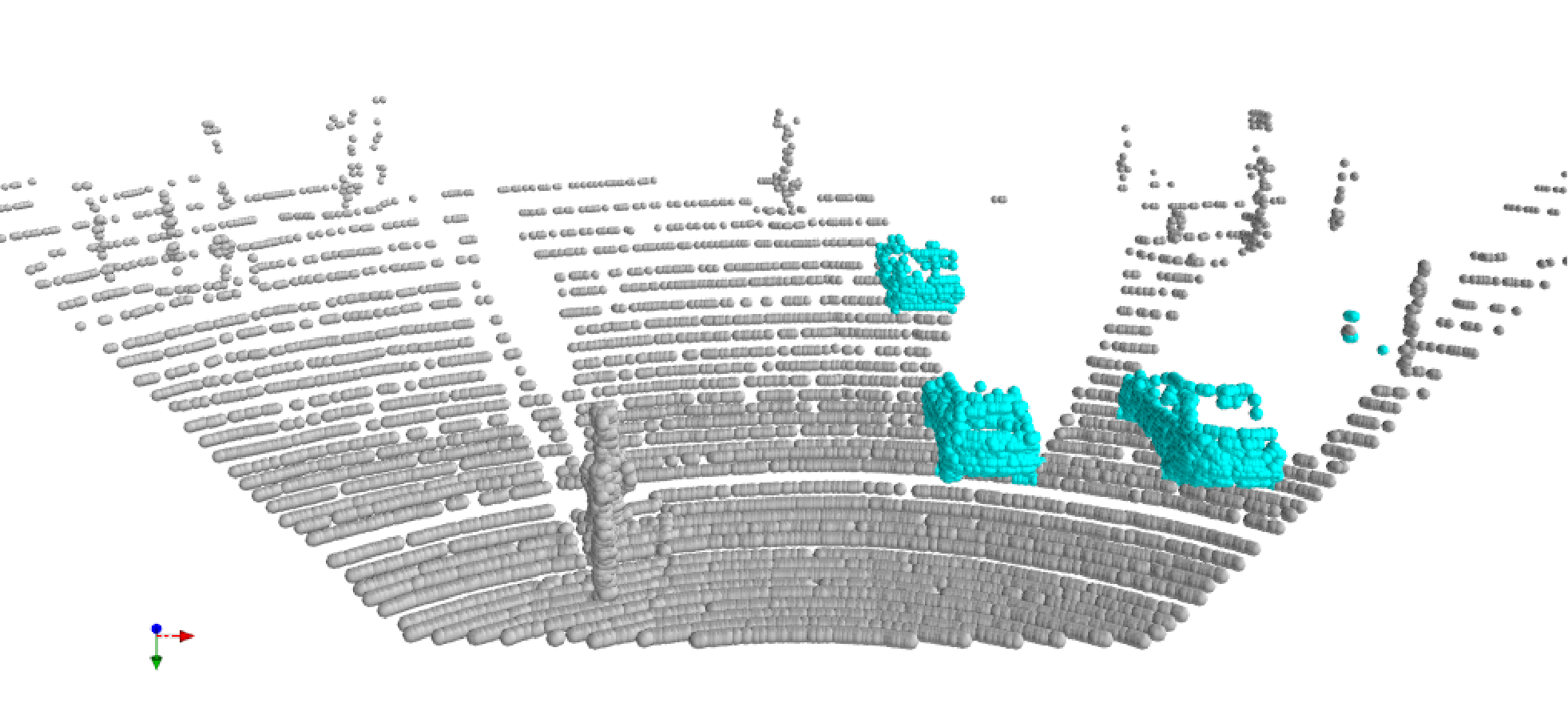}&
		\includegraphics[width=0.29\linewidth,valign=c]{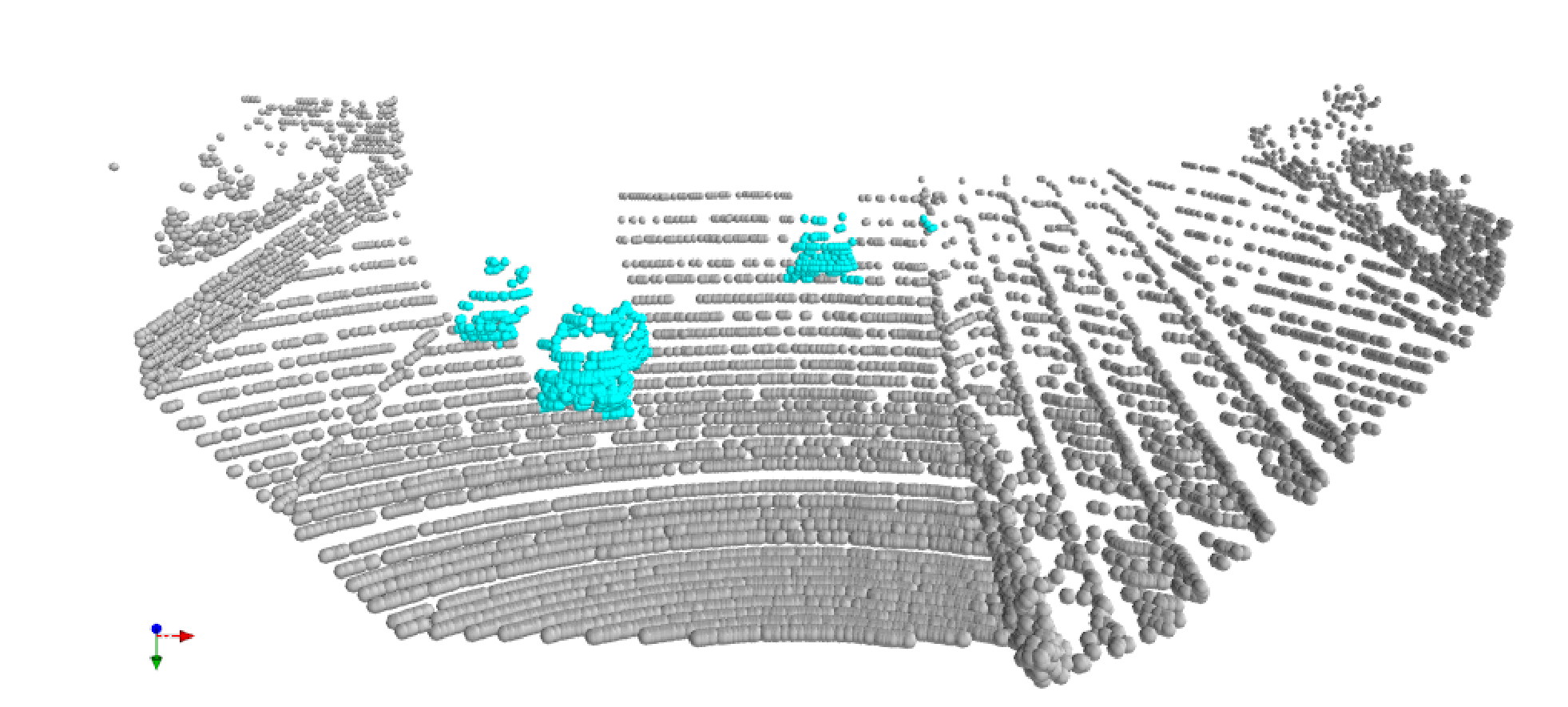}&
		\includegraphics[width=0.29\linewidth,valign=c]{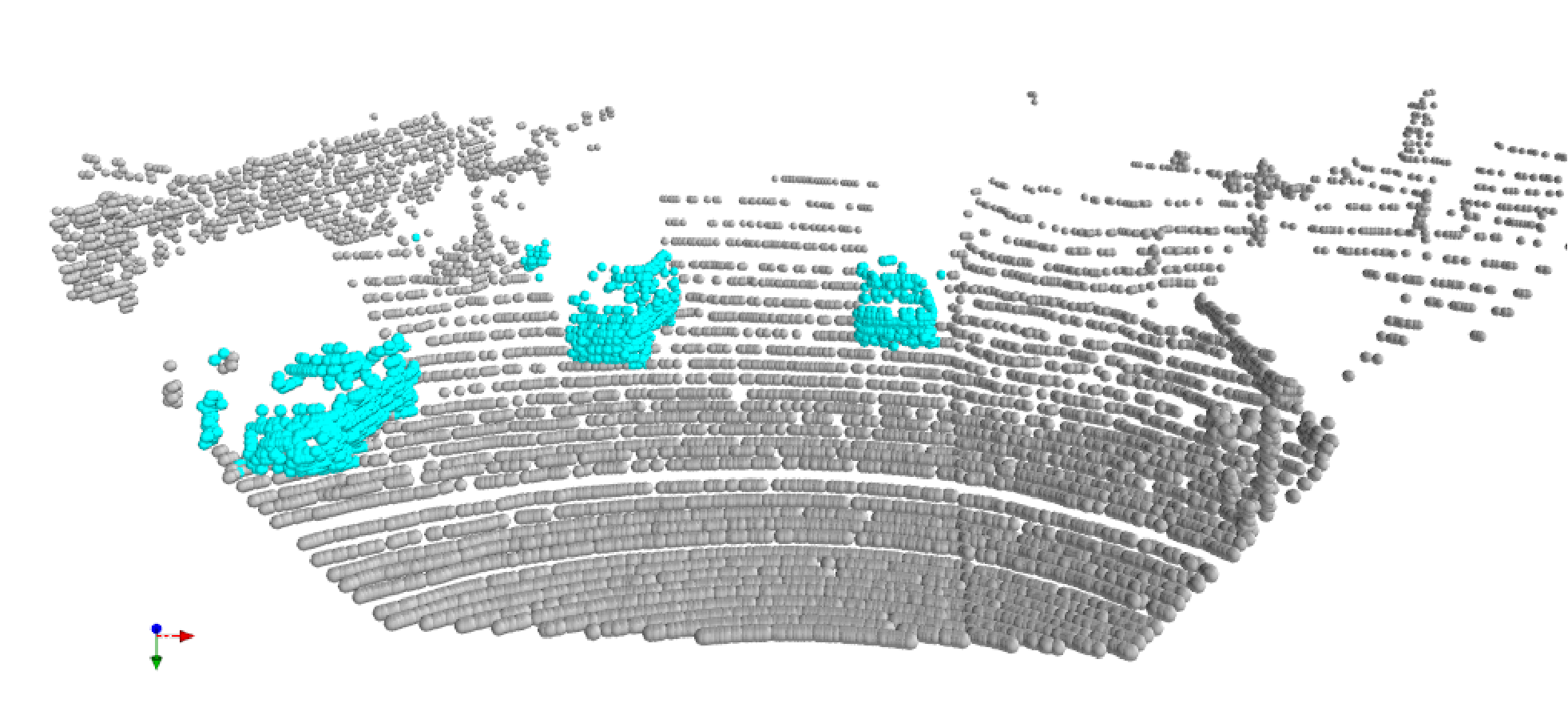}
		\Tstrut\Bstrut\\
		
		\rotatebox[origin=c]{90}{\textit{Error Map}} &
		\includegraphics[width=0.29\linewidth,valign=c]{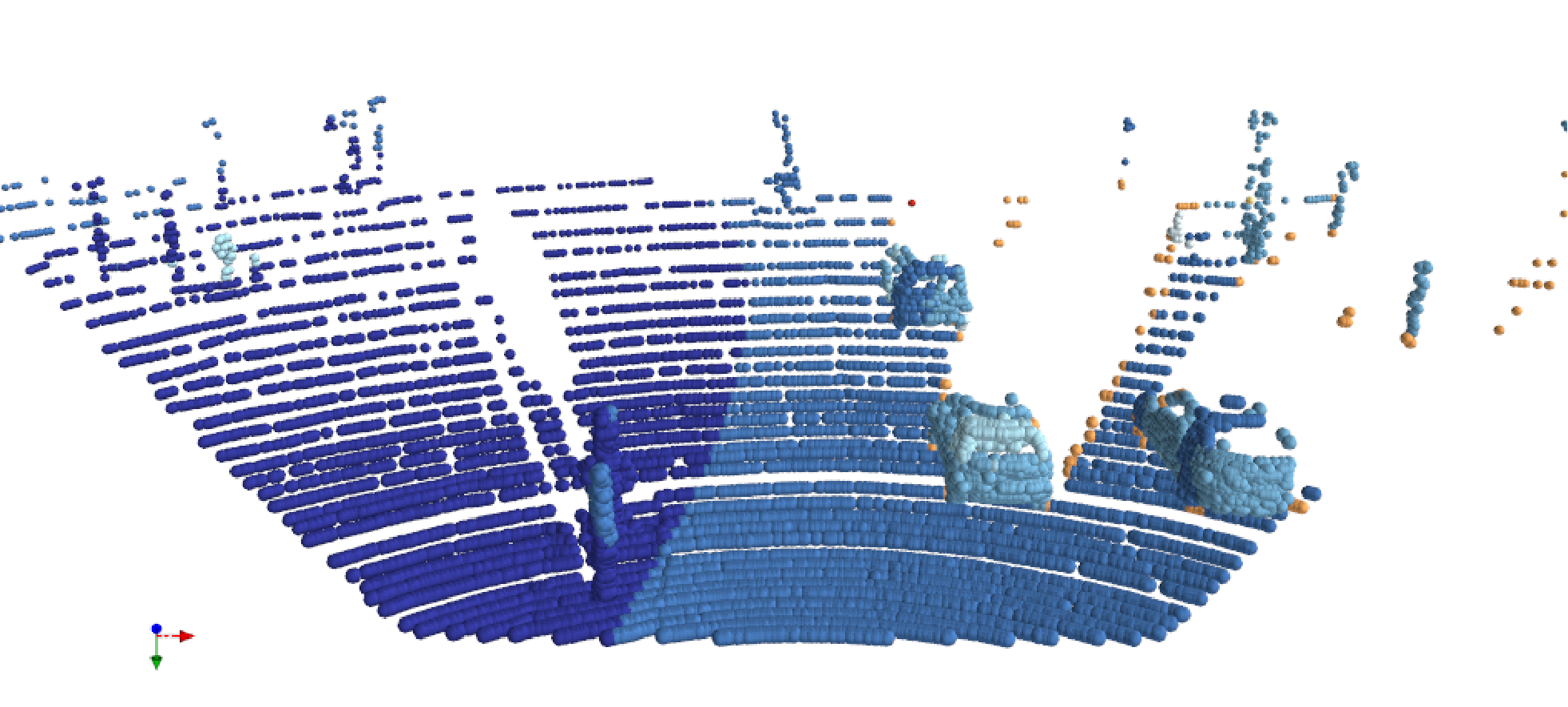}&
		\includegraphics[width=0.29\linewidth,valign=c]{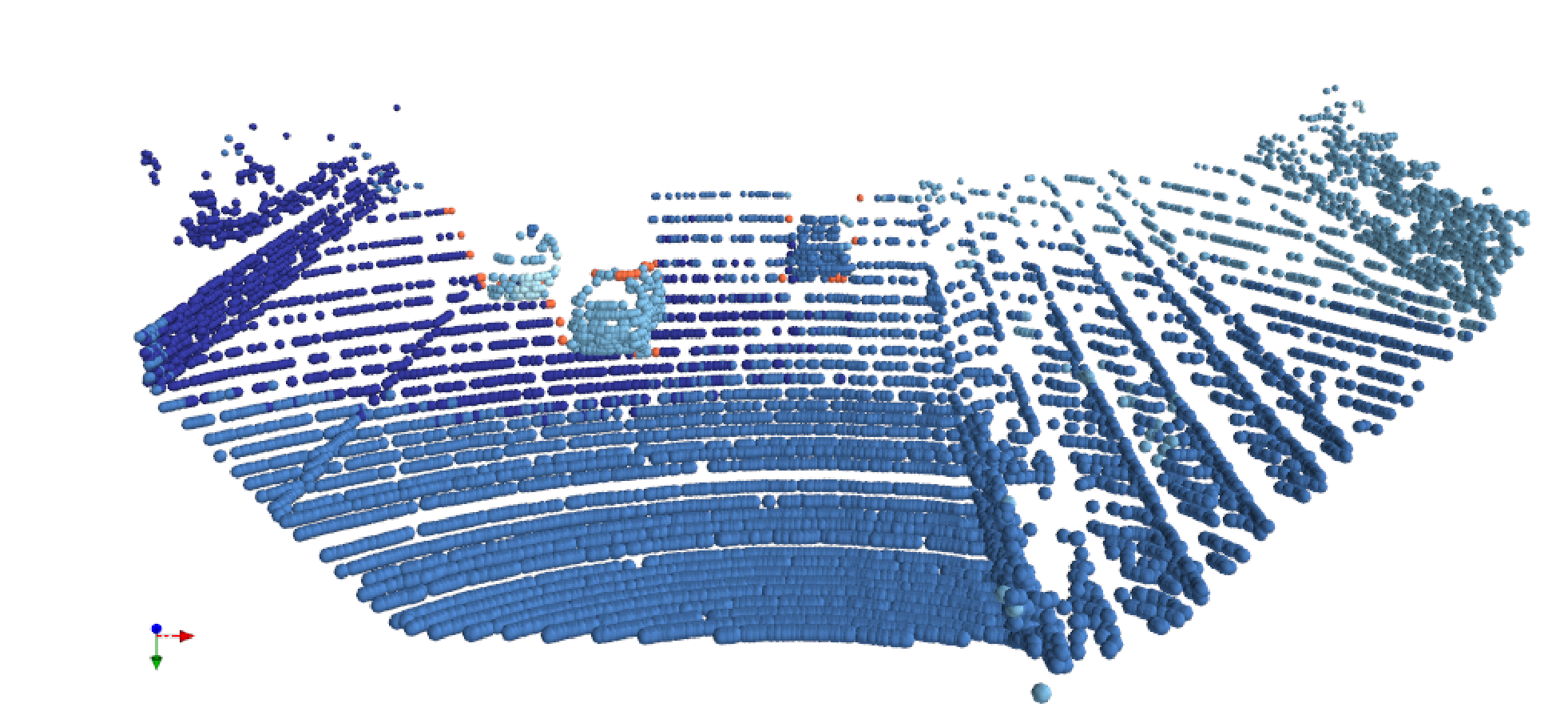}&
		\includegraphics[width=0.29\linewidth,valign=c]{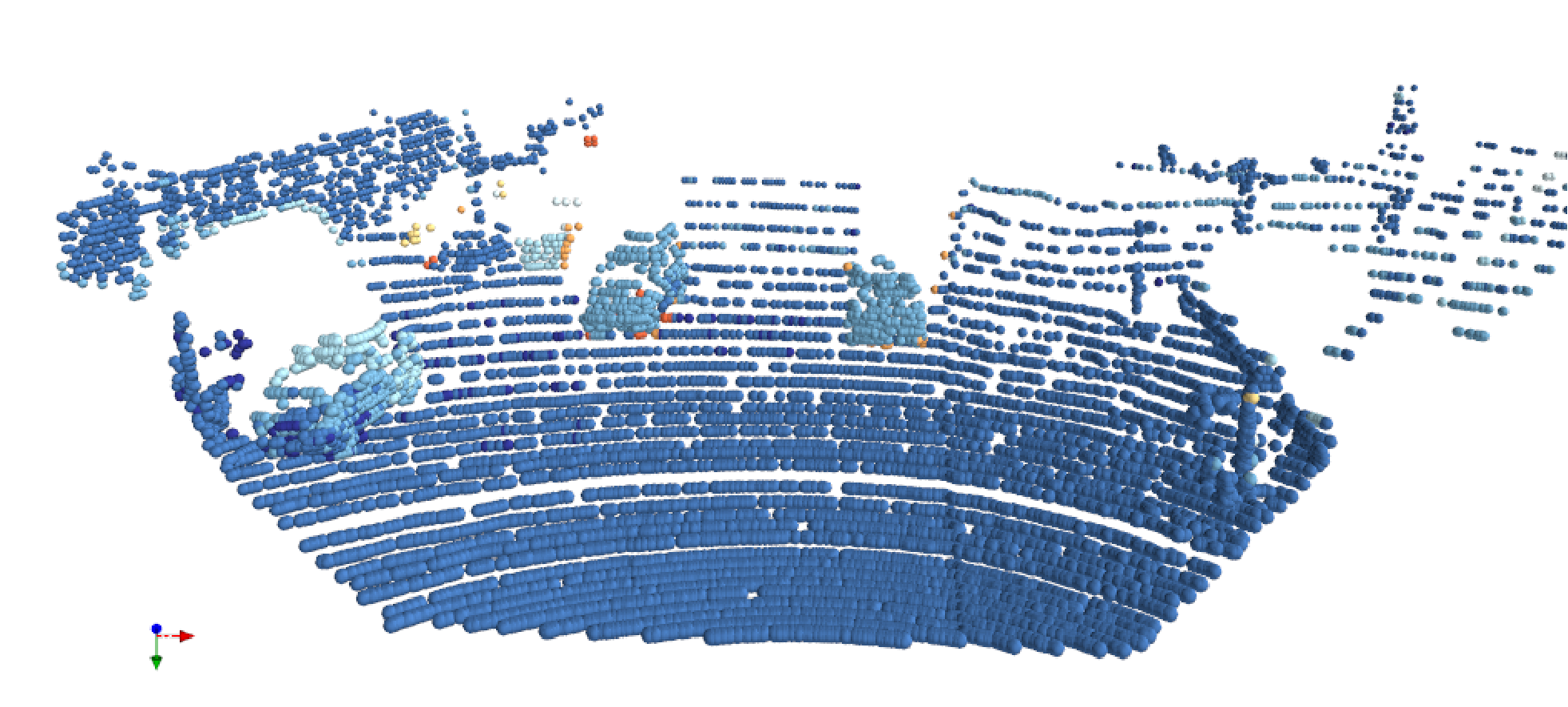}
		\Tstrut\Bstrut\\
		& \multicolumn{3}{c}{\includegraphics[width=0.9\linewidth,valign=c]{eps_min/KITTI_errorcolors_3D-min.eps}}\Tstrut\Bstrut\\
	\end{tabular}
	\caption{Six examples from $\mathrm{lidarKITTI}$~\cite{geiger2012we} show the qualitative results of our \name{}.}
	\label{Figure:qual_lidar}
\end{figure}

We visualize our predicted masks and the error maps of scene flow of six examples from $\mathrm{stereoKITTI}$ in \Cref{Figure:qual_stereo} and another six examples from $\mathrm{lidarKITTI}$ in \Cref{Figure:qual_lidar}.

\bibliography{EgoFlowNet}
\end{document}